\pdfoutput=1

\documentclass[11pt]{article}

\usepackage{acl}

\usepackage{times}
\usepackage{latexsym}

\usepackage[T1]{fontenc}

\usepackage[utf8]{inputenc}

\usepackage{microtype}

\usepackage{subcaption}
\usepackage{microtype}
\usepackage{amsmath}
\usepackage{multirow}
\usepackage{soul}
\usepackage{pifont}
\usepackage{times}
\usepackage{latexsym}
\usepackage{graphicx}
\usepackage{booktabs}
\usepackage{subcaption}
\usepackage[ruled,vlined]{algorithm2e}
\usepackage{multirow}
\usepackage{tabularx}
\usepackage{amsmath}
\usepackage{bbold}
\usepackage{mathtools}
\usepackage{xcolor}
\usepackage{booktabs}
\usepackage{tabularx,ragged2e}
\usepackage{enumitem}
\usepackage{scrextend}
\newcommand*\samethanks[1][\value{footnote}]{\footnotemark[#1]}
\usepackage{amssymb}

\let\emptyset\varnothing

\usepackage{xcolor}
\title{NDA: A Multi-Stage Task for \underline{N}ovelty \underline{D}etection and \underline{A}ccommodation in Authorship Attribution}

\title{NoveltyTask: A Multi-Stage Benchmark for Novelty Detection and Accommodation in NLP with an Instantiation in Authorship Attribution}

\title{A Unified Evaluation Framework for Novelty Detection and Accommodation in NLP with an Instantiation in Authorship Attribution}

\author{Neeraj Varshney$^{1}$\thanks{~~Equal Contribution, Contact email: hgupta35@asu.edu} \hspace{9pt} Himanshu Gupta$^{1}$\samethanks \hspace{9pt} Eric Robertson$^{2}$ \hspace{9pt} Bing Liu$^{3}$\hspace{9pt} Chitta Baral$^{1}$\\
$^1$ Arizona State University \; $^2$ PAR Government Systems Corporation \\\; $^3$ University of Illinois at Chicago }
\begin{document}
\maketitle
\begin{abstract}

State-of-the-art natural language processing models have been shown to achieve remarkable performance in `closed-world' settings where all the labels in the evaluation set are known at training time. 
However, in real-world settings, `novel' instances that do not belong to any known class are often observed. 
This renders the ability to deal with novelties crucial.
To initiate a systematic research in this important area of `dealing with novelties', we introduce \textit{NoveltyTask}, a multi-stage task to evaluate a system's performance on pipelined novelty `detection' and `accommodation' tasks.
We provide mathematical formulation of NoveltyTask and instantiate it with the authorship attribution task that pertains to identifying the correct author of a given text.
We use Amazon reviews corpus and compile a large dataset (consisting of $250k$ instances across $200$ authors/labels) for NoveltyTask.
We conduct comprehensive experiments and explore several baseline methods for the task. 
Our results show that the methods achieve considerably low performance making the task challenging and leaving sufficient room for improvement.
Finally, we believe our work will encourage research in this underexplored area of dealing with novelties, an important step en route to developing robust systems.


\end{abstract}

\section{Introduction}


Recent advancements in Natural Language Processing (NLP) have led to the development of several pre-trained large-scale language models such as BERT \cite{devlin-etal-2019-bert}, RoBERTa \cite{Liu2019RoBERTaAR}, and ELECTRA \cite{clark2020electra}.
These models have been shown to achieve remarkable performance in \textit{closed-world} settings where all the labels in the evaluation set are known at training time. 
However, in real-world settings, this assumption is often violated as instances that do not belong to any known label (`novel' instances) are also observed.
This renders the ability to deal with novelties crucial in order to develop robust systems for real-world applications.


The topic of novelty is getting increased attention in the broad AI research \cite{Boult2020AUF, KR2021-43,Rambhatla2021ThePO}.
Also, in NLP, the `novelty detection' task in which novel instances need to be identified is being explored \cite{ghosal-etal-2018-novelty, ma-etal-2021-semantic}; related problems such as anomaly detection \cite{chalapathy2019deep}, out-of-domain detection, and open-set recognition \cite{hendrycks17baseline,hendrycks-etal-2020-pretrained,NEURIPS2019_8558cb40} are also being studied.
In addition to the task of `detection', dealing with novelties also requires `\textit{accommodation}' that pertains to learning from the correctly detected novelties.
Despite having practical significance, this aspect of dealing with novelties has remained underexplored. 
Furthermore, dealing with novelties is a crucial step in numerous other practical applications such as concept learning, continual learning, and domain adaptation.






\begin{figure*}
    \centering 
    \small
    \includegraphics[width=0.98\linewidth]{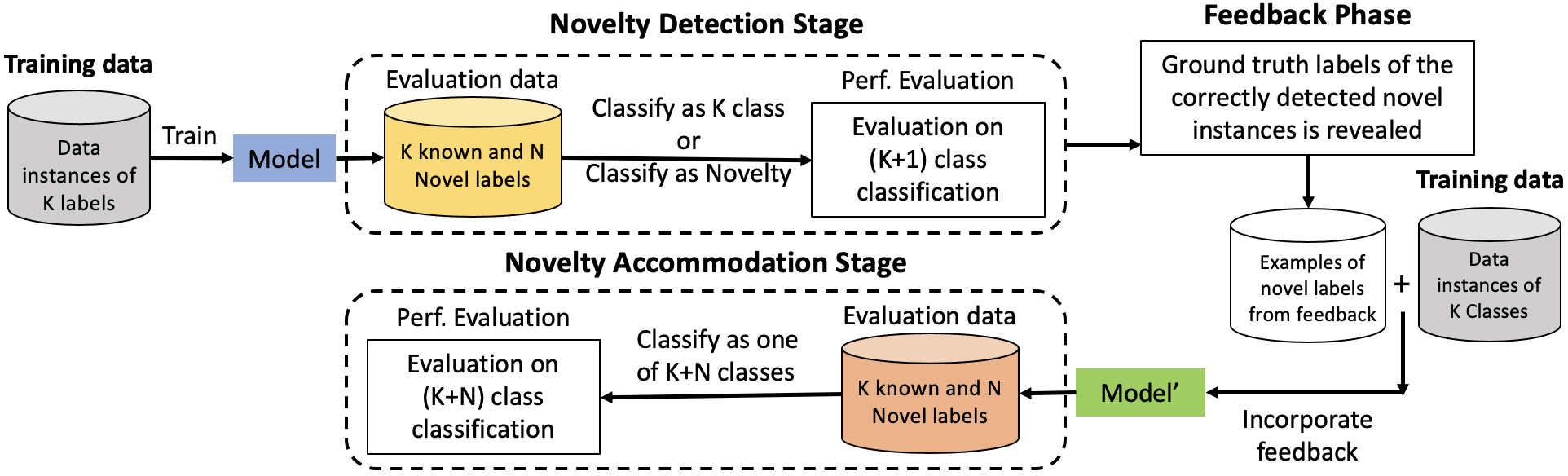}
    \caption{
    Illustrating the multi-stage pipelined formulation of \textbf{NoveltyTask}. 
    Initially, examples of a set of $K$ labels (`\textbf{known labels}') are provided for training a classification system. The first evaluation stage i.e. the `\textbf{novelty detection}' stage consists of evaluation instances from the $K$ known labels and `$N$' novel labels. For each instance, the system needs to either classify it to one of the $K$ known classes or report it as novel (not from any of the $K$ known classes) i.e. the system is evaluated on a $(K+1)$ class classification problem. This stage is followed by a \textbf{Feedback phase} in which  the ground truth label of the novel instances that the system correctly reports as novel is revealed. 
    The system then needs to leverage these new examples (of the novel labels) for the second evaluation stage (\textbf{novelty accommodation}) in which it is evaluated on a $(K+N)$ class classification problem.
    }
    \label{fig:block}
\end{figure*}

To initiate systematic research in this area of `dealing with novelties', we formulate a multi-stage task called \textbf{NoveltyTask}.
Initially, a dataset consisting of examples of a set of labels (referred to as `known labels') is provided for training and then sequential evaluation is conducted in two stages: \textbf{Novelty Detection} and \textbf{Novelty Accommodation}.
Both these stages include distinct unseen evaluation instances belonging to both `known labels' (labels present in the training dataset) and `novel labels' (labels not present in the training dataset).

In the first evaluation stage i.e. the \textbf{novelty detection} stage, the system needs to either identify an instance as novel or classify it to one of the `$K$' known labels.
This is the same as the $(K+1)$ class classification problem (where $K$ corresponds to the number of known labels) used in standard anomaly/OOD detection tasks.
This evaluation stage is followed by a \textbf{feedback} phase in which the ground truth label of the novel instances (from the detection stage) that get correctly reported as novel is revealed. 
Essentially, in the feedback phase, the system gets some examples of the novel labels (from the evaluation instances of the detection stage) that it correctly identified as novel. 

In addition to the initially provided training examples of the $K$ known labels, the system can leverage these new examples of the novel labels for the next evaluation stage, the \textbf{novelty accommodation} stage. 
This stage also has evaluation instances from both the known and the novel labels (distinct and mutually exclusive from the detection stage); however, in this stage, the system needs to identify the true label of the evaluation instances, i.e. it's a $(K + N)$ class classification problem where $N$ corresponds to the number of novel labels.
We summarize this multi-stage task in Figure \ref{fig:block}.
We note that \textbf{NoveltyTask is a controlled task/framework for evaluating a system's ability to deal with novelties and not a method to improve its ability.}




It is intuitive that the ability to deal with the novelties should be directly correlated with the ability to detect the novelties; our two-stage pipelined formulation of NoveltyTask allows achieving this desiderata as higher accuracy in correctly detecting the novelties will result in more feedback i.e. more examples of the novel labels that will eventually help in achieving higher performance in the accommodation stage.
However, in the detection stage, the system needs to balance the trade-off between reporting instances as novel and classifying them to the known labels.
Consider a trivial system that simply flags all the evaluation instances of the detection stage as novel in order to get the maximum feedback; 
such a system will get the true ground-truth label (novel label) of all the novel instances present in the detection stage and will eventually perform better in the accommodation stage but it would have to sacrifice its classification accuracy in the detection stage (especially on instances of the known labels). 
We address several such concerns in formulating the performance metrics for NoveltyTask (Section \ref{sec_noveltybench}).

In this work, we instantiate NoveltyTask with \textbf{authorship attribution} task in which each author represents a label and the task is to identify the correct author of a given unseen text.
However, we note that the formulation of NoveltyTask is general and applicable to all tasks.
We leverage product reviews from Amazon corpus \cite{mcauley2015image,he2016ups} for the attribution task.
We explore several baseline methods for both detection and accommodation tasks (Section \ref{sec_experiments}).

In summary, our contributions are as follows:
\begin{enumerate}[noitemsep,nosep,leftmargin=*]

    \item We \textbf{define a unified task for `dealing with novelties'} consisting of both novelty detection and novelty accommodation.
    

    \item We \textbf{provide a controlled evaluation framework} with its mathematical formulation.
    
    \item We \textbf{instantiate NoveltyTask} with the Authorship Attribution task.
    
    \item We \textbf{study the performance of several baseline methods} for NoveltyTask.

\end{enumerate}

\section{Background and Related Work}
In this section, we first discuss the related work on novelty/OOD/anomaly detection tasks and then detail the authorship attribution task.

\subsection{Novelty/OOD/Anomaly Detection}
Novelty Detection and its related tasks such as out-of-distribution detection, selective prediction, and anomaly detection have attracted a lot of research attention from both computer vision \cite{fort2021exploring,esmaeilpour2022zero,sun2021m2iosr,lu2022pmal,liu2020few,perera2020generative, whitehead2022reliablevqa} and language \cite{qin2020text,venkataram2018open,yang2022one, varshney-etal-2022-investigating,kamath-etal-2020-selective, varshney-etal-2022-towards} research communities. 
OOD detection for text classification is an active area of research in NLP. 
\citet{qin2020text} follow a pairwise matching paradigm and calculate the probability of a pair of samples belonging to the same class. 
\citet{yang2022one} investigate how to detect open classes efficiently under domain shift. 
\citet{ai-etal-2022-whodunit} propose a contrastive learning paradigm, a technique that brings similar samples close and pushes dissimilar samples apart in the vector representation space. 
\citet{yilmaz2022d2u} propose a method for detecting out-of-scope utterances utilizing the confidence score for a given utterance. 

\subsection{Authorship Attribution}
Authorship attribution task (AA) pertains to identifying the correct author of a given text.
AA has been studied for short texts \cite{Aborisade2018ClassificationFA} such as tweets as well as long texts such as court judgments \cite{sari-etal-2018-topic}. 
Traditional approaches for AA explore techniques based on n-grams, word embeddings, and stylometric features such as the use of punctuation, average word length, sentence length, and number of upper cases \cite{sari-etal-2018-topic,8424720,soler-company-wanner-2017-relevance}.
Transformer-based models have been shown to outperform the traditional methods on this task \cite{fabien-etal-2020-bertaa, tyo2021siamese, manolache2021transferring,10.1007/978-3-030-28577-7_17}.

\section{NoveltyTask}
\label{sec_noveltybench}

NoveltyTask is a two-stage pipelined framework to evaluate a system's ability to deal with novelties.
In this task, examples of a set of labels (referred to as \textbf{known} labels) are made available for initial training. The system is sequentially evaluated in two stages: novelty detection and novelty accommodation.
Both these stages consist of distinct unseen evaluation instances belonging to both `known' and `novel' labels.
We define a label as \textbf{novel} if it is not one of the known labels provided for initial training and all instances belonging to the novel labels are referred to as novel instances.
We summarize this multi-stage task in Figure \ref{fig:block}.
In this section, we provide a mathematical formulation of NoveltyTask, detail its performance metrics, and describe the baseline methods.


\subsection{Formulation}


\subsubsection{Initial Training ($D^{T}$)}

Consider a dataset $D^{T}$ of $(x, y)$ pairs where $x$ denotes the input instance and $y$ $\in$ $\{1,2,...,K\}$ denotes the class label. 
We refer to this label set of $K$ classes as `known labels.'
In NoveltyTask, the classification dataset $D^T$ is provided for initial training. Then, the trained system is evaluated in the novelty detection stage as described in the next subsection.




\subsubsection{Novelty Detection ($Eval_{Det}$)}
The evaluation dataset of this stage ($Eval_{Det}$) consists of unseen instances of both known and novel labels, i.e., $Eval_{Det}$ includes instances from $K \cup N$ labels where $N$ corresponds to the number of novel labels not seen in the initial training dataset $D^T$.
Here, the system needs to do a $(K+1)$ class classification, i.e., for each instance, it can either output one of the $K$ known classes or report it as novel (not belonging to any known class) by outputting the $(0)^{th}$ class.
This is followed by the feedback phase described in \ref{sec_feedbak}.






\subsubsection{Feedback Phase ($D^F$)}
\label{sec_feedbak}
For each instance of the $Eval_{Det}$ dataset, we use an indicator function `$f$' whose value is $1$ if the instance is novel ( i.e. not from the $K$ known labels) and $0$ otherwise:
\begin{equation*}
    f(x) = \mathbb{1}[x \not\in \{1,2, ..., K\}]
\end{equation*}

In the feedback phase, we reveal the ground truth label of those novel instances (from $Eval_{Det}$) that the system correctly reports as novel, i.e., feedback results in a dataset ($D^F$) which is a subset of the novel instances of $Eval_{Det}$ where $f(x)$ is $1$ and the system's prediction on $x$ is the $(0)^{th}$ class.
\begin{equation*}
        D^F = 
            \begin{cases}
               & \text{$\in$ $Eval_{Det}$} \\
              (x,y), & f(x) = 1 \\
              & pred(x) = (0)^{th} class
            \end{cases}
\end{equation*}
Essentially, $D^F$ is a dataset that consists of examples of the novel labels.
The system can incorporate the feedback by leveraging $D^F$ in addition to the initial training dataset $D^T$ (refer to Section \ref{subsec_accommodation} for novelty accommodation methods) to adapt itself for the next evaluation stage, which is the novelty accommodation stage. 







\subsubsection{Novelty Accommodation ($Eval_{Acc}$)}
The system incorporates the feedback and is evaluated in the novelty accommodation stage on the $Eval_{Acc}$ dataset. 
Like the detection stage dataset, $Eval_{Acc}$ also includes instances of both $K$ known and $N$ novel labels (mutually exclusive from $Eval_{Det}$ i.e. $Eval_{Det} \cap Eval_{Acc} = \emptyset$).
However, in this stage, the system needs to identify the true label of \textbf{all} the evaluation instances (including those belonging to the novel labels) i.e. the task for the system is to do a $(K+N)$ class classification instead of a $(K+1)$ class classification.
Here, $N$ corresponds to the number of novel labels.
Essentially, in the feedback phase, the system gets some examples of the novel labels, and it needs to leverage them along with $D^T$ to classify the evaluation instances correctly. 

\textbf{Note that the feedback data $D^F$ may or may not contain examples of all the $N$ novel classes} as it totally depends on the system's ability to correctly detect novelties in the detection stage. The inability to detect instances of all the novel classes will accordingly impact the system's performance in the accommodation stage.
Next, we describe the performance metrics for both the stages.



\subsection{Performance Evaluation}


\paragraph{Novelty Detection:}
For the novelty detection stage, we use F1 score over all classes to evaluate the performance of the system. 
We also calculate the F1 score for the known classes ($\text{F1}_{\text{Known}}$) and for the novel instances ($\text{F1}_{\text{Novel}}$) to evaluate the fine-grained performances. 

Let $\{C_{1}, \cdots, C_{K}\}$ be the set of known classes and $C_{0}$ be the class corresponding to the novel instances, we calculate the micro F1 score using:
\begin{align*}
\text{F1} &= 2 \times  \frac{\text{P} \times \text{R}}{\text{P} + \text{R}},
\end{align*}

where P and R are precision and recall values.

Similarly, the F1 scores over known classes ($\text{F1}_{\text{known}}$) and novel class ($\text{F1}_{\text{Novel}}$) are computed.





\begin{figure*}[t]
\centering
    \begin{subfigure}{.32\linewidth}
        \includegraphics[width=\linewidth]{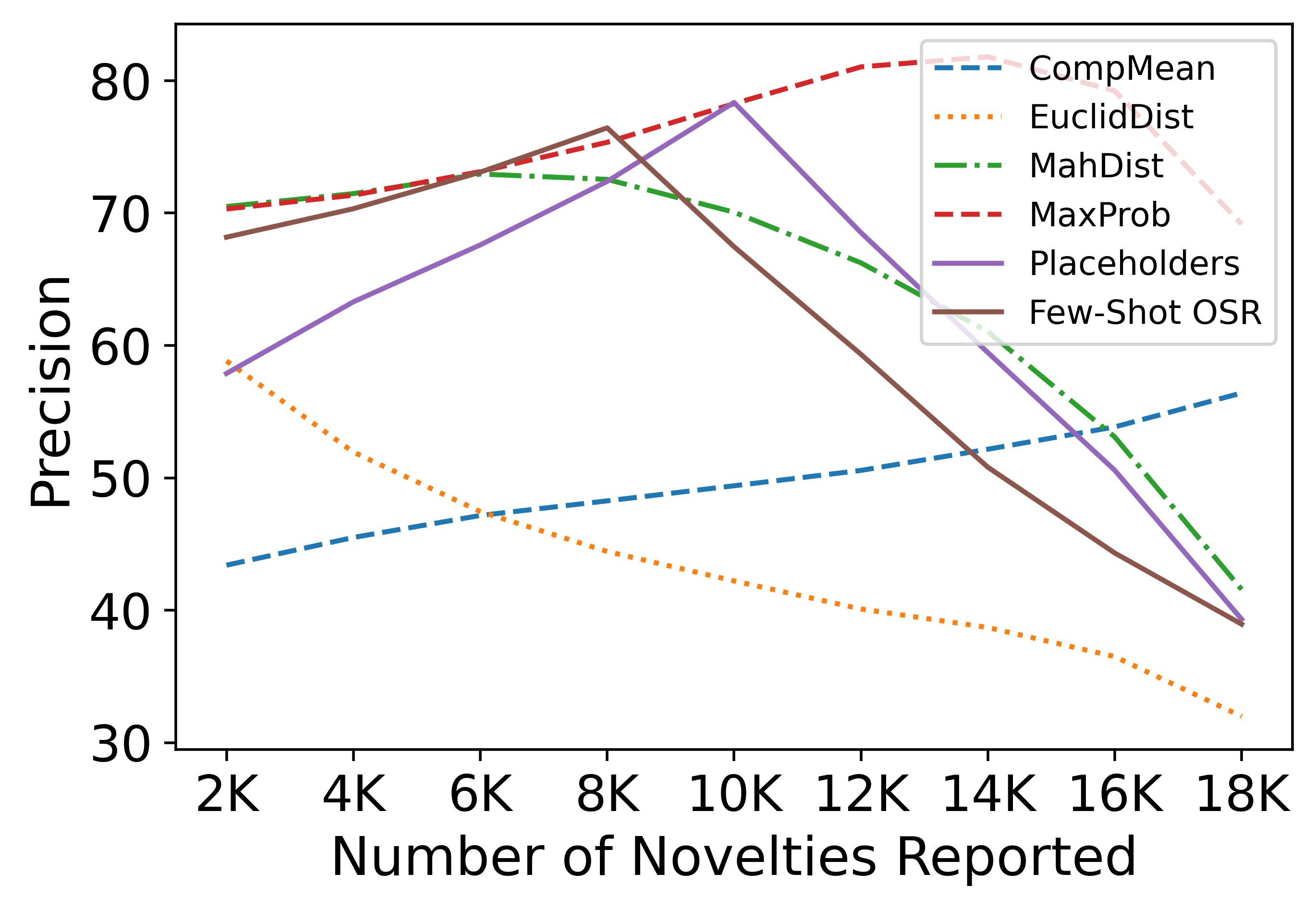}
    \end{subfigure}
    \begin{subfigure}{.32\linewidth}
         \includegraphics[width=\linewidth]{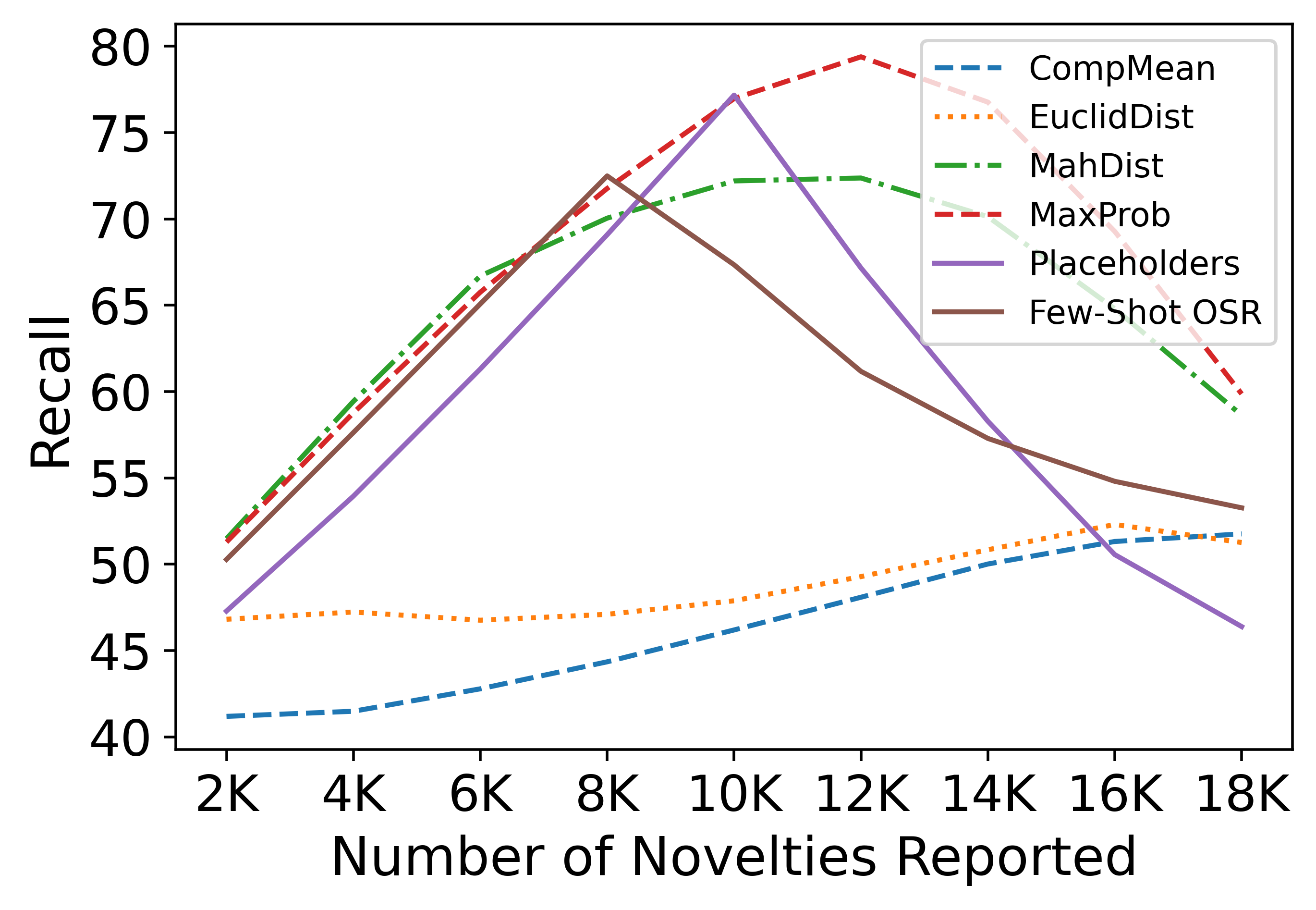}
    \end{subfigure}
    \begin{subfigure}{.32\linewidth}
        \includegraphics[width=\linewidth]{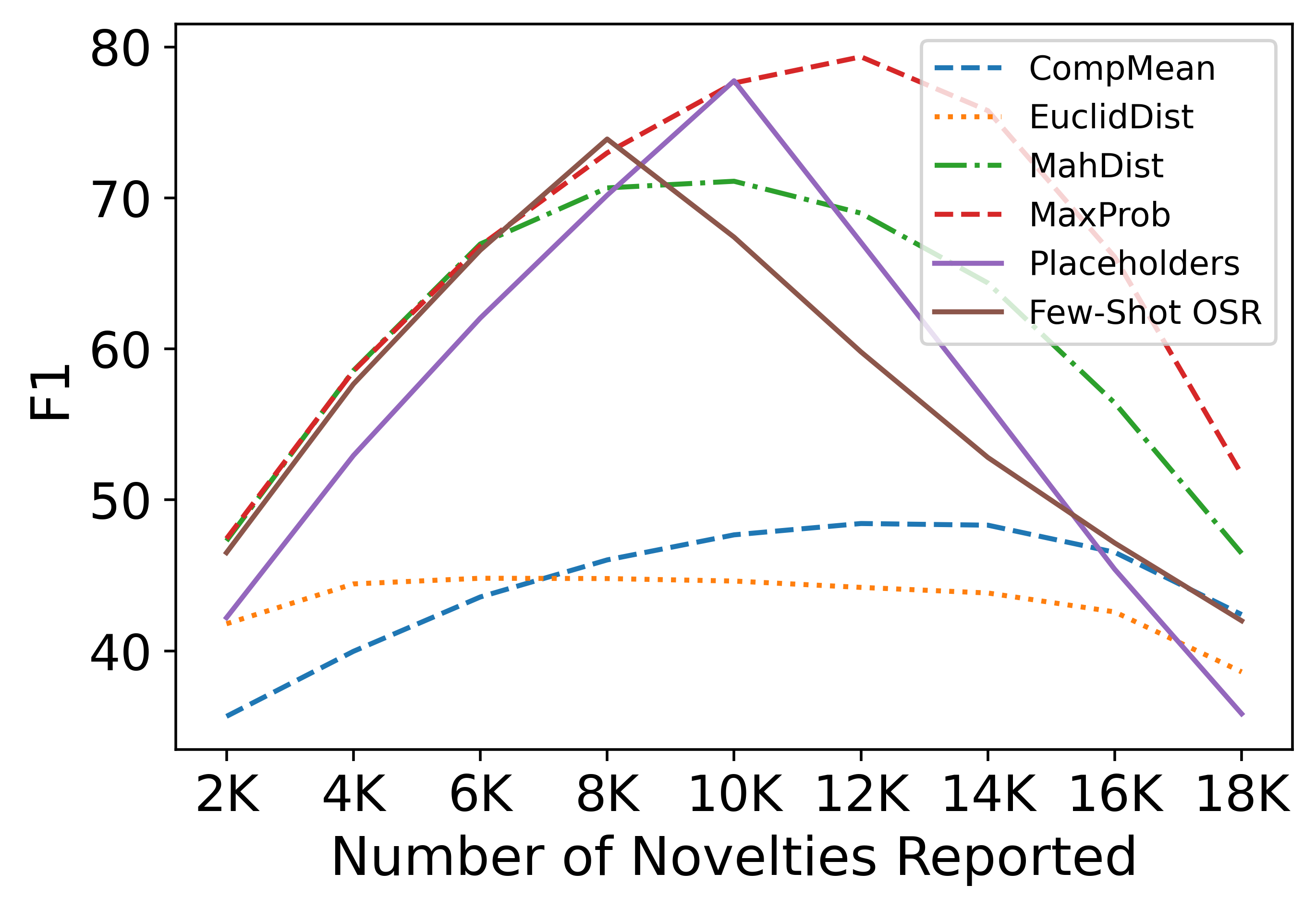}
    \end{subfigure}
    \caption{\textbf{Novelty Detection Performance on the base setting} -  \underline{Overall} Precision, Recall, and F1 achieved by various methods across the range of reported novelties on $Eval_{Det}$. 
    Specifically, each point on the curve represents the P, R, or F1 when its corresponding method reports the specified number of novelties (x-axis value) out of all instances in $Eval_{Det}$.
    We note that in the base setting, $Eval_{Det}$ has 20k instances out of which 10k are novel.
    }
    \label{fig:results_novelty_detection}    
\end{figure*}

Note that all the above measures are threshold dependent i.e. the system needs to select a confidence threshold (based on which it classifies instances on which it fails to surpass that threshold as novel) and its performance measures depend on that.
This is not a fair performance metric as its performance heavily depends on the number of novelties present in the evaluation dataset ($Eval_{Det}$).
To comprehensively evaluate a system, we use a threshold-independent performance metric in which we compute these precision, recall, and F1 values for a \textbf{range of reported novelties}.
To achieve this, we order the evaluation instances of $Eval_{Det}$ based on the system's prediction score (calculated using various techniques described in the next subsection) and take the least confident instances as reported novelties (for each number in the range of reported novelties).
Then, we plot a curve for these performance measures and aggregate the values (AUC) to calculate the overall performance of the method (refer to Figure \ref{fig:results_novelty_detection}).
This evaluation methodology (similar to the OOD detection method) makes the performance measurement comprehensive and also accounts for the number of novelties present in the evaluation dataset.

\paragraph{Novelty Accommodation:}
In this stage, the task for the system is to do $(K+N)$ class classification instead of $(K+1)$ class classification.
The system leverages the feedback ($D^F$) (which is contingent on the number of reported novelties) to adapt it for the task, and its performance also depends on that.
Following the methodology described for the detection stage, we evaluate the system's performance over a range of reported novelties and hence over a range of feedback. 
Specifically, we find the feedback dataset $D^F$ for a range of reported novelties and for each individual feedback, we incorporate it into the system and then evaluate its prediction performance on the $(K+N)$ classification task. 
Similar to the detection stage, we plot a curve (across a range of reported novelties) and calculate its area under the curve value to quantify the overall performance of novelty accommodation.

\subsection{Methods for Novelty Detection}

As described in the previous subsection, we calculate the system's performance on a range of reported novelties.
To achieve this, we order the evaluation instances of $Eval_{Det}$ based on the system's prediction confidence score (calculated using various techniques described in this subsection) and take the least confident instances as reported novelties (for each number in the range of reported novelties).
This implies that the performance depends on the system's method of computing this prediction score. 
We explore the following methods of computing this score for the evaluation instances:

\paragraph{Maximum Softmax Probability (MaxProb):}
Usually, the last layer of models has a softmax activation function that gives the probability distribution $P(y)$ over all possible answer candidates $Y$. 
For the classification tasks, $Y$ corresponds to the set of labels. 
\citet{hendrycks17baseline} introduced a simple method that uses the maximum softmax probability across all answer candidates as the confidence estimator i.e. the prediction confidence score corresponds to $\max_{y \in Y} P(y)$.
In this method, we order the evaluation instances of $Eval_{Det}$ based on this confidence measure, and for each value in the range of reported novelties, we report those instances as novel on which the model is least confident.
For the remaining instances, we output the label (out of $K$ classes) having the maximum softmax probability.




\paragraph{Euclidean Distance (EuclidDist)}: In this approach, we consider each sample as a point in K-dimensional space. For each sample, the probabilities from the K class classifier are chosen as coordinates in the space. We then calculate Euclidean distances between each sample and the entire distribution. The points furthest away from the distribution are classified as novel instances. 

The {\em Euclidean distance} is given by $d = \sqrt { \sum_{i=1}^{K} \left( {x_i - x_{mu} } \right)^2 }$ where $x_{mu}$ is the distribution of all the samples.


\paragraph{Mahalanobis Distance (MahDist):} This approach is similar to the previous approach with the only difference that Mahalanobis distance is used to compute the distance between the sample and the distribution.  

The {\em Mahalanobis distance} \cite{ghorbani2019mahalanobis} between $\boldsymbol{x}^i$ and $\boldsymbol{x}^j$ is given by $\Delta^2=(\boldsymbol{x}^i-\boldsymbol{x}^j)^\top\Sigma^{-1}(\boldsymbol{x}^i-\boldsymbol{x}^j)$, where $\Sigma$ is a $d\times d$ covariance matrix. 
$\Delta^2$ is equivalent to the squared Euclidean distance between $\boldsymbol{y}^i$ and $\boldsymbol{y}^j$, where $\boldsymbol{y}$ is a linearly transformed version of $\boldsymbol{x}$. 

\paragraph{Mean (CompMean):} For each sample, the mean of K-1 classes is computed. The class with the highest probability is left out. 
The mean is later subtracted from 1.
The resultant score for all the samples is sorted in descending order. 
The last Y elements are classified as Novel instances.

\paragraph{Learning Placeholders Algorithm (Placeholder):} \citet{9578797} propose a Placeholder algorithm for increasing the separation between clusters of samples in different classes. It addresses the challenge of open-set recognition by increasing the distance between class clusters and shrinking the classification boundary, allowing the classifier to classify samples as novel that fall outside these clusters. It demonstrates the effectiveness of the Placeholders algorithm through experiments and comparison with other state-of-the-art open-set recognition methods.

\paragraph{Few Shot Open set Recognition (Few Shot OSR):}
\citet{jeong2021few} presents a method for recognizing novel classes with few examples available for each class. 
It uses prototypes to represent each class and a similarity function to compare new examples to these prototypes, allowing for the effective recognition of novel classes. 
The paper includes experiments on multiple datasets and compares the method's performance to other state-of-the-art few-shot open-set recognition methods.

We further detail these methods in Appendix \ref{sec_methods}. We note that other OOD/anomaly detection methods can also be explored here. However, we study only a limited set of methods since the focus of this work is on formulating and exploring NoveltyTask. 

\subsection{Methods for Novelty Accommodation:}
\label{subsec_accommodation}

After the detection stage, the system gets feedback i.e. examples of novel labels ($D^F$).
We explore the following methods of leveraging this feedback:

\paragraph{Retrain using $D^T$ and $D^F$:}
$D_T$ consists of examples of known labels, and $D_F$ consists of examples of novel labels.
In this approach, we train a new model $(K+N)$ classifier by combining data instances of $D^T$ and $D^F$. 


\paragraph{Further Fine-tune using $D^F$:}
In this method, we first train a model on $D^T$ with extra dummy labels, i.e., we train a model having more than $K$ logits. This allows modifying the same model to learn to output the novel labels.
To incorporate the feedback, the model initially trained on $D^T$ with dummy labels is further fine-tuned using $D^F$. 

\paragraph{Further Fine-tune using $D^T$ (sampled) and $D^F$:}
Here, we follow the same strategy as the previous method, but instead of further fine-tuning only on $D^F$, we further fine-tuning using both $D^T$ (downsampled) and $D^F$.
This is done to reduce catastrophic forgetting \cite{carpenter1988art} of the known labels.

\section{Experiments and Results}
\label{sec_experiments}

\subsection{Experimental Setup}
\label{subsec_experimental_setup}
\paragraph{Configurations:} We use Amazon reviews \cite{mcauley2015image,he2016ups} for the authorship attribution task. 
In this task, each author corresponds to a class.
We compile a dataset consisting of $250k$ instances across $200$ authors and use it for NoveltyTask.
We define experimental settings using a set of configuration parameters; for the base setting, we use the following values:

\begin{itemize}[noitemsep,nosep,leftmargin=*]
    \item Number of Known Classes (K): 100
    \item Training Data $D^T$ Class Balanced: True
    \item \# Instances Per Known Label in $D^T$: 500
    \item Number of Novel Classes (N): 100
    \item \# Instances Per Class in $Eval_{Det}$: 100
    \item \# Instances Per Class in $Eval_{Acc}$: 500
\end{itemize}

In the above setting, the total number of evaluation instances in $Eval_{Det}$ is 20k out of which 10k are novel.
In this work, we also study other settings by varying the values of these parameters.


\paragraph{Models:} We run all our experiments using the BERT-base model \cite{devlin-etal-2019-bert}. 
For classification, we add a linear layer on top of BERT representation and train the model with a standard learning rate ranging in $\{1{-}5\}e{-}5$. 
All experiments are done with Nvidia V100 16GB GPUs.



\begin{figure*}[t]
\centering
    \begin{subfigure}{.32\linewidth}
        \includegraphics[width=\linewidth]{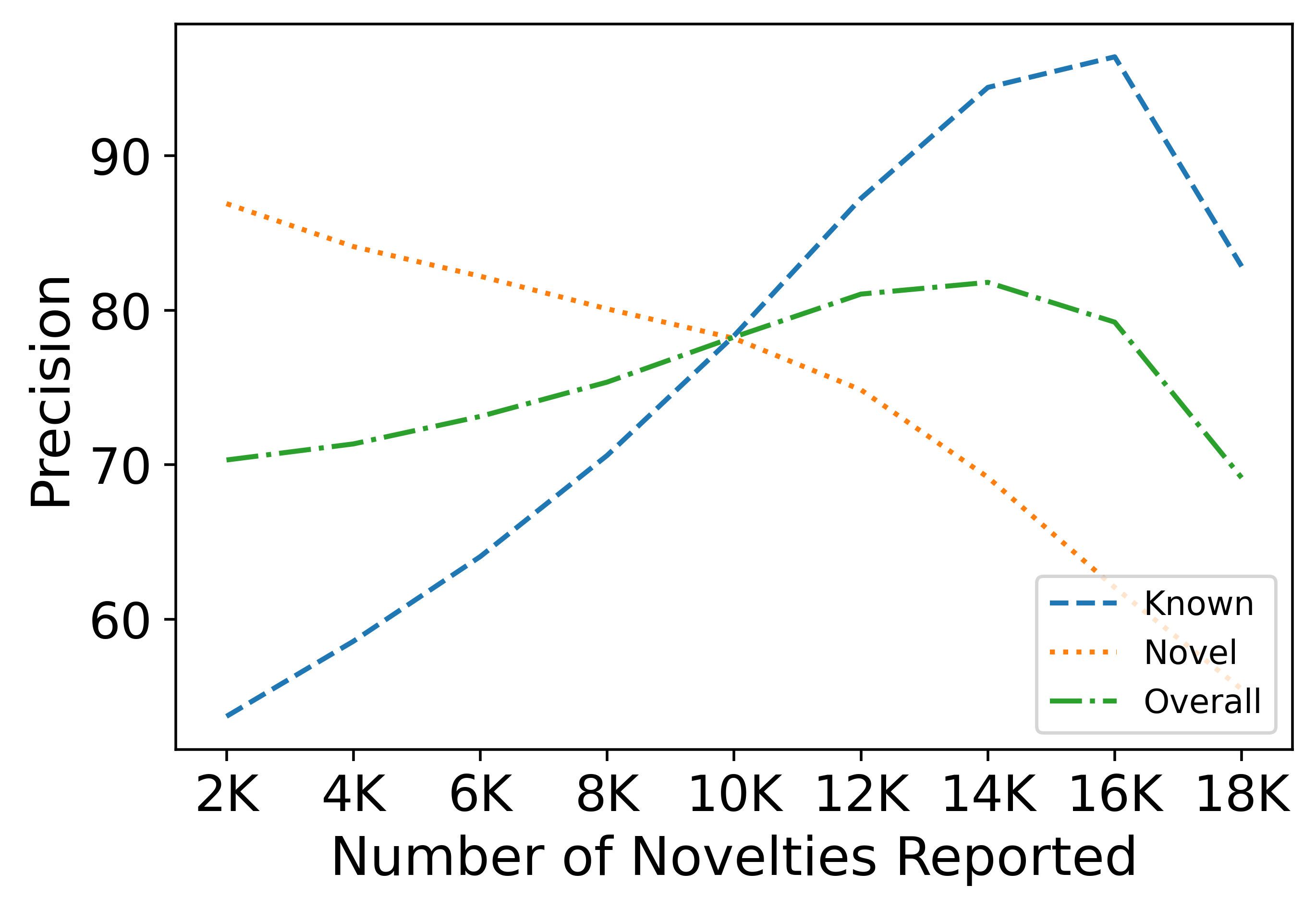}
    \end{subfigure}
    \begin{subfigure}{.32\linewidth}
         \includegraphics[width=\linewidth]{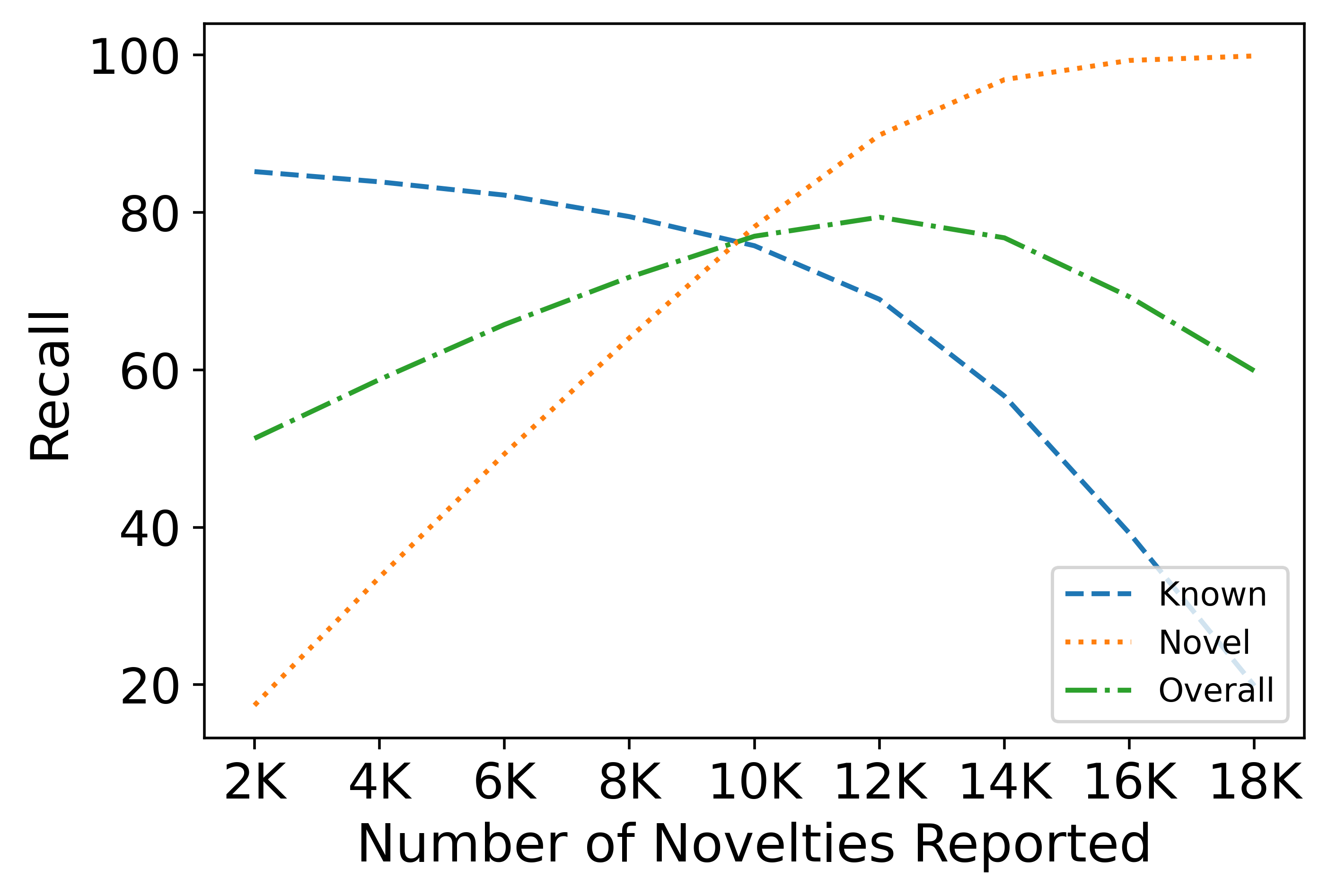}
    \end{subfigure}
    \begin{subfigure}{.32\linewidth}
        \includegraphics[width=\linewidth]{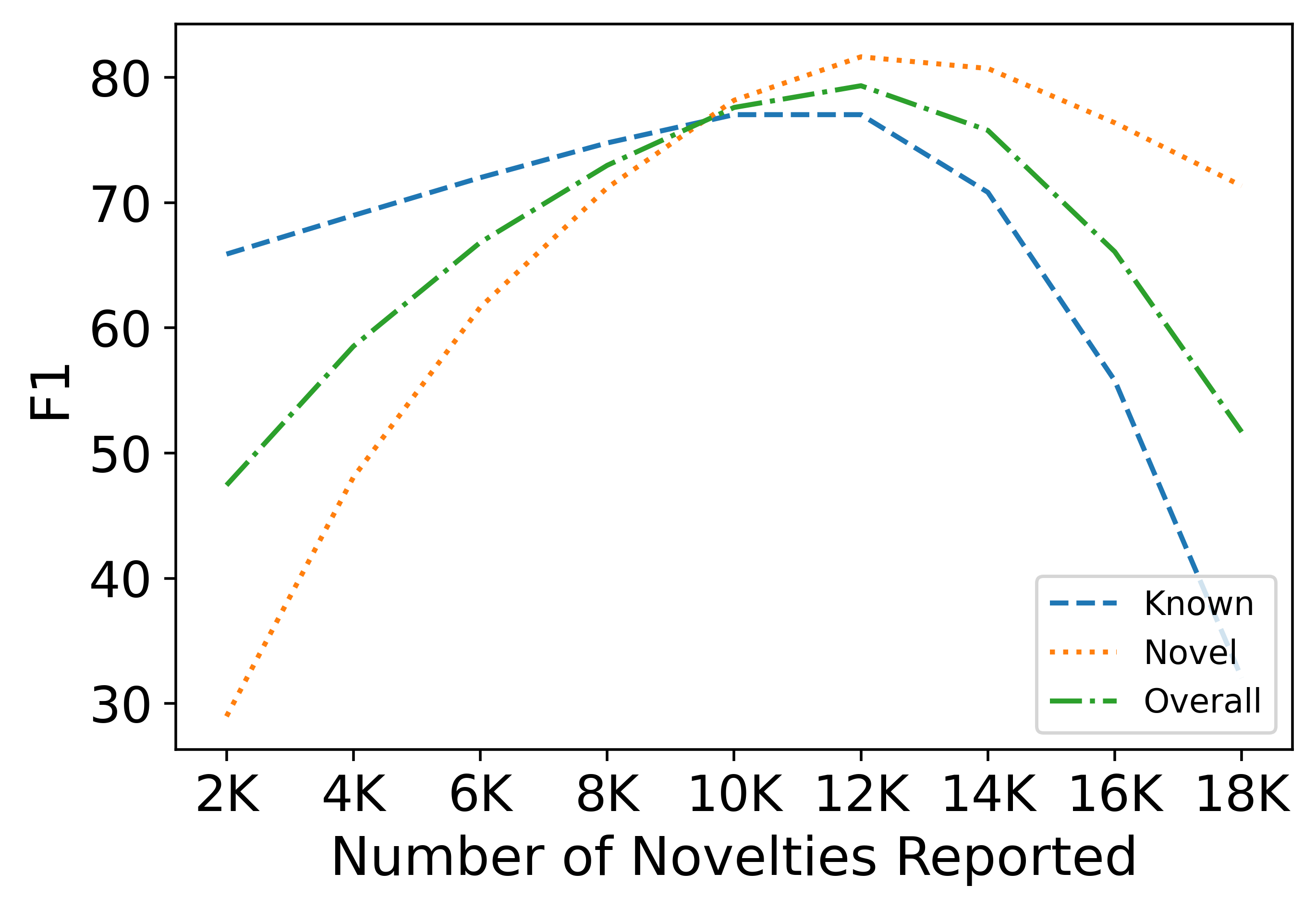}
    \end{subfigure}
    \caption{\textbf{MaxProb's Novelty Detection performance on the base setting} separately on Known classes, Novel classes, and Overall data. }
    \label{fig:results_novelty_detection_maxprob}    
\end{figure*}

\begin{figure*}[t]
\centering
    \begin{subfigure}{.32\linewidth}
        \includegraphics[width=\linewidth]{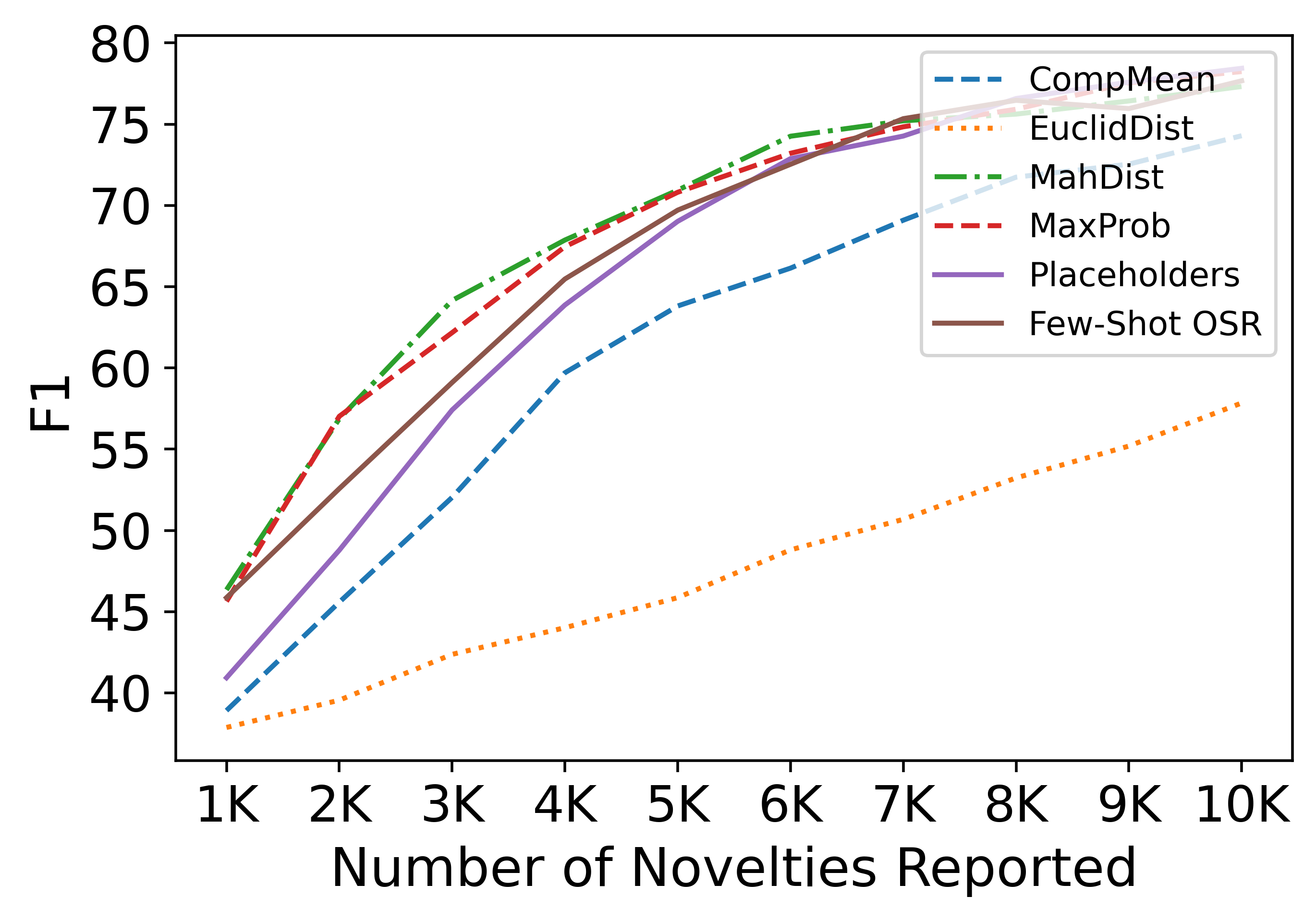}
        \caption{Retraining w/ $D^T$ + $D^F$}
    \end{subfigure}
    \begin{subfigure}{.32\linewidth}
         \includegraphics[width=\linewidth]{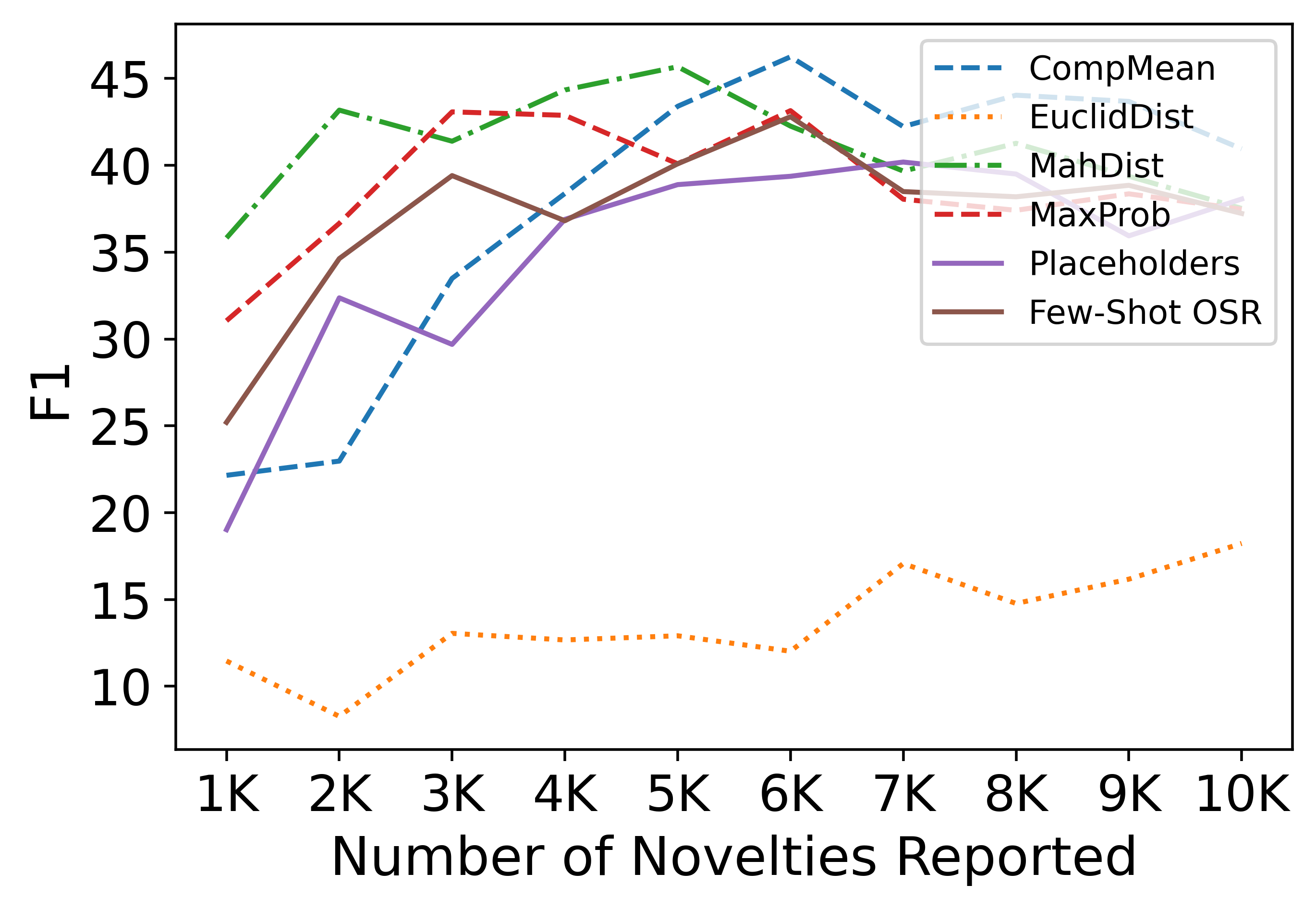}
         \caption{Further finetune w/ $D^F$}
    \end{subfigure}
    \begin{subfigure}{.32\linewidth}
        \includegraphics[width=\linewidth]{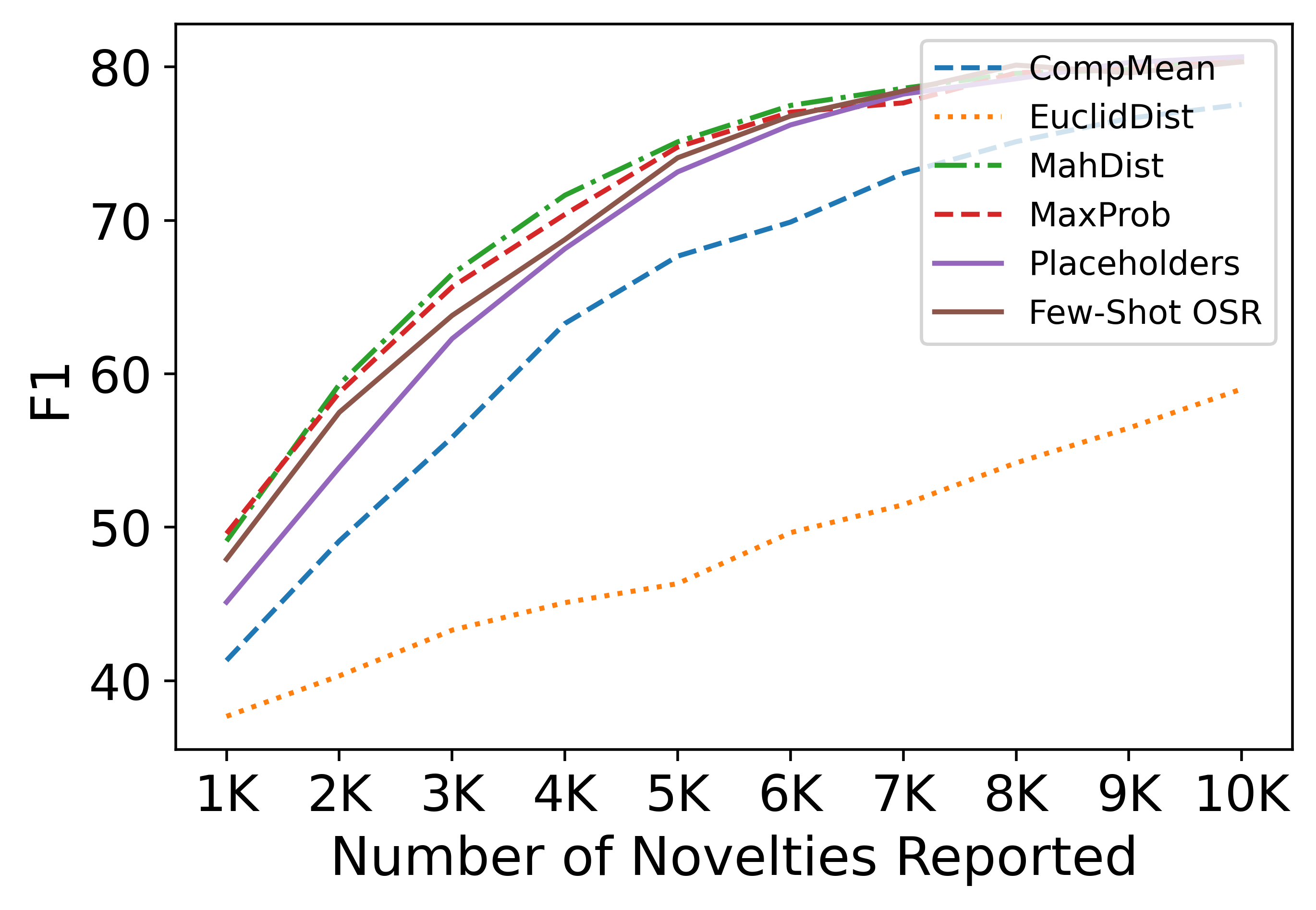}
        \caption{Further finetune w/ sampled $D^T$+$D^F$}
    \end{subfigure}
    \caption{\textbf{Novelty Accommodation performance on the base setting} - Overall F1 achieved by systems trained by leveraging the feedback (using different accommodation methods (a, b, and c)) resulting from different detection methods across the range of reported novelties. }
    \label{fig:results_novelty_accomodation_21}    
\end{figure*}

\subsection{Results}



\subsubsection{Novelty Detection}

Figure \ref{fig:results_novelty_detection} shows the novelty detection performance on the base setting ($Eval_{Det}$ has 20k instances out of which 10k are novel) i.e. overall Precision, Recall, and F1 achieved by various methods across the range of reported novelties on $Eval_{Det}$. 
Specifically, each point on the curve represents the P, R, or F1 when its corresponding method reports the specified number of novelties (x-axis value) out of all instances in $Eval_{Det}$.


\paragraph{MaxProb achieves the best overall performance:} From the plots, it can be observed  that MaxProb achieves the highest AUC value and hence the best overall performance. 
This result supports the prior finding that complex methods fail to consistently outperform the simiple MaxProb method \cite{varshney-etal-2022-investigating, azizmalayeri2022ood}.

\paragraph{Performance Analysis of MaxProb: }
To further study the performance of MaxProb in detail, we show its P, R, and F1 curves for Known, Novel, and Overall data in Figure \ref{fig:results_novelty_detection_maxprob}.
As expected, the precision on Known classes tends to increase as more novelties get reported. This is because the system predicts the known classes only for those instances on which it is most confident (highest MaxProb).
Similarly, the precision on novel instances tends to decrease as more and more novelties get reported.
The overall precision on the (K+1) classes tends to increase with the increase in the number of reported novelties.
We provide a detailed performance analysis on the known classes, novel classes, and overall data in Appendix \ref{sec_results_appendix}.



\subsubsection{Novelty Accommodation}
Figure \ref{fig:results_novelty_accomodation_21} shows the novelty accommodation performance on the base setting ($Eval_{Acc}$ has 100k instances uniformly split across 200 classes - 100 known and 100 novel) i.e. Overall F1 achieved by systems trained by leveraging the feedback (using different accommodation methods (a, b, and c)) resulting from different detection methods across the range of reported novelties. 
Note that for a value of reported novelty, each detection method results in a different feedback dataset and hence will have a different accommodation performance.
We show the Precision and Recall curves in the Appendix.

\paragraph{Retraining w/ $D^T$ + $D^F$: }
We note that MaxProb and MahDist turned out to be the best detection methods. This implies that their corresponding feedback dataset would contain more examples of the novel labels. This further reflects in the novelty accommodation performance as incorporating the feedback of these methods results in the best overall accommodation performance using the retraining method.

\begin{figure}[t]
\centering
        \includegraphics[width=\linewidth]{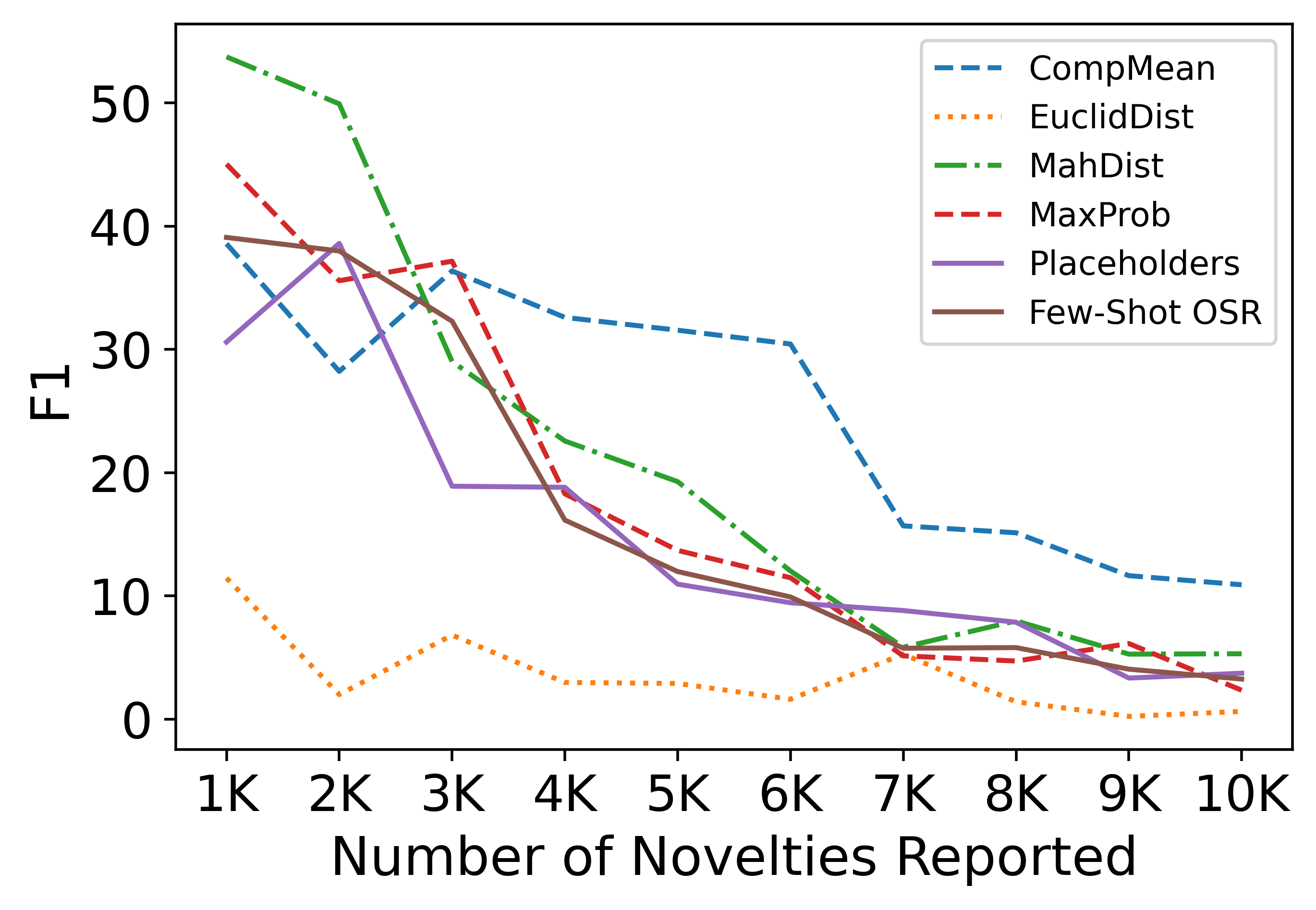}
    \caption{Demonstrating catastrophic forgetting on Known classes on further fine-tuning with $D^F$ only.}
    \label{fig:catastrophic_forgetting}    
\end{figure}

\paragraph{Catastrophic Forgetting Increases in further fine-tuning with $D^F$: }
As previously mentioned, this method leads to catastrophic forgetting of the known classes resulting in low overall F1 performance. 
We demonstrate this trend in Figure \ref{fig:catastrophic_forgetting}. 
Furthermore, with the increase in the number of novelties reported, the extent of catastrophic forgetting also increases.

\paragraph{Further fine-tuning with Sampled $D^T$ and $D^F$ improves performance: }
This method not only mitigates catastrophic forgetting but also results in a slight improvement over the retraining method.
For sampling, we use the maximum number of correctly detected instances of a class in $D^F$ as the threshold for sampling instances of known labels from $D^T$.
Furthermore, this method is more \textbf{training efficient} than the retraining method as the number of training instances is significantly lower in this method and yet it achieves better performance.

\begin{figure}[t]
\centering
        \includegraphics[width=0.8\linewidth]{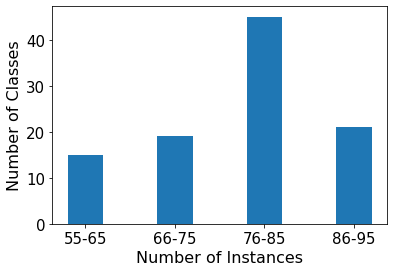}
    \caption{Distribution of instances over novel classes in $D^F$ when the number of reported novelties (by MaxProb) is 10k in the base setting.
    }
    \label{fig:histogram}    
\end{figure}

\begin{figure}[t]
\centering
        \includegraphics[width=0.9\linewidth]{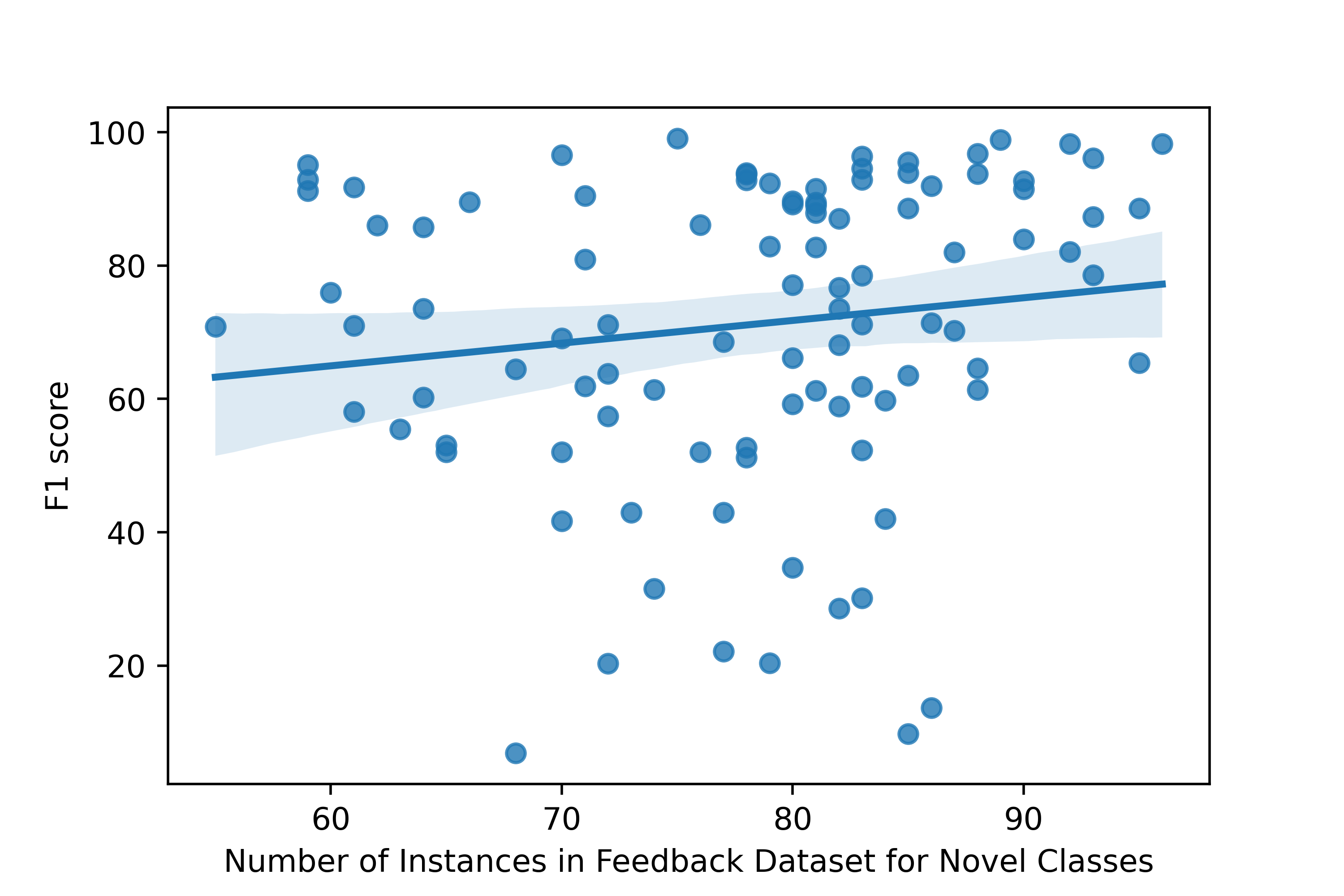}

    \caption{
    Scatter plot showing F1 performance achieved by each novel class in the accommodation stage vs the number of its instances in the feedback dataset.
    The plot is for the MaxProb detection method when 10k novelties are detected for Retrain using $D^T$ and $D^F$ accommodation method. 
    }
    \label{fig:scatter_plot}    
\end{figure}



\subsection{Analysis}

\paragraph{Distribution of Instances over classes in the Feedback dataset: }
We show the distribution of instances over all the classes (novel) in the feedback dataset $D^F$ when the number of reported novelties is 10k in Figure \ref{fig:histogram}.
It can be observed from the histogram that for all the novel classes, novel instances between 55 and 95 are correctly detected. For majority of the classes, 76-85 instances are detected. 
This further shows that the detection method is not biased towards or against any set of novel classes in identifying novel instances.

\paragraph{Trend of class level performance in the accommodation stage vs the number of instances in the feedback dataset:}
In Figure \ref{fig:scatter_plot}, we show a scatter plot of accommodation F1 performance achieved by each class vs the number of its instances in the feedback dataset. 
The plot is for the MaxProb detection method when 10k novelties are detected and retrain with $D^T$, and $D^F$ accommodation method is used. 
From the trend, it can be inferred that with the increase in the number of instances, the performance generally tends to increase.

\begin{figure}[t]
\centering
        \includegraphics[width=0.8\linewidth]{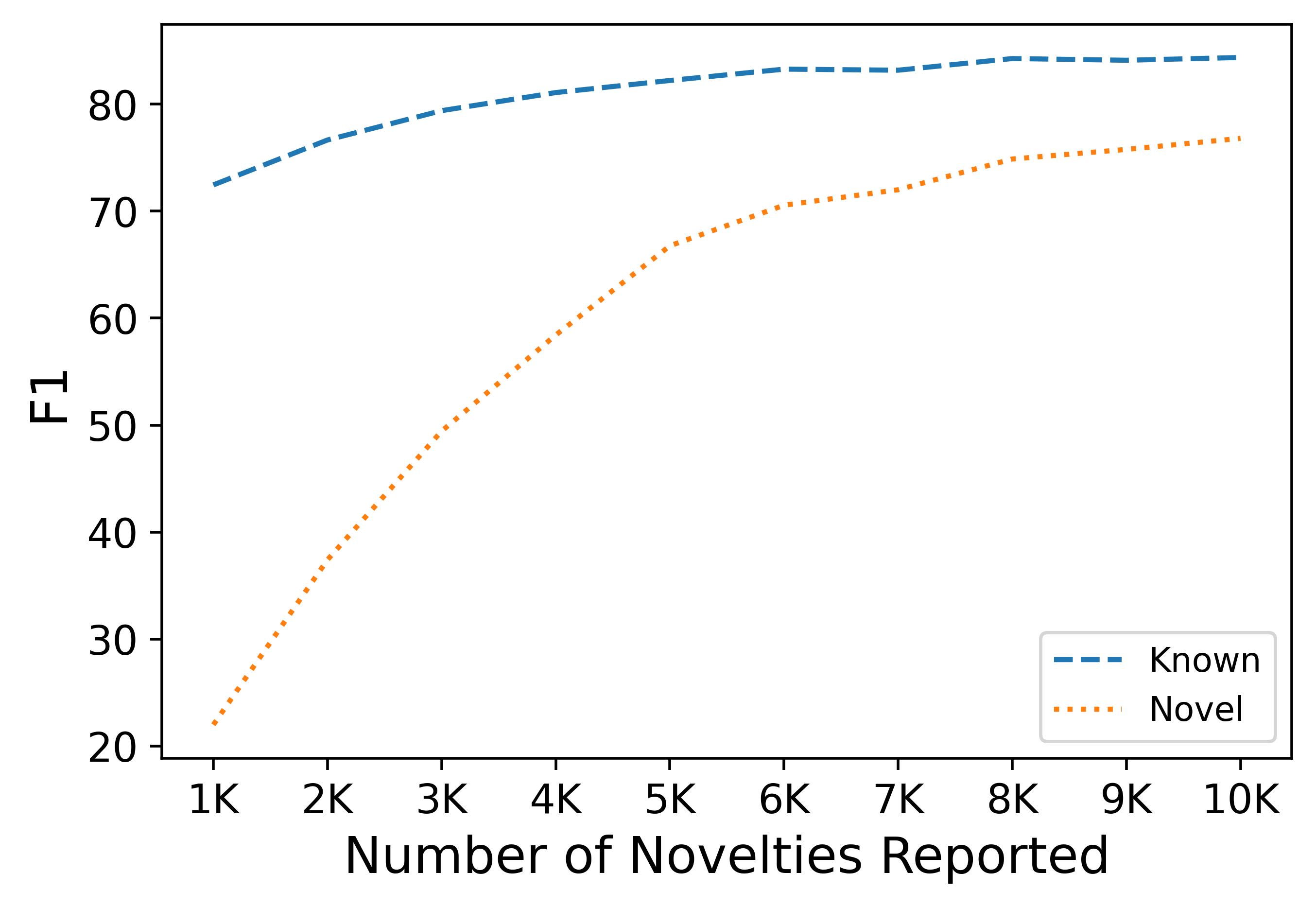}
    \caption{
    Comparing Performance on the system on Known and Novel classes in the accommodation stage.
    }
    \label{fig:known_vs_novel}    
\end{figure}

\paragraph{Comparing Performance of Known and Novel Classes in the Accommodation Stage:}
In Figure \ref{fig:known_vs_novel}, we compare the performance of the system (in the accommodation stage) on Known and Novel classes.
It clearly shows that the system finds it challenging to adapt itself to the novel classes.
This can be partly attributed to the availability of limited number of training examples of novel classes. 
This also provides opportunities for developing better accommodation techniques that can overcome this limitation.

\subsection{Other Configuration Settings} 
In this work, we also study NoveltyTask for different settings (different configuration parameters defined in \ref{subsec_experimental_setup}).
We observe findings and trends similar to the base setting. 
We provide detailed results and discussion in the Appendix.

\section{Conclusion and Discussion}

To initiate a systematic research in the important yet underexplored area of `dealing with novelties', we introduce \textit{NoveltyTask}, a multi-stage task to evaluate a system's performance on pipelined novelty `detection' and `accommodation' tasks.
We provided mathematical formulation of NoveltyTask and instantiated it with the authorship attribution task.
To this end, we also compiled a large dataset (consisting of $250k$ instances across $200$ authors/labels) from Amazon reviews corpus.
We conducted comprehensive experiments and explored several baseline methods for both detection and accommodation tasks.

Looking forward, we believe that our work opens up several avenues for interesting research avenues in this space, such as improving performance of detecting the novel instances and leveraging the feedback in a way that helps the system adapt with just a few examples of the novel labels.



\section*{Limitations}

Though the formulation of the task allows exploring several different settings (by varying the configuration parameters), in this work, we investigated only the label-balanced setting. 
Exploring the label-imbalanced setting is another very interesting research direction, and we leave that for future work. 
Another limitation was the limited exploration of novelty detection methods, as a number of methods have been proposed in the recent times. However, we study only a limited set of methods since the focus of this work is on formulating and exploring NoveltyTask. 
Lastly, we note that NoveltyTask is a controlled task/framework for evaluating a system's ability to deal with novelties and not a method to improve its ability.

\section*{Acknowledgement}
We thank the anonymous reviewers for their insightful feedback. This research was supported by DARPA SAIL-ON program.

\bibliography{anthology,custom}
\bibliographystyle{acl_natbib}

\section*{Appendix}
\appendix

\section{Other Related Work}

\subsection{Novelty/OOD/Anomaly Detection}

\subsubsection{Vision} Novelty/OOD/Anomaly detection is an active area of research in computer vision \cite{fort2021exploring,esmaeilpour2022zero,sun2021m2iosr,lu2022pmal,liu2020few,sun2020conditional,perera2020generative}. 
Datasets such as CIFAR 10 and 100 \cite{Krizhevsky09learningmultiple} are typically used to evaluate the efficacy of various detection methods. 

\citet{fort2021exploring} demonstrated that pre-training of transformer-based models using large datasets is fairly robust in detecting near-OOD instances using few examples.  
\citet{esmaeilpour2022zero} proposed to detect OOD instances using pairwise similarity score. They generate synthetic unseen examples and use their closed-set classifier to compute pairwise similarity. 
\citet{wei2021open} use open-set samples with dynamic, noisy labels and assign random labels to open-set examples, and use them for developing a system for OOD detection. 
\citet{sun2021react} analyze activation functions of the penultimate layer of pretrained models and rectify the activations to an upper limit for OOD detection.

\subsubsection{Language} 
\citet{zhou2021learning} propose to add an additional classifier in addition to a closed domain classifier for getting a class-specific threshold of known and unknown classes. They generate data placeholders to mimic open set categories. 
\citet{venkataram2018open} use an ensemble-based approach and replace the softmax layer with an OpenMAX layer.
The hypothesis is that the closest (most similar) class to any known class is an unknown one. This allows the classifier to be trained, enforcing the most probable class to be the ground truth class and the runner-up class to be the background class for all source data.

\citet{zhou-etal-2021-contrastive} employ a contrastive learning framework for unsupervised OOD detection, which is composed of a contrastive loss and an OOD scoring function. The contrastive loss increases the discrepancy of the representations of instances from different classes in the task, while the OOD scoring function maps the representations of instances to OOD detection scores. 

\citet{xu2019open} propose Learning to Accept Classes (L2AC) method based on meta-learning and does not require re-training the model when new classes are added. 
L2AC works by maintaining a set of seen classes and comparing new data points to the nearest example from each seen class. 

Detection approaches are also used in selective prediction \cite{varshney-etal-2022-investigating, kamath-etal-2020-selective, xin-etal-2021-art, varshney2023post} and cascading techniques \cite{varshney-baral-2022-model, varshney2022can, li-etal-2021-cascadebert-accelerating} where under-confident predictions are detected to avoid incorrect predictions.

\subsection{Authorship Attribution}
BERT (and its different variants like BertAA, RoBERTa) based, Siamese-based, and ensemble-based approaches have been used for authorship attribution. 
\citet{tyo2021siamese} propose an approach that uses a pretrained BERT model in a siamese configuration for audio-visual classification. 
They experiment with using triplet loss, contrastive loss, and a modified version of contrastive loss and compare the results.
\citet{bagdon2021profiling} combine the results of a n-gram-based logistic regression classifier with a transformer model based on RoBERTa \cite{Liu2019RoBERTaAR} via a SVM meta-classifier.
\citet{altakrori-etal-2021-topic-confusion} explore a new evaluation setting topic confusion task. 
The topic distribution is controlled by making it dependent on the author, switching the topic-author pairs between training and testing. This setup allows for measuring the degree to which certain features are influenced by the topic, as opposed to the author’s identity. 
Other works include \cite{10.1007/978-3-030-49161-1_22,fabien-etal-2020-bertaa,10.1007/978-3-030-85896-4_32}. N-grams, word embeddings, and other stylometric features have been used as input feature vectors for the task \cite{caballero2021explainable,boenninghoff2019explainable,li2022tasr,lagutina2021authorship,lagutina2020influence,lagutina2021survey}.

\section{Novelty Detection Algorithms}
\label{sec_methods}

\paragraph{Learning placeholders:} The Placeholders algorithm consists of two main components: "Learning Classifier Placeholders" and "Learning Data Placeholders". "Learning Classifier Placeholders" involves adding a set of weights called classifier placeholders to the linear classifier layer at the end of the network. This modified classifier function denoted as $f(x)=[WT(x),wT(x)]$, where w represents the weights of the additional k+1 class, is trained using a modified loss function that encourages the classifier to predict the k+1 class as the second most likely class for every sample. This loss function helps the classifier learn an embedding function such that the k+1 class is always the closest class to each class cluster boundary. In addition to the k+1 class, the Placeholders algorithm includes a tunable number (C) of additional classifiers to make decisions' boundaries smoother. The final classifier function is, therefore, $f(x)=[WT(x),maxk=1,...,C wkT(x)]$, meaning that the closest open-set region in the embedding space is taken into consideration.
"Learning Data Placeholders" involves tightening the decision boundaries around the known-class clusters in the embedding space through a process called manifold mixup. This involves creating "unknown" class data from known class data and using an additional loss function to penalize classifying this new data as any of the known classes. Manifold mixup works by interpolating the embeddings of two samples from closed-set classes to create an embedding for a new sample, which is considered to belong to an unknown class. After training the model using both classifier and data placeholders, the Placeholders algorithm includes a final calibration step in which an additional bias is added to the open-set logits. This bias is tuned using a validation set of closed-set samples such that 95\% of all closed-set samples are classified as known. The combination of these two components and final calibration allows the Placeholders algorithm to train a classifier to identify novel samples even when only trained on closed-set data.

\paragraph{Few Shot Open set Recognition using Meta-Learning}: In the paper "Few-Shot Open-Set Recognition using Meta-Learning", the authors propose a method for few-shot open-set recognition using meta-learning. The main idea is to train a meta-learner that can recognize new classes given a few examples of each class.

The meta-learner consists of a feature extractor network and a linear classifier. The feature extractor network is responsible for learning an embedding function that maps samples from different classes into a common embedding space. The goal is to learn an embedding function that clusters samples from the same class together while separating samples from different classes by a large margin.

To train the meta-learner, the authors use a meta-learning loss function that encourages the embedding function to learn a "smooth" embedding space. This loss function consists of two terms: a classification loss and a separation loss.

During training, the meta-learner is presented with a small number of examples from each new class and is required to classify these examples correctly. The meta-learner is trained to optimize the meta-learning loss function, which encourages the embedding function to learn a smooth embedding space where samples from different classes are well separated.

After training, the meta-learner can be used to classify new samples by first projecting them into the embedding space using the feature extractor network, and then using the linear classifier to assign them to the appropriate class. The final classifier is able to generalize to new classes not seen during training, as it has learned to recognize the underlying structure of the embedding space.





\section{Results}
\label{sec_results_appendix}
\paragraph{Hyperparameters of the model:} 
Hidden layer dropout probability of 0.15,
input Sequence length of 512 tokens,
batch size of 32, and standard learning rate ranging in $\{1{-}5\}e{-}5$.

\subsection{Novelty Accomodation Stage}

Table \ref{tab:ds1_s21}, \ref{tab:ds1_s22}, and \ref{tab:ds1_s23} show the results of all six novelty detection methods across all three accommodation settings on the first dataset whose results are described in details in the main paper.

Similar results are obtained for datasets 2 and 3 as well. 
Novelty Accommodation results for dataset 2 are present in table \ref{tab:ds2_s21}, \ref{tab:d2_s22}, and \ref{tab:ds2_s23} and in table \ref{tab:ds3_s21}, \ref{tab:ds3_s22}, and \ref{tab:ds3_s23} for dataset 3.

\begin{table*}[t!]
    \centering
    \small
    \resizebox{\linewidth}{!}
    {
        \begin{tabular}{llllllllll}
\hline
\multicolumn{1}{l|}{\textbf{\begin{tabular}[c]{@{}l@{}}\# of \\ Novelties\end{tabular}}} & \textbf{\begin{tabular}[c]{@{}l@{}}Known \\ Class\\ precision\end{tabular}} & \textbf{\begin{tabular}[c]{@{}l@{}}Known\\ class \\ recall\end{tabular}} & \multicolumn{1}{l|}{\textbf{\begin{tabular}[c]{@{}l@{}}Known \\ Class\\ F1\end{tabular}}} & \textbf{\begin{tabular}[c]{@{}l@{}}Novel\\ Class \\ Precision\end{tabular}} & \textbf{\begin{tabular}[c]{@{}l@{}}Novel\\ Class\\ Recall\end{tabular}} & \multicolumn{1}{l|}{\textbf{\begin{tabular}[c]{@{}l@{}}Novel \\ Class \\ F1\end{tabular}}} & \textbf{\begin{tabular}[c]{@{}l@{}}Overall\\ Precision\end{tabular}} & \textbf{\begin{tabular}[c]{@{}l@{}}Overall\\ Recall\end{tabular}} & \textbf{\begin{tabular}[c]{@{}l@{}}Overall\\ F1\end{tabular}} \\ \hline
\multicolumn{1}{l|}{\textbf{1000}}                                                       & 51.62                                                                       & 90.10                                                                    & \multicolumn{1}{l|}{65.64}                                                                & 15.62                                                                       & 2.23                                                                    & \multicolumn{1}{l|}{3.90}                                                                  & 33.62                                                                & 46.17                                                             & 38.91                                                         \\
\multicolumn{1}{l|}{\textbf{2000}}                                                       & 54.17                                                                       & 90.62                                                                    & \multicolumn{1}{l|}{67.81}                                                                & 30.21                                                                       & 8.51                                                                    & \multicolumn{1}{l|}{13.28}                                                                 & 42.19                                                                & 49.56                                                             & 45.58                                                         \\
\multicolumn{1}{l|}{\textbf{3000}}                                                       & 55.25                                                                       & 89.92                                                                    & \multicolumn{1}{l|}{68.44}                                                                & 50.29                                                                       & 12.69                                                                   & \multicolumn{1}{l|}{20.27}                                                                 & 52.77                                                                & 51.31                                                             & 52.03                                                         \\
\multicolumn{1}{l|}{\textbf{4000}}                                                       & 58.64                                                                       & 89.61                                                                    & \multicolumn{1}{l|}{70.89}                                                                & 67.22                                                                       & 23.97                                                                   & \multicolumn{1}{l|}{35.34}                                                                 & 62.93                                                                & 56.79                                                             & 59.70                                                         \\
\multicolumn{1}{l|}{\textbf{5000}}                                                       & 60.77                                                                       & 90.16                                                                    & \multicolumn{1}{l|}{72.60}                                                                & 74.02                                                                       & 30.99                                                                   & \multicolumn{1}{l|}{43.69}                                                                 & 67.40                                                                & 60.58                                                             & 63.81                                                         \\
\multicolumn{1}{l|}{\textbf{6000}}                                                       & 62.60                                                                       & 89.59                                                                    & \multicolumn{1}{l|}{73.70}                                                                & 76.85                                                                       & 36.23                                                                   & \multicolumn{1}{l|}{49.24}                                                                 & 69.73                                                                & 62.91                                                             & 66.14                                                         \\
\multicolumn{1}{l|}{\textbf{7000}}                                                       & 64.90                                                                       & 89.62                                                                    & \multicolumn{1}{l|}{75.28}                                                                & 80.26                                                                       & 42.21                                                                   & \multicolumn{1}{l|}{55.32}                                                                 & 72.58                                                                & 65.92                                                             & 69.09                                                         \\
\multicolumn{1}{l|}{\textbf{8000}}                                                       & 67.03                                                                       & 89.61                                                                    & \multicolumn{1}{l|}{76.69}                                                                & 82.63                                                                       & 48.16                                                                   & \multicolumn{1}{l|}{60.85}                                                                 & 74.83                                                                & 68.89                                                             & 71.74                                                         \\
\multicolumn{1}{l|}{\textbf{9000}}                                                       & 67.90                                                                       & 89.35                                                                    & \multicolumn{1}{l|}{77.16}                                                                & 82.83                                                                       & 50.46                                                                   & \multicolumn{1}{l|}{62.71}                                                                 & 75.37                                                                & 69.91                                                             & 72.54                                                         \\
\multicolumn{1}{l|}{\textbf{10000}}                                                      & 69.91                                                                       & 89.67                                                                    & \multicolumn{1}{l|}{78.57}                                                                & 83.81                                                                       & 54.10                                                                   & \multicolumn{1}{l|}{65.75}                                                                 & 76.86                                                                & 71.89                                                             & 74.29                                                         \\ \hline
\multicolumn{10}{c}{\textbf{Compute Mean}}                                                                                                                                                                                                                                                                                                                                                                                                                                                                                                                                                                                                                                                                                                                                                                    \\ \hline
\multicolumn{1}{l|}{\textbf{1000}}                                                       & 55.57                                                                       & 89.67                                                                    & \multicolumn{1}{l|}{68.62}                                                                & 6.41                                                                        & 7.71                                                                    & \multicolumn{1}{l|}{7.00}                                                                  & 30.99                                                                & 48.69                                                             & 37.87                                                         \\
\multicolumn{1}{l|}{\textbf{2000}}                                                       & 55.86                                                                       & 90.10                                                                    & \multicolumn{1}{l|}{68.96}                                                                & 9.09                                                                        & 10.98                                                                   & \multicolumn{1}{l|}{9.95}                                                                  & 32.48                                                                & 50.54                                                             & 39.55                                                         \\
\multicolumn{1}{l|}{\textbf{3000}}                                                       & 57.29                                                                       & 89.92                                                                    & \multicolumn{1}{l|}{69.99}                                                                & 13.54                                                                       & 15.52                                                                   & \multicolumn{1}{l|}{14.46}                                                                 & 35.42                                                                & 52.72                                                             & 42.37                                                         \\
\multicolumn{1}{l|}{\textbf{4000}}                                                       & 58.63                                                                       & 89.40                                                                    & \multicolumn{1}{l|}{70.82}                                                                & 15.88                                                                       & 18.07                                                                   & \multicolumn{1}{l|}{16.90}                                                                 & 37.26                                                                & 53.74                                                             & 44.01                                                         \\
\multicolumn{1}{l|}{\textbf{5000}}                                                       & 59.39                                                                       & 89.72                                                                    & \multicolumn{1}{l|}{71.47}                                                                & 19.00                                                                       & 20.82                                                                   & \multicolumn{1}{l|}{19.87}                                                                 & 39.19                                                                & 55.27                                                             & 45.86                                                         \\
\multicolumn{1}{l|}{\textbf{6000}}                                                       & 61.03                                                                       & 89.34                                                                    & \multicolumn{1}{l|}{72.52}                                                                & 23.67                                                                       & 25.77                                                                   & \multicolumn{1}{l|}{24.68}                                                                 & 42.35                                                                & 57.56                                                             & 48.80                                                         \\
\multicolumn{1}{l|}{\textbf{7000}}                                                       & 61.67                                                                       & 89.63                                                                    & \multicolumn{1}{l|}{73.07}                                                                & 27.36                                                                       & 28.05                                                                   & \multicolumn{1}{l|}{27.70}                                                                 & 44.51                                                                & 58.84                                                             & 50.68                                                         \\
\multicolumn{1}{l|}{\textbf{8000}}                                                       & 63.27                                                                       & 89.30                                                                    & \multicolumn{1}{l|}{74.06}                                                                & 31.58                                                                       & 32.01                                                                   & \multicolumn{1}{l|}{31.79}                                                                 & 47.42                                                                & 60.66                                                             & 53.23                                                         \\
\multicolumn{1}{l|}{\textbf{9000}}                                                       & 64.12                                                                       & 89.15                                                                    & \multicolumn{1}{l|}{74.59}                                                                & 35.75                                                                       & 34.22                                                                   & \multicolumn{1}{l|}{34.97}                                                                 & 49.93                                                                & 61.69                                                             & 55.19                                                         \\
\multicolumn{1}{l|}{\textbf{10000}}                                                      & 65.70                                                                       & 89.18                                                                    & \multicolumn{1}{l|}{75.66}                                                                & 40.30                                                                       & 38.12                                                                   & \multicolumn{1}{l|}{39.18}                                                                 & 53.00                                                                & 63.65                                                             & 57.84                                                         \\ \hline
\multicolumn{10}{c}{\textbf{Compute Euclid Distance}}                                                                                                                                                                                                                                                                                                                                                                                                                                                                                                                                                                                                                                                                                                                                                         \\ \hline
\multicolumn{1}{l|}{\textbf{1000}}                                                       & 54.84                                                                       & 90.10                                                                    & \multicolumn{1}{l|}{68.18}                                                                & 30.39                                                                       & 11.45                                                                   & \multicolumn{1}{l|}{16.63}                                                                 & 42.61                                                                & 50.78                                                             & 46.34                                                         \\
\multicolumn{1}{l|}{\textbf{2000}}                                                       & 59.14                                                                       & 90.22                                                                    & \multicolumn{1}{l|}{71.45}                                                                & 56.80                                                                       & 21.40                                                                   & \multicolumn{1}{l|}{31.09}                                                                 & 57.97                                                                & 55.81                                                             & 56.87                                                         \\
\multicolumn{1}{l|}{\textbf{3000}}                                                       & 62.09                                                                       & 90.02                                                                    & \multicolumn{1}{l|}{73.49}                                                                & 73.87                                                                       & 31.43                                                                   & \multicolumn{1}{l|}{44.10}                                                                 & 67.98                                                                & 60.72                                                             & 64.15                                                         \\
\multicolumn{1}{l|}{\textbf{4000}}                                                       & 64.82                                                                       & 89.66                                                                    & \multicolumn{1}{l|}{75.24}                                                                & 78.40                                                                       & 39.37                                                                   & \multicolumn{1}{l|}{52.42}                                                                 & 71.61                                                                & 64.51                                                             & 67.87                                                         \\
\multicolumn{1}{l|}{\textbf{5000}}                                                       & 67.45                                                                       & 89.29                                                                    & \multicolumn{1}{l|}{76.85}                                                                & 80.12                                                                       & 47.33                                                                   & \multicolumn{1}{l|}{59.51}                                                                 & 73.78                                                                & 68.31                                                             & 70.94                                                         \\
\multicolumn{1}{l|}{\textbf{6000}}                                                       & 69.63                                                                       & 89.93                                                                    & \multicolumn{1}{l|}{78.49}                                                                & 83.73                                                                       & 54.05                                                                   & \multicolumn{1}{l|}{65.69}                                                                 & 76.68                                                                & 71.99                                                             & 74.26                                                         \\
\multicolumn{1}{l|}{\textbf{7000}}                                                       & 71.18                                                                       & 89.45                                                                    & \multicolumn{1}{l|}{79.28}                                                                & 83.37                                                                       & 57.08                                                                   & \multicolumn{1}{l|}{67.76}                                                                 & 77.27                                                                & 73.26                                                             & 75.21                                                         \\
\multicolumn{1}{l|}{\textbf{8000}}                                                       & 71.48                                                                       & 89.20                                                                    & \multicolumn{1}{l|}{79.36}                                                                & 83.93                                                                       & 58.09                                                                   & \multicolumn{1}{l|}{68.66}                                                                 & 77.71                                                                & 73.64                                                             & 75.62                                                         \\
\multicolumn{1}{l|}{\textbf{9000}}                                                       & 72.10                                                                       & 89.09                                                                    & \multicolumn{1}{l|}{79.70}                                                                & 84.41                                                                       & 60.30                                                                   & \multicolumn{1}{l|}{70.35}                                                                 & 78.26                                                                & 74.69                                                             & 76.43                                                         \\
\multicolumn{1}{l|}{\textbf{10000}}                                                      & 73.45                                                                       & 88.92                                                                    & \multicolumn{1}{l|}{80.45}                                                                & 85.03                                                                       & 62.08                                                                   & \multicolumn{1}{l|}{71.76}                                                                 & 79.24                                                                & 75.50                                                             & 77.32                                                         \\ \hline
\multicolumn{10}{c}{\textbf{Compute Mahalanobis Distance}}                                                                                                                                                                                                                                                                                                                                                                                                                                                                                                                                                                                                                                                                                                                                                    \\ \hline
\multicolumn{1}{l|}{\textbf{1000}}                                                       & 54.72                                                                       & 89.65                                                                    & \multicolumn{1}{l|}{67.96}                                                                & 29.72                                                                       & 9.65                                                                    & \multicolumn{1}{l|}{14.57}                                                                 & 42.22                                                                & 49.65                                                             & 45.63                                                         \\
\multicolumn{1}{l|}{\textbf{2000}}                                                       & 59.75                                                                       & 90.12                                                                    & \multicolumn{1}{l|}{71.86}                                                                & 55.68                                                                       & 22.54                                                                   & \multicolumn{1}{l|}{32.09}                                                                 & 57.72                                                                & 56.33                                                             & 57.02                                                         \\
\multicolumn{1}{l|}{\textbf{3000}}                                                       & 61.80                                                                       & 89.69                                                                    & \multicolumn{1}{l|}{73.18}                                                                & 66.02                                                                       & 31.36                                                                   & \multicolumn{1}{l|}{42.52}                                                                 & 63.91                                                                & 60.53                                                             & 62.17                                                         \\
\multicolumn{1}{l|}{\textbf{4000}}                                                       & 64.73                                                                       & 90.20                                                                    & \multicolumn{1}{l|}{75.37}                                                                & 75.62                                                                       & 39.72                                                                   & \multicolumn{1}{l|}{52.08}                                                                 & 70.17                                                                & 64.96                                                             & 67.46                                                         \\
\multicolumn{1}{l|}{\textbf{5000}}                                                       & 67.11                                                                       & 89.53                                                                    & \multicolumn{1}{l|}{76.72}                                                                & 80.63                                                                       & 46.47                                                                   & \multicolumn{1}{l|}{58.96}                                                                 & 73.87                                                                & 68.00                                                             & 70.81                                                         \\
\multicolumn{1}{l|}{\textbf{6000}}                                                       & 69.20                                                                       & 89.61                                                                    & \multicolumn{1}{l|}{78.09}                                                                & 82.91                                                                       & 51.50                                                                   & \multicolumn{1}{l|}{63.53}                                                                 & 76.06                                                                & 70.56                                                             & 73.21                                                         \\
\multicolumn{1}{l|}{\textbf{7000}}                                                       & 70.90                                                                       & 89.39                                                                    & \multicolumn{1}{l|}{79.08}                                                                & 83.39                                                                       & 55.98                                                                   & \multicolumn{1}{l|}{66.99}                                                                 & 77.14                                                                & 72.68                                                             & 74.84                                                         \\
\multicolumn{1}{l|}{\textbf{8000}}                                                       & 71.13                                                                       & 89.20                                                                    & \multicolumn{1}{l|}{79.15}                                                                & 84.92                                                                       & 58.64                                                                   & \multicolumn{1}{l|}{69.37}                                                                 & 78.03                                                                & 73.92                                                             & 75.92                                                         \\
\multicolumn{1}{l|}{\textbf{9000}}                                                       & 73.21                                                                       & 89.59                                                                    & \multicolumn{1}{l|}{80.58}                                                                & 85.68                                                                       & 61.90                                                                   & \multicolumn{1}{l|}{71.87}                                                                 & 79.44                                                                & 75.75                                                             & 77.55                                                         \\
\multicolumn{1}{l|}{\textbf{10000}}                                                      & 73.92                                                                       & 89.11                                                                    & \multicolumn{1}{l|}{80.81}                                                                & 86.08                                                                       & 64.00                                                                   & \multicolumn{1}{l|}{73.42}                                                                 & 80.00                                                                & 76.55                                                             & 78.24                                                         \\ \hline
\multicolumn{10}{c}{\textbf{Compute Max Probability}}                                                                                                                                                                                                                                                                                                                                                                                                                                                                                                                                                                                                                                                                                                                                                         \\ \hline
\multicolumn{1}{l|}{\textbf{1000}}                                                       & 53.87                                                                       & 89.72                                                                    & \multicolumn{1}{l|}{67.32}                                                                & 17.00                                                                       & 7.15                                                                    & \multicolumn{1}{l|}{10.07}                                                                 & 35.44                                                                & 48.43                                                             & 40.93                                                         \\
\multicolumn{1}{l|}{\textbf{2000}}                                                       & 56.30                                                                       & 90.26                                                                    & \multicolumn{1}{l|}{69.35}                                                                & 33.36                                                                       & 16.72                                                                   & \multicolumn{1}{l|}{22.28}                                                                 & 44.83                                                                & 53.49                                                             & 48.78                                                         \\
\multicolumn{1}{l|}{\textbf{3000}}                                                       & 59.59                                                                       & 89.59                                                                    & \multicolumn{1}{l|}{71.57}                                                                & 56.36                                                                       & 24.13                                                                   & \multicolumn{1}{l|}{33.79}                                                                 & 57.97                                                                & 56.86                                                             & 57.41                                                         \\
\multicolumn{1}{l|}{\textbf{4000}}                                                       & 62.23                                                                       & 89.46                                                                    & \multicolumn{1}{l|}{73.40}                                                                & 69.30                                                                       & 34.70                                                                   & \multicolumn{1}{l|}{46.24}                                                                 & 65.77                                                                & 62.08                                                             & 63.87                                                         \\
\multicolumn{1}{l|}{\textbf{5000}}                                                       & 66.08                                                                       & 89.81                                                                    & \multicolumn{1}{l|}{76.14}                                                                & 76.18                                                                       & 44.24                                                                   & \multicolumn{1}{l|}{55.97}                                                                 & 71.13                                                                & 67.03                                                             & 69.02                                                         \\
\multicolumn{1}{l|}{\textbf{6000}}                                                       & 68.11                                                                       & 89.66                                                                    & \multicolumn{1}{l|}{77.41}                                                                & 84.26                                                                       & 50.07                                                                   & \multicolumn{1}{l|}{62.81}                                                                 & 76.18                                                                & 69.86                                                             & 72.88                                                         \\
\multicolumn{1}{l|}{\textbf{7000}}                                                       & 69.64                                                                       & 89.52                                                                    & \multicolumn{1}{l|}{78.34}                                                                & 83.82                                                                       & 54.43                                                                   & \multicolumn{1}{l|}{66.00}                                                                 & 76.73                                                                & 71.97                                                             & 74.27                                                         \\
\multicolumn{1}{l|}{\textbf{8000}}                                                       & 72.31                                                                       & 89.36                                                                    & \multicolumn{1}{l|}{79.94}                                                                & 85.28                                                                       & 59.61                                                                   & \multicolumn{1}{l|}{70.17}                                                                 & 78.80                                                                & 74.48                                                             & 76.58                                                         \\
\multicolumn{1}{l|}{\textbf{9000}}                                                       & 73.03                                                                       & 89.30                                                                    & \multicolumn{1}{l|}{80.35}                                                                & 85.77                                                                       & 62.45                                                                   & \multicolumn{1}{l|}{72.28}                                                                 & 79.40                                                                & 75.87                                                             & 77.59                                                         \\
\multicolumn{1}{l|}{\textbf{10000}}                                                      & 74.69                                                                       & 88.94                                                                    & \multicolumn{1}{l|}{81.19}                                                                & 85.24                                                                       & 65.03                                                                   & \multicolumn{1}{l|}{73.78}                                                                 & 79.96                                                                & 76.98                                                             & 78.44                                                         \\ \hline
\multicolumn{10}{c}{\textbf{Placeholders Algorithm}}                                                                                                                                                                                                                                                                                                                                                                                                                                                                                                                                                                                                                                                                                                                                                          \\ \hline
\multicolumn{1}{l|}{\textbf{1000}}                                                       & 54.45                                                                       & 90.31                                                                    & \multicolumn{1}{l|}{67.94}                                                                & 29.47                                                                       & 10.84                                                                   & \multicolumn{1}{l|}{15.85}                                                                 & 41.96                                                                & 50.57                                                             & 45.86                                                         \\
\multicolumn{1}{l|}{\textbf{2000}}                                                       & 57.17                                                                       & 90.10                                                                    & \multicolumn{1}{l|}{69.95}                                                                & 44.44                                                                       & 18.82                                                                   & \multicolumn{1}{l|}{26.44}                                                                 & 50.80                                                                & 54.46                                                             & 52.57                                                         \\
\multicolumn{1}{l|}{\textbf{3000}}                                                       & 61.10                                                                       & 89.96                                                                    & \multicolumn{1}{l|}{72.77}                                                                & 55.81                                                                       & 29.59                                                                   & \multicolumn{1}{l|}{38.67}                                                                 & 58.45                                                                & 59.77                                                             & 59.10                                                         \\
\multicolumn{1}{l|}{\textbf{4000}}                                                       & 62.74                                                                       & 89.85                                                                    & \multicolumn{1}{l|}{73.89}                                                                & 73.03                                                                       & 36.60                                                                   & \multicolumn{1}{l|}{48.76}                                                                 & 67.88                                                                & 63.23                                                             & 65.47                                                         \\
\multicolumn{1}{l|}{\textbf{5000}}                                                       & 65.90                                                                       & 89.69                                                                    & \multicolumn{1}{l|}{75.98}                                                                & 79.94                                                                       & 43.86                                                                   & \multicolumn{1}{l|}{56.64}                                                                 & 72.92                                                                & 66.77                                                             & 69.71                                                         \\
\multicolumn{1}{l|}{\textbf{6000}}                                                       & 67.59                                                                       & 89.34                                                                    & \multicolumn{1}{l|}{76.96}                                                                & 83.69                                                                       & 49.98                                                                   & \multicolumn{1}{l|}{62.58}                                                                 & 75.64                                                                & 69.66                                                             & 72.53                                                         \\
\multicolumn{1}{l|}{\textbf{7000}}                                                       & 70.52                                                                       & 89.73                                                                    & \multicolumn{1}{l|}{78.97}                                                                & 84.42                                                                       & 56.92                                                                   & \multicolumn{1}{l|}{67.99}                                                                 & 77.47                                                                & 73.33                                                             & 75.34                                                         \\
\multicolumn{1}{l|}{\textbf{8000}}                                                       & 71.90                                                                       & 89.64                                                                    & \multicolumn{1}{l|}{79.80}                                                                & 85.32                                                                       & 59.24                                                                   & \multicolumn{1}{l|}{69.93}                                                                 & 78.61                                                                & 74.44                                                             & 76.47                                                         \\
\multicolumn{1}{l|}{\textbf{9000}}                                                       & 72.34                                                                       & 89.04                                                                    & \multicolumn{1}{l|}{79.83}                                                                & 83.72                                                                       & 58.95                                                                   & \multicolumn{1}{l|}{69.18}                                                                 & 78.03                                                                & 73.99                                                             & 75.96                                                         \\
\multicolumn{1}{l|}{\textbf{10000}}                                                      & 73.41                                                                       & 89.71                                                                    & \multicolumn{1}{l|}{80.75}                                                                & 85.84                                                                       & 61.92                                                                   & \multicolumn{1}{l|}{71.94}                                                                 & 79.63                                                                & 75.81                                                             & 77.67                                                         \\ \hline
\multicolumn{10}{c}{\textbf{Few shot Open set Recognition}}                                                                                                                                                                                                                                                                                                                                                                                                                                                                                                                                                                                                                                                                                                                                                   \\ \hline
\end{tabular}
    }
    \caption{
   Novelty Accommodation Stage: Dataset 1: Retrain using $D^T$ and $D^F$
    }
    \label{tab:ds1_s21}
\end{table*}
\begin{table*}[t!]
    \centering
    \small
    \resizebox{\linewidth}{!}
    {
       \begin{tabular}{llllllllll}
\hline
\multicolumn{1}{l|}{\textbf{\begin{tabular}[c]{@{}l@{}}\# of \\ Novelties\end{tabular}}} & \textbf{\begin{tabular}[c]{@{}l@{}}Known \\ Class\\ precision\end{tabular}} & \textbf{\begin{tabular}[c]{@{}l@{}}Known\\ class \\ recall\end{tabular}} & \multicolumn{1}{l|}{\textbf{\begin{tabular}[c]{@{}l@{}}Known \\ Class\\ F1\end{tabular}}} & \textbf{\begin{tabular}[c]{@{}l@{}}Novel\\ Class \\ Precision\end{tabular}} & \textbf{\begin{tabular}[c]{@{}l@{}}Novel\\ Class\\ Recall\end{tabular}} & \multicolumn{1}{l|}{\textbf{\begin{tabular}[c]{@{}l@{}}Novel \\ Class \\ F1\end{tabular}}} & \textbf{\begin{tabular}[c]{@{}l@{}}Overall\\ Precision\end{tabular}} & \textbf{\begin{tabular}[c]{@{}l@{}}Overall\\ Recall\end{tabular}} & \textbf{\begin{tabular}[c]{@{}l@{}}Overall\\ F1\end{tabular}} \\ \hline
\multicolumn{1}{l|}{\textbf{1000}}                                                       & 74.55                                                                       & 26.02                                                                    & \multicolumn{1}{l|}{38.58}                                                                & 3.43                                                                        & 4.92                                                                    & \multicolumn{1}{l|}{4.04}                                                                  & 38.99                                                                & 15.47                                                             & 22.15                                                         \\
\multicolumn{1}{l|}{\textbf{2000}}                                                       & 61.86                                                                       & 18.28                                                                    & \multicolumn{1}{l|}{28.22}                                                                & 11.74                                                                       & 15.12                                                                   & \multicolumn{1}{l|}{13.22}                                                                 & 36.80                                                                & 16.70                                                             & 22.97                                                         \\
\multicolumn{1}{l|}{\textbf{3000}}                                                       & 75.14                                                                       & 23.99                                                                    & \multicolumn{1}{l|}{36.37}                                                                & 27.67                                                                       & 25.67                                                                   & \multicolumn{1}{l|}{26.63}                                                                 & 51.41                                                                & 24.83                                                             & 33.49                                                         \\
\multicolumn{1}{l|}{\textbf{4000}}                                                       & 73.70                                                                       & 20.93                                                                    & \multicolumn{1}{l|}{32.60}                                                                & 34.57                                                                       & 38.53                                                                   & \multicolumn{1}{l|}{36.44}                                                                 & 54.14                                                                & 29.73                                                             & 38.38                                                         \\
\multicolumn{1}{l|}{\textbf{5000}}                                                       & 77.08                                                                       & 19.84                                                                    & \multicolumn{1}{l|}{31.56}                                                                & 40.72                                                                       & 48.88                                                                   & \multicolumn{1}{l|}{44.43}                                                                 & 58.90                                                                & 34.36                                                             & 43.40                                                         \\
\multicolumn{1}{l|}{\textbf{6000}}                                                       & 72.90                                                                       & 19.24                                                                    & \multicolumn{1}{l|}{30.44}                                                                & 43.49                                                                       & 57.53                                                                   & \multicolumn{1}{l|}{49.53}                                                                 & 58.19                                                                & 38.38                                                             & 46.25                                                         \\
\multicolumn{1}{l|}{\textbf{7000}}                                                       & 61.77                                                                       & 8.98                                                                     & \multicolumn{1}{l|}{15.68}                                                                & 44.02                                                                       & 61.30                                                                   & \multicolumn{1}{l|}{51.24}                                                                 & 52.90                                                                & 35.14                                                             & 42.23                                                         \\
\multicolumn{1}{l|}{\textbf{8000}}                                                       & 58.13                                                                       & 8.69                                                                     & \multicolumn{1}{l|}{15.12}                                                                & 47.92                                                                       & 66.64                                                                   & \multicolumn{1}{l|}{55.75}                                                                 & 53.03                                                                & 37.66                                                             & 44.04                                                         \\
\multicolumn{1}{l|}{\textbf{9000}}                                                       & 53.48                                                                       & 6.53                                                                     & \multicolumn{1}{l|}{11.64}                                                                & 48.35                                                                       & 69.98                                                                   & \multicolumn{1}{l|}{57.19}                                                                 & 50.91                                                                & 38.25                                                             & 43.68                                                         \\
\multicolumn{1}{l|}{\textbf{10000}}                                                      & 38.57                                                                       & 6.35                                                                     & \multicolumn{1}{l|}{10.90}                                                                & 48.24                                                                       & 71.15                                                                   & \multicolumn{1}{l|}{57.50}                                                                 & 43.41                                                                & 38.75                                                             & 40.95                                                         \\ \hline
\multicolumn{10}{c}{\textbf{Compute Mean}}                                                                                                                                                                                                                                                                                                                                                                                                                                                                                                                                                                                                                                                                                                                                                                    \\ \hline
\multicolumn{1}{l|}{\textbf{1000}}                                                       & 39.95                                                                       & 6.68                                                                     & \multicolumn{1}{l|}{11.45}                                                                & 1.03                                                                        & 9.21                                                                    & \multicolumn{1}{l|}{1.85}                                                                  & 20.49                                                                & 7.95                                                              & 11.46                                                         \\
\multicolumn{1}{l|}{\textbf{2000}}                                                       & 16.75                                                                       & 1.06                                                                     & \multicolumn{1}{l|}{1.99}                                                                 & 2.01                                                                        & 13.72                                                                   & \multicolumn{1}{l|}{3.51}                                                                  & 9.38                                                                 & 7.39                                                              & 8.27                                                          \\
\multicolumn{1}{l|}{\textbf{3000}}                                                       & 27.00                                                                       & 3.90                                                                     & \multicolumn{1}{l|}{6.82}                                                                 & 3.91                                                                        & 18.67                                                                   & \multicolumn{1}{l|}{6.47}                                                                  & 15.45                                                                & 11.29                                                             & 13.05                                                         \\
\multicolumn{1}{l|}{\textbf{4000}}                                                       & 23.00                                                                       & 1.59                                                                     & \multicolumn{1}{l|}{2.97}                                                                 & 4.62                                                                        & 21.82                                                                   & \multicolumn{1}{l|}{7.63}                                                                  & 13.81                                                                & 11.71                                                             & 12.67                                                         \\
\multicolumn{1}{l|}{\textbf{5000}}                                                       & 19.92                                                                       & 1.56                                                                     & \multicolumn{1}{l|}{2.89}                                                                 & 5.25                                                                        & 24.94                                                                   & \multicolumn{1}{l|}{8.67}                                                                  & 12.59                                                                & 13.25                                                             & 12.91                                                         \\
\multicolumn{1}{l|}{\textbf{6000}}                                                       & 11.00                                                                       & 0.87                                                                     & \multicolumn{1}{l|}{1.61}                                                                 & 8.57                                                                        & 30.25                                                                   & \multicolumn{1}{l|}{13.36}                                                                 & 9.79                                                                 & 15.56                                                             & 12.02                                                         \\
\multicolumn{1}{l|}{\textbf{7000}}                                                       & 20.67                                                                       & 3.02                                                                     & \multicolumn{1}{l|}{5.27}                                                                 & 11.31                                                                       & 33.58                                                                   & \multicolumn{1}{l|}{16.92}                                                                 & 15.99                                                                & 18.30                                                             & 17.07                                                         \\
\multicolumn{1}{l|}{\textbf{8000}}                                                       & 11.00                                                                       & 0.75                                                                     & \multicolumn{1}{l|}{1.40}                                                                 & 12.96                                                                       & 37.76                                                                   & \multicolumn{1}{l|}{19.30}                                                                 & 11.98                                                                & 19.26                                                             & 14.77                                                         \\
\multicolumn{1}{l|}{\textbf{9000}}                                                       & 11.00                                                                       & 0.11                                                                     & \multicolumn{1}{l|}{0.22}                                                                 & 15.43                                                                       & 41.53                                                                   & \multicolumn{1}{l|}{22.50}                                                                 & 13.22                                                                & 20.82                                                             & 16.17                                                         \\
\multicolumn{1}{l|}{\textbf{10000}}                                                      & 12.00                                                                       & 0.32                                                                     & \multicolumn{1}{l|}{0.62}                                                                 & 18.28                                                                       & 45.52                                                                   & \multicolumn{1}{l|}{26.08}                                                                 & 15.14                                                                & 22.92                                                             & 18.23                                                         \\ \hline
\multicolumn{10}{c}{\textbf{Compute Euclid Distance}}                                                                                                                                                                                                                                                                                                                                                                                                                                                                                                                                                                                                                                                                                                                                                         \\ \hline
\multicolumn{1}{l|}{\textbf{1000}}                                                       & 79.11                                                                       & 40.70                                                                    & \multicolumn{1}{l|}{53.75}                                                                & 17.10                                                                       & 16.37                                                                   & \multicolumn{1}{l|}{16.73}                                                                 & 48.10                                                                & 28.54                                                             & 35.82                                                         \\
\multicolumn{1}{l|}{\textbf{2000}}                                                       & 85.75                                                                       & 35.22                                                                    & \multicolumn{1}{l|}{49.93}                                                                & 35.85                                                                       & 31.74                                                                   & \multicolumn{1}{l|}{33.67}                                                                 & 60.80                                                                & 33.48                                                             & 43.18                                                         \\
\multicolumn{1}{l|}{\textbf{3000}}                                                       & 76.62                                                                       & 17.90                                                                    & \multicolumn{1}{l|}{29.02}                                                                & 46.81                                                                       & 44.39                                                                   & \multicolumn{1}{l|}{45.57}                                                                 & 61.72                                                                & 31.14                                                             & 41.39                                                         \\
\multicolumn{1}{l|}{\textbf{4000}}                                                       & 72.88                                                                       & 13.35                                                                    & \multicolumn{1}{l|}{22.57}                                                                & 47.42                                                                       & 56.88                                                                   & \multicolumn{1}{l|}{51.72}                                                                 & 60.15                                                                & 35.11                                                             & 44.34                                                         \\
\multicolumn{1}{l|}{\textbf{5000}}                                                       & 59.96                                                                       & 11.48                                                                    & \multicolumn{1}{l|}{19.27}                                                                & 49.27                                                                       & 67.04                                                                   & \multicolumn{1}{l|}{56.80}                                                                 & 54.62                                                                & 39.26                                                             & 45.68                                                         \\
\multicolumn{1}{l|}{\textbf{6000}}                                                       & 43.27                                                                       & 6.98                                                                     & \multicolumn{1}{l|}{12.02}                                                                & 49.74                                                                       & 70.48                                                                   & \multicolumn{1}{l|}{58.32}                                                                 & 46.50                                                                & 38.73                                                             & 42.26                                                         \\
\multicolumn{1}{l|}{\textbf{7000}}                                                       & 33.50                                                                       & 3.19                                                                     & \multicolumn{1}{l|}{5.83}                                                                 & 48.67                                                                       & 73.52                                                                   & \multicolumn{1}{l|}{58.57}                                                                 & 41.09                                                                & 38.35                                                             & 39.67                                                         \\
\multicolumn{1}{l|}{\textbf{8000}}                                                       & 35.45                                                                       & 4.49                                                                     & \multicolumn{1}{l|}{7.97}                                                                 & 49.63                                                                       & 75.66                                                                   & \multicolumn{1}{l|}{59.94}                                                                 & 42.54                                                                & 40.08                                                             & 41.27                                                         \\
\multicolumn{1}{l|}{\textbf{9000}}                                                       & 26.33                                                                       & 2.93                                                                     & \multicolumn{1}{l|}{5.27}                                                                 & 50.30                                                                       & 78.02                                                                   & \multicolumn{1}{l|}{61.17}                                                                 & 38.32                                                                & 40.47                                                             & 39.37                                                         \\
\multicolumn{1}{l|}{\textbf{10000}}                                                      & 17.00                                                                       & 3.14                                                                     & \multicolumn{1}{l|}{5.30}                                                                 & 51.87                                                                       & 79.31                                                                   & \multicolumn{1}{l|}{62.72}                                                                 & 34.43                                                                & 41.23                                                             & 37.52                                                         \\ \hline
\multicolumn{10}{c}{\textbf{Compute Mahalanobis Distance}}                                                                                                                                                                                                                                                                                                                                                                                                                                                                                                                                                                                                                                                                                                                                                    \\ \hline
\multicolumn{1}{l|}{\textbf{1000}}                                                       & 77.64                                                                       & 31.72                                                                    & \multicolumn{1}{l|}{45.04}                                                                & 11.59                                                                       & 15.90                                                                   & \multicolumn{1}{l|}{13.41}                                                                 & 44.61                                                                & 23.81                                                             & 31.05                                                         \\
\multicolumn{1}{l|}{\textbf{2000}}                                                       & 79.25                                                                       & 22.94                                                                    & \multicolumn{1}{l|}{35.58}                                                                & 27.57                                                                       & 32.88                                                                   & \multicolumn{1}{l|}{29.99}                                                                 & 53.41                                                                & 27.91                                                             & 36.66                                                         \\
\multicolumn{1}{l|}{\textbf{3000}}                                                       & 88.60                                                                       & 23.51                                                                    & \multicolumn{1}{l|}{37.16}                                                                & 35.89                                                                       & 42.37                                                                   & \multicolumn{1}{l|}{38.86}                                                                 & 62.25                                                                & 32.94                                                             & 43.08                                                         \\
\multicolumn{1}{l|}{\textbf{4000}}                                                       & 78.20                                                                       & 10.35                                                                    & \multicolumn{1}{l|}{18.28}                                                                & 42.53                                                                       & 56.15                                                                   & \multicolumn{1}{l|}{48.40}                                                                 & 60.37                                                                & 33.25                                                             & 42.88                                                         \\
\multicolumn{1}{l|}{\textbf{5000}}                                                       & 50.86                                                                       & 7.91                                                                     & \multicolumn{1}{l|}{13.69}                                                                & 45.89                                                                       & 60.54                                                                   & \multicolumn{1}{l|}{52.21}                                                                 & 48.38                                                                & 34.23                                                             & 40.09                                                         \\
\multicolumn{1}{l|}{\textbf{6000}}                                                       & 50.45                                                                       & 6.47                                                                     & \multicolumn{1}{l|}{11.47}                                                                & 47.37                                                                       & 70.73                                                                   & \multicolumn{1}{l|}{56.74}                                                                 & 48.91                                                                & 38.60                                                             & 43.15                                                         \\
\multicolumn{1}{l|}{\textbf{7000}}                                                       & 27.26                                                                       & 2.83                                                                     & \multicolumn{1}{l|}{5.13}                                                                 & 47.90                                                                       & 74.23                                                                   & \multicolumn{1}{l|}{58.23}                                                                 & 37.58                                                                & 38.53                                                             & 38.05                                                         \\
\multicolumn{1}{l|}{\textbf{8000}}                                                       & 21.00                                                                       & 2.65                                                                     & \multicolumn{1}{l|}{4.71}                                                                 & 49.53                                                                       & 77.02                                                                   & \multicolumn{1}{l|}{60.29}                                                                 & 35.27                                                                & 39.83                                                             & 37.41                                                         \\
\multicolumn{1}{l|}{\textbf{9000}}                                                       & 22.00                                                                       & 3.56                                                                     & \multicolumn{1}{l|}{6.13}                                                                 & 50.31                                                                       & 78.15                                                                   & \multicolumn{1}{l|}{61.21}                                                                 & 36.15                                                                & 40.85                                                             & 38.36                                                         \\
\multicolumn{1}{l|}{\textbf{10000}}                                                      & 17.97                                                                       & 1.25                                                                     & \multicolumn{1}{l|}{2.34}                                                                 & 51.21                                                                       & 80.49                                                                   & \multicolumn{1}{l|}{62.60}                                                                 & 34.59                                                                & 40.87                                                             & 37.47                                                         \\ \hline
\multicolumn{10}{c}{\textbf{Compute Max Probability}}                                                                                                                                                                                                                                                                                                                                                                                                                                                                                                                                                                                                                                                                                                                                                         \\ \hline
\multicolumn{1}{l|}{\textbf{1000}}                                                       & 64.52                                                                       & 20.06                                                                    & \multicolumn{1}{l|}{30.60}                                                                & 3.20                                                                        & 6.38                                                                    & \multicolumn{1}{l|}{4.26}                                                                  & 33.86                                                                & 13.22                                                             & 19.02                                                         \\
\multicolumn{1}{l|}{\textbf{2000}}                                                       & 84.61                                                                       & 25.01                                                                    & \multicolumn{1}{l|}{38.61}                                                                & 18.63                                                                       & 22.15                                                                   & \multicolumn{1}{l|}{20.24}                                                                 & 51.62                                                                & 23.58                                                             & 32.37                                                         \\
\multicolumn{1}{l|}{\textbf{3000}}                                                       & 57.24                                                                       & 11.32                                                                    & \multicolumn{1}{l|}{18.90}                                                                & 24.95                                                                       & 35.16                                                                   & \multicolumn{1}{l|}{29.19}                                                                 & 41.10                                                                & 23.24                                                             & 29.69                                                         \\
\multicolumn{1}{l|}{\textbf{4000}}                                                       & 63.88                                                                       & 11.03                                                                    & \multicolumn{1}{l|}{18.81}                                                                & 36.70                                                                       & 47.23                                                                   & \multicolumn{1}{l|}{41.30}                                                                 & 50.29                                                                & 29.13                                                             & 36.89                                                         \\
\multicolumn{1}{l|}{\textbf{5000}}                                                       & 56.02                                                                       & 6.07                                                                     & \multicolumn{1}{l|}{10.95}                                                                & 42.32                                                                       & 58.27                                                                   & \multicolumn{1}{l|}{49.03}                                                                 & 49.17                                                                & 32.17                                                             & 38.89                                                         \\
\multicolumn{1}{l|}{\textbf{6000}}                                                       & 43.84                                                                       & 5.29                                                                     & \multicolumn{1}{l|}{9.44}                                                                 & 45.00                                                                       & 65.42                                                                   & \multicolumn{1}{l|}{53.32}                                                                 & 44.42                                                                & 35.35                                                             & 39.37                                                         \\
\multicolumn{1}{l|}{\textbf{7000}}                                                       & 37.02                                                                       & 5.00                                                                     & \multicolumn{1}{l|}{8.81}                                                                 & 47.77                                                                       & 71.41                                                                   & \multicolumn{1}{l|}{57.25}                                                                 & 42.39                                                                & 38.21                                                             & 40.19                                                         \\
\multicolumn{1}{l|}{\textbf{8000}}                                                       & 28.85                                                                       & 4.55                                                                     & \multicolumn{1}{l|}{7.86}                                                                 & 48.98                                                                       & 75.65                                                                   & \multicolumn{1}{l|}{59.46}                                                                 & 38.91                                                                & 40.10                                                             & 39.50                                                         \\
\multicolumn{1}{l|}{\textbf{9000}}                                                       & 16.00                                                                       & 1.86                                                                     & \multicolumn{1}{l|}{3.33}                                                                 & 49.15                                                                       & 78.29                                                                   & \multicolumn{1}{l|}{60.39}                                                                 & 32.58                                                                & 40.07                                                             & 35.94                                                         \\
\multicolumn{1}{l|}{\textbf{10000}}                                                      & 19.90                                                                       & 2.05                                                                     & \multicolumn{1}{l|}{3.72}                                                                 & 51.34                                                                       & 79.65                                                                   & \multicolumn{1}{l|}{62.44}                                                                 & 35.62                                                                & 40.85                                                             & 38.06                                                         \\ \hline
\multicolumn{10}{c}{\textbf{Placeholders Algorithm}}                                                                                                                                                                                                                                                                                                                                                                                                                                                                                                                                                                                                                                                                                                                                                          \\ \hline
\multicolumn{1}{l|}{\textbf{1000}}                                                       & 75.81                                                                       & 26.33                                                                    & \multicolumn{1}{l|}{39.09}                                                                & 5.38                                                                        & 10.21                                                                   & \multicolumn{1}{l|}{7.05}                                                                  & 40.60                                                                & 18.27                                                             & 25.20                                                         \\
\multicolumn{1}{l|}{\textbf{2000}}                                                       & 83.63                                                                       & 24.58                                                                    & \multicolumn{1}{l|}{37.99}                                                                & 18.81                                                                       & 27.74                                                                   & \multicolumn{1}{l|}{22.42}                                                                 & 51.22                                                                & 26.16                                                             & 34.63                                                         \\
\multicolumn{1}{l|}{\textbf{3000}}                                                       & 84.29                                                                       & 19.97                                                                    & \multicolumn{1}{l|}{32.29}                                                                & 28.87                                                                       & 40.48                                                                   & \multicolumn{1}{l|}{33.70}                                                                 & 56.58                                                                & 30.23                                                             & 39.41                                                         \\
\multicolumn{1}{l|}{\textbf{4000}}                                                       & 59.16                                                                       & 9.35                                                                     & \multicolumn{1}{l|}{16.15}                                                                & 35.60                                                                       & 50.84                                                                   & \multicolumn{1}{l|}{41.88}                                                                 & 47.38                                                                & 30.09                                                             & 36.81                                                         \\
\multicolumn{1}{l|}{\textbf{5000}}                                                       & 60.67                                                                       & 6.64                                                                     & \multicolumn{1}{l|}{11.97}                                                                & 42.08                                                                       & 59.11                                                                   & \multicolumn{1}{l|}{49.16}                                                                 & 51.38                                                                & 32.87                                                             & 40.09                                                         \\
\multicolumn{1}{l|}{\textbf{6000}}                                                       & 50.79                                                                       & 5.49                                                                     & \multicolumn{1}{l|}{9.91}                                                                 & 48.35                                                                       & 69.81                                                                   & \multicolumn{1}{l|}{57.13}                                                                 & 49.57                                                                & 37.65                                                             & 42.80                                                         \\
\multicolumn{1}{l|}{\textbf{7000}}                                                       & 29.00                                                                       & 3.19                                                                     & \multicolumn{1}{l|}{5.75}                                                                 & 47.59                                                                       & 74.18                                                                   & \multicolumn{1}{l|}{57.98}                                                                 & 38.30                                                                & 38.68                                                             & 38.49                                                         \\
\multicolumn{1}{l|}{\textbf{8000}}                                                       & 26.99                                                                       & 3.25                                                                     & \multicolumn{1}{l|}{5.80}                                                                 & 48.55                                                                       & 73.98                                                                   & \multicolumn{1}{l|}{58.63}                                                                 & 37.77                                                                & 38.62                                                             & 38.19                                                         \\
\multicolumn{1}{l|}{\textbf{9000}}                                                       & 25.70                                                                       & 2.20                                                                     & \multicolumn{1}{l|}{4.05}                                                                 & 50.46                                                                       & 77.10                                                                   & \multicolumn{1}{l|}{61.00}                                                                 & 38.08                                                                & 39.65                                                             & 38.85                                                         \\
\multicolumn{1}{l|}{\textbf{10000}}                                                      & 19.94                                                                       & 1.77                                                                     & \multicolumn{1}{l|}{3.25}                                                                 & 49.57                                                                       & 78.38                                                                   & \multicolumn{1}{l|}{60.73}                                                                 & 34.76                                                                & 40.07                                                             & 37.23                                                         \\ \hline
\multicolumn{10}{c}{\textbf{Few shot Open set Recognition}}                                                                                                                                                                                                                                                                                                                                                                                                                                                                                                                                                                                                                                                                                                                                                   \\ \hline
\end{tabular}
    }
    \caption{
   Novelty Accommodation Stage: Dataset 1: Further Fine-tune using $D^F$.
    }
    \label{tab:ds1_s22}
\end{table*}
\begin{table*}[t!]
    \centering
    \small
    \resizebox{\linewidth}{!}
    {
       \begin{tabular}{llllllllll}
\hline
\multicolumn{1}{l|}{\textbf{\begin{tabular}[c]{@{}l@{}}\# of \\ Novelties\end{tabular}}} & \textbf{\begin{tabular}[c]{@{}l@{}}Known \\ Class\\ precision\end{tabular}} & \textbf{\begin{tabular}[c]{@{}l@{}}Known\\ class \\ recall\end{tabular}} & \multicolumn{1}{l|}{\textbf{\begin{tabular}[c]{@{}l@{}}Known \\ Class\\ F1\end{tabular}}} & \textbf{\begin{tabular}[c]{@{}l@{}}Novel\\ Class \\ Precision\end{tabular}} & \textbf{\begin{tabular}[c]{@{}l@{}}Novel\\ Class\\ Recall\end{tabular}} & \multicolumn{1}{l|}{\textbf{\begin{tabular}[c]{@{}l@{}}Novel \\ Class \\ F1\end{tabular}}} & \textbf{\begin{tabular}[c]{@{}l@{}}Overall\\ Precision\end{tabular}} & \textbf{\begin{tabular}[c]{@{}l@{}}Overall\\ Recall\end{tabular}} & \textbf{\begin{tabular}[c]{@{}l@{}}Overall\\ F1\end{tabular}} \\ \hline
\multicolumn{1}{l|}{\textbf{1000}}                                                       & 59.69                                                                       & 87.26                                                                    & \multicolumn{1}{l|}{70.89}                                                                & 15.37                                                                       & 4.64                                                                    & \multicolumn{1}{l|}{7.13}                                                                  & 37.53                                                                & 45.95                                                             & 41.32                                                         \\
\multicolumn{1}{l|}{\textbf{2000}}                                                       & 62.66                                                                       & 87.70                                                                    & \multicolumn{1}{l|}{73.09}                                                                & 32.20                                                                       & 14.14                                                                   & \multicolumn{1}{l|}{19.65}                                                                 & 47.43                                                                & 50.92                                                             & 49.11                                                         \\
\multicolumn{1}{l|}{\textbf{3000}}                                                       & 65.50                                                                       & 86.96                                                                    & \multicolumn{1}{l|}{74.72}                                                                & 48.58                                                                       & 22.47                                                                   & \multicolumn{1}{l|}{30.73}                                                                 & 57.04                                                                & 54.72                                                             & 55.86                                                         \\
\multicolumn{1}{l|}{\textbf{4000}}                                                       & 68.97                                                                       & 87.08                                                                    & \multicolumn{1}{l|}{76.97}                                                                & 62.66                                                                       & 34.79                                                                   & \multicolumn{1}{l|}{44.74}                                                                 & 65.81                                                                & 60.94                                                             & 63.28                                                         \\
\multicolumn{1}{l|}{\textbf{5000}}                                                       & 73.44                                                                       & 86.87                                                                    & \multicolumn{1}{l|}{79.59}                                                                & 65.70                                                                       & 44.87                                                                   & \multicolumn{1}{l|}{53.32}                                                                 & 69.57                                                                & 65.87                                                             & 67.67                                                         \\
\multicolumn{1}{l|}{\textbf{6000}}                                                       & 75.33                                                                       & 85.70                                                                    & \multicolumn{1}{l|}{80.18}                                                                & 67.43                                                                       & 51.22                                                                   & \multicolumn{1}{l|}{58.22}                                                                 & 71.38                                                                & 68.46                                                             & 69.89                                                         \\
\multicolumn{1}{l|}{\textbf{7000}}                                                       & 78.45                                                                       & 85.66                                                                    & \multicolumn{1}{l|}{81.90}                                                                & 70.25                                                                       & 57.98                                                                   & \multicolumn{1}{l|}{63.53}                                                                 & 74.35                                                                & 71.82                                                             & 73.06                                                         \\
\multicolumn{1}{l|}{\textbf{8000}}                                                       & 80.33                                                                       & 85.03                                                                    & \multicolumn{1}{l|}{82.61}                                                                & 72.67                                                                       & 62.59                                                                   & \multicolumn{1}{l|}{67.25}                                                                 & 76.50                                                                & 73.81                                                             & 75.13                                                         \\
\multicolumn{1}{l|}{\textbf{9000}}                                                       & 81.31                                                                       & 84.65                                                                    & \multicolumn{1}{l|}{82.95}                                                                & 73.82                                                                       & 66.87                                                                   & \multicolumn{1}{l|}{70.17}                                                                 & 77.56                                                                & 75.76                                                             & 76.65                                                         \\
\multicolumn{1}{l|}{\textbf{10000}}                                                      & 82.88                                                                       & 84.44                                                                    & \multicolumn{1}{l|}{83.65}                                                                & 74.13                                                                       & 68.83                                                                   & \multicolumn{1}{l|}{71.38}                                                                 & 78.50                                                                & 76.64                                                             & 77.56                                                         \\ \hline
\multicolumn{10}{c}{\textbf{Compute Mean}}                                                                                                                                                                                                                                                                                                                                                                                                                                                                                                                                                                                                                                                                                                                                                                    \\ \hline
\multicolumn{1}{l|}{\textbf{1000}}                                                       & 57.36                                                                       & 88.99                                                                    & \multicolumn{1}{l|}{69.76}                                                                & 3.95                                                                        & 8.73                                                                    & \multicolumn{1}{l|}{5.44}                                                                  & 30.66                                                                & 48.86                                                             & 37.68                                                         \\
\multicolumn{1}{l|}{\textbf{2000}}                                                       & 60.92                                                                       & 88.75                                                                    & \multicolumn{1}{l|}{72.25}                                                                & 5.96                                                                        & 12.81                                                                   & \multicolumn{1}{l|}{8.14}                                                                  & 33.44                                                                & 50.78                                                             & 40.32                                                         \\
\multicolumn{1}{l|}{\textbf{3000}}                                                       & 63.45                                                                       & 88.03                                                                    & \multicolumn{1}{l|}{73.75}                                                                & 9.74                                                                        & 17.96                                                                   & \multicolumn{1}{l|}{12.63}                                                                 & 36.60                                                                & 52.99                                                             & 43.30                                                         \\
\multicolumn{1}{l|}{\textbf{4000}}                                                       & 65.72                                                                       & 87.84                                                                    & \multicolumn{1}{l|}{75.19}                                                                & 11.33                                                                       & 20.86                                                                   & \multicolumn{1}{l|}{14.68}                                                                 & 38.52                                                                & 54.35                                                             & 45.09                                                         \\
\multicolumn{1}{l|}{\textbf{5000}}                                                       & 65.68                                                                       & 87.91                                                                    & \multicolumn{1}{l|}{75.19}                                                                & 13.52                                                                       & 23.76                                                                   & \multicolumn{1}{l|}{17.23}                                                                 & 39.60                                                                & 55.83                                                             & 46.33                                                         \\
\multicolumn{1}{l|}{\textbf{6000}}                                                       & 68.42                                                                       & 87.28                                                                    & \multicolumn{1}{l|}{76.71}                                                                & 18.28                                                                       & 28.87                                                                   & \multicolumn{1}{l|}{22.39}                                                                 & 43.35                                                                & 58.08                                                             & 49.65                                                         \\
\multicolumn{1}{l|}{\textbf{7000}}                                                       & 69.71                                                                       & 86.83                                                                    & \multicolumn{1}{l|}{77.33}                                                                & 21.18                                                                       & 31.86                                                                   & \multicolumn{1}{l|}{25.44}                                                                 & 45.44                                                                & 59.34                                                             & 51.47                                                         \\
\multicolumn{1}{l|}{\textbf{8000}}                                                       & 71.32                                                                       & 87.33                                                                    & \multicolumn{1}{l|}{78.52}                                                                & 25.46                                                                       & 35.86                                                                   & \multicolumn{1}{l|}{29.78}                                                                 & 48.39                                                                & 61.60                                                             & 54.20                                                         \\
\multicolumn{1}{l|}{\textbf{9000}}                                                       & 73.99                                                                       & 86.78                                                                    & \multicolumn{1}{l|}{79.88}                                                                & 28.26                                                                       & 39.38                                                                   & \multicolumn{1}{l|}{32.91}                                                                 & 51.12                                                                & 63.08                                                             & 56.47                                                         \\
\multicolumn{1}{l|}{\textbf{10000}}                                                      & 75.89                                                                       & 86.23                                                                    & \multicolumn{1}{l|}{80.73}                                                                & 32.54                                                                       & 43.27                                                                   & \multicolumn{1}{l|}{37.15}                                                                 & 54.21                                                                & 64.75                                                             & 59.01                                                         \\ \hline
\multicolumn{10}{c}{\textbf{Compute Euclid Distance}}                                                                                                                                                                                                                                                                                                                                                                                                                                                                                                                                                                                                                                                                                                                                                         \\ \hline
\multicolumn{1}{l|}{\textbf{1000}}                                                       & 63.23                                                                       & 87.81                                                                    & \multicolumn{1}{l|}{73.52}                                                                & 30.48                                                                       & 15.33                                                                   & \multicolumn{1}{l|}{20.40}                                                                 & 46.86                                                                & 51.57                                                             & 49.10                                                         \\
\multicolumn{1}{l|}{\textbf{2000}}                                                       & 68.99                                                                       & 87.47                                                                    & \multicolumn{1}{l|}{77.14}                                                                & 52.53                                                                       & 28.39                                                                   & \multicolumn{1}{l|}{36.86}                                                                 & 60.76                                                                & 57.93                                                             & 59.31                                                         \\
\multicolumn{1}{l|}{\textbf{3000}}                                                       & 72.93                                                                       & 86.67                                                                    & \multicolumn{1}{l|}{79.21}                                                                & 64.15                                                                       & 42.46                                                                   & \multicolumn{1}{l|}{51.10}                                                                 & 68.54                                                                & 64.57                                                             & 66.50                                                         \\
\multicolumn{1}{l|}{\textbf{4000}}                                                       & 75.98                                                                       & 86.22                                                                    & \multicolumn{1}{l|}{80.78}                                                                & 70.70                                                                       & 53.82                                                                   & \multicolumn{1}{l|}{61.12}                                                                 & 73.34                                                                & 70.02                                                             & 71.64                                                         \\
\multicolumn{1}{l|}{\textbf{5000}}                                                       & 79.22                                                                       & 85.36                                                                    & \multicolumn{1}{l|}{82.18}                                                                & 73.09                                                                       & 62.86                                                                   & \multicolumn{1}{l|}{67.59}                                                                 & 76.15                                                                & 74.11                                                             & 75.12                                                         \\
\multicolumn{1}{l|}{\textbf{6000}}                                                       & 81.70                                                                       & 85.02                                                                    & \multicolumn{1}{l|}{83.33}                                                                & 74.91                                                                       & 68.35                                                                   & \multicolumn{1}{l|}{71.48}                                                                 & 78.31                                                                & 76.69                                                             & 77.49                                                         \\
\multicolumn{1}{l|}{\textbf{7000}}                                                       & 82.01                                                                       & 85.48                                                                    & \multicolumn{1}{l|}{83.71}                                                                & 76.69                                                                       & 70.31                                                                   & \multicolumn{1}{l|}{73.36}                                                                 & 79.35                                                                & 77.89                                                             & 78.61                                                         \\
\multicolumn{1}{l|}{\textbf{8000}}                                                       & 82.80                                                                       & 85.56                                                                    & \multicolumn{1}{l|}{84.16}                                                                & 77.90                                                                       & 72.09                                                                   & \multicolumn{1}{l|}{74.88}                                                                 & 80.35                                                                & 78.83                                                             & 79.58                                                         \\
\multicolumn{1}{l|}{\textbf{9000}}                                                       & 83.54                                                                       & 84.62                                                                    & \multicolumn{1}{l|}{84.08}                                                                & 77.92                                                                       & 73.29                                                                   & \multicolumn{1}{l|}{75.53}                                                                 & 80.73                                                                & 78.95                                                             & 79.83                                                         \\
\multicolumn{1}{l|}{\textbf{10000}}                                                      & 84.50                                                                       & 84.31                                                                    & \multicolumn{1}{l|}{84.40}                                                                & 78.39                                                                       & 75.05                                                                   & \multicolumn{1}{l|}{76.68}                                                                 & 81.45                                                                & 79.68                                                             & 80.56                                                         \\ \hline
\multicolumn{10}{c}{\textbf{Compute Mahalanobis Distance}}                                                                                                                                                                                                                                                                                                                                                                                                                                                                                                                                                                                                                                                                                                                                                    \\ \hline
\multicolumn{1}{l|}{\textbf{1000}}                                                       & 61.52                                                                       & 88.11                                                                    & \multicolumn{1}{l|}{72.45}                                                                & 32.72                                                                       & 16.57                                                                   & \multicolumn{1}{l|}{22.00}                                                                 & 47.12                                                                & 52.34                                                             & 49.59                                                         \\
\multicolumn{1}{l|}{\textbf{2000}}                                                       & 68.34                                                                       & 87.25                                                                    & \multicolumn{1}{l|}{76.65}                                                                & 49.43                                                                       & 30.10                                                                   & \multicolumn{1}{l|}{37.42}                                                                 & 58.89                                                                & 58.67                                                             & 58.78                                                         \\
\multicolumn{1}{l|}{\textbf{3000}}                                                       & 73.16                                                                       & 86.78                                                                    & \multicolumn{1}{l|}{79.39}                                                                & 61.57                                                                       & 41.30                                                                   & \multicolumn{1}{l|}{49.44}                                                                 & 67.36                                                                & 64.04                                                             & 65.66                                                         \\
\multicolumn{1}{l|}{\textbf{4000}}                                                       & 76.49                                                                       & 86.25                                                                    & \multicolumn{1}{l|}{81.08}                                                                & 67.36                                                                       & 51.58                                                                   & \multicolumn{1}{l|}{58.42}                                                                 & 71.93                                                                & 68.92                                                             & 70.39                                                         \\
\multicolumn{1}{l|}{\textbf{5000}}                                                       & 78.93                                                                       & 85.77                                                                    & \multicolumn{1}{l|}{82.21}                                                                & 72.94                                                                       & 61.55                                                                   & \multicolumn{1}{l|}{66.76}                                                                 & 75.94                                                                & 73.66                                                             & 74.78                                                         \\
\multicolumn{1}{l|}{\textbf{6000}}                                                       & 80.99                                                                       & 85.69                                                                    & \multicolumn{1}{l|}{83.27}                                                                & 74.78                                                                       & 66.77                                                                   & \multicolumn{1}{l|}{70.55}                                                                 & 77.88                                                                & 76.23                                                             & 77.05                                                         \\
\multicolumn{1}{l|}{\textbf{7000}}                                                       & 81.71                                                                       & 84.69                                                                    & \multicolumn{1}{l|}{83.17}                                                                & 75.56                                                                       & 68.76                                                                   & \multicolumn{1}{l|}{72.00}                                                                 & 78.63                                                                & 76.72                                                             & 77.66                                                         \\
\multicolumn{1}{l|}{\textbf{8000}}                                                       & 83.11                                                                       & 85.45                                                                    & \multicolumn{1}{l|}{84.26}                                                                & 77.29                                                                       & 72.60                                                                   & \multicolumn{1}{l|}{74.87}                                                                 & 80.20                                                                & 79.03                                                             & 79.61                                                         \\
\multicolumn{1}{l|}{\textbf{9000}}                                                       & 83.76                                                                       & 84.45                                                                    & \multicolumn{1}{l|}{84.10}                                                                & 77.86                                                                       & 73.79                                                                   & \multicolumn{1}{l|}{75.77}                                                                 & 80.81                                                                & 79.12                                                             & 79.96                                                         \\
\multicolumn{1}{l|}{\textbf{10000}}                                                      & 84.55                                                                       & 84.18                                                                    & \multicolumn{1}{l|}{84.36}                                                                & 78.58                                                                       & 75.10                                                                   & \multicolumn{1}{l|}{76.80}                                                                 & 81.57                                                                & 79.64                                                             & 80.59                                                         \\ \hline
\multicolumn{10}{c}{\textbf{Compute Max Probability}}                                                                                                                                                                                                                                                                                                                                                                                                                                                                                                                                                                                                                                                                                                                                                         \\ \hline
\multicolumn{1}{l|}{\textbf{1000}}                                                       & 58.20                                                                       & 88.66                                                                    & \multicolumn{1}{l|}{70.27}                                                                & 23.09                                                                       & 12.71                                                                   & \multicolumn{1}{l|}{16.40}                                                                 & 40.65                                                                & 50.69                                                             & 45.12                                                         \\
\multicolumn{1}{l|}{\textbf{2000}}                                                       & 65.05                                                                       & 88.25                                                                    & \multicolumn{1}{l|}{74.89}                                                                & 37.00                                                                       & 26.00                                                                   & \multicolumn{1}{l|}{30.54}                                                                 & 51.02                                                                & 57.12                                                             & 53.90                                                         \\
\multicolumn{1}{l|}{\textbf{3000}}                                                       & 68.66                                                                       & 87.88                                                                    & \multicolumn{1}{l|}{77.09}                                                                & 57.39                                                                       & 35.26                                                                   & \multicolumn{1}{l|}{43.68}                                                                 & 63.03                                                                & 61.57                                                             & 62.29                                                         \\
\multicolumn{1}{l|}{\textbf{4000}}                                                       & 73.22                                                                       & 87.58                                                                    & \multicolumn{1}{l|}{79.76}                                                                & 64.21                                                                       & 47.60                                                                   & \multicolumn{1}{l|}{54.67}                                                                 & 68.71                                                                & 67.59                                                             & 68.15                                                         \\
\multicolumn{1}{l|}{\textbf{5000}}                                                       & 77.43                                                                       & 86.70                                                                    & \multicolumn{1}{l|}{81.80}                                                                & 70.83                                                                       & 57.74                                                                   & \multicolumn{1}{l|}{63.62}                                                                 & 74.13                                                                & 72.22                                                             & 73.16                                                         \\
\multicolumn{1}{l|}{\textbf{6000}}                                                       & 79.69                                                                       & 86.50                                                                    & \multicolumn{1}{l|}{82.96}                                                                & 74.79                                                                       & 63.95                                                                   & \multicolumn{1}{l|}{68.95}                                                                 & 77.24                                                                & 75.23                                                             & 76.22                                                         \\
\multicolumn{1}{l|}{\textbf{7000}}                                                       & 81.81                                                                       & 85.93                                                                    & \multicolumn{1}{l|}{83.82}                                                                & 76.06                                                                       & 69.15                                                                   & \multicolumn{1}{l|}{72.44}                                                                 & 78.93                                                                & 77.54                                                             & 78.23                                                         \\
\multicolumn{1}{l|}{\textbf{8000}}                                                       & 82.86                                                                       & 85.38                                                                    & \multicolumn{1}{l|}{84.10}                                                                & 76.89                                                                       & 71.83                                                                   & \multicolumn{1}{l|}{74.27}                                                                 & 79.88                                                                & 78.60                                                             & 79.23                                                         \\
\multicolumn{1}{l|}{\textbf{9000}}                                                       & 84.09                                                                       & 85.00                                                                    & \multicolumn{1}{l|}{84.54}                                                                & 77.74                                                                       & 74.36                                                                   & \multicolumn{1}{l|}{76.01}                                                                 & 80.91                                                                & 79.68                                                             & 80.29                                                         \\
\multicolumn{1}{l|}{\textbf{10000}}                                                      & 84.60                                                                       & 84.47                                                                    & \multicolumn{1}{l|}{84.53}                                                                & 78.24                                                                       & 75.36                                                                   & \multicolumn{1}{l|}{76.77}                                                                 & 81.42                                                                & 79.91                                                             & 80.66                                                         \\ \hline
\multicolumn{10}{c}{\textbf{Placeholders Algorithm}}                                                                                                                                                                                                                                                                                                                                                                                                                                                                                                                                                                                                                                                                                                                                                          \\ \hline
\multicolumn{1}{l|}{\textbf{1000}}                                                       & 58.94                                                                       & 88.66                                                                    & \multicolumn{1}{l|}{70.81}                                                                & 28.65                                                                       & 17.12                                                                   & \multicolumn{1}{l|}{21.43}                                                                 & 43.80                                                                & 52.89                                                             & 47.92                                                         \\
\multicolumn{1}{l|}{\textbf{2000}}                                                       & 64.40                                                                       & 88.71                                                                    & \multicolumn{1}{l|}{74.63}                                                                & 49.10                                                                       & 27.78                                                                   & \multicolumn{1}{l|}{35.48}                                                                 & 56.75                                                                & 58.25                                                             & 57.49                                                         \\
\multicolumn{1}{l|}{\textbf{3000}}                                                       & 68.68                                                                       & 88.11                                                                    & \multicolumn{1}{l|}{77.19}                                                                & 59.69                                                                       & 38.73                                                                   & \multicolumn{1}{l|}{46.98}                                                                 & 64.19                                                                & 63.42                                                             & 63.80                                                         \\
\multicolumn{1}{l|}{\textbf{4000}}                                                       & 72.46                                                                       & 87.32                                                                    & \multicolumn{1}{l|}{79.20}                                                                & 68.83                                                                       & 46.58                                                                   & \multicolumn{1}{l|}{55.56}                                                                 & 70.64                                                                & 66.95                                                             & 68.75                                                         \\
\multicolumn{1}{l|}{\textbf{5000}}                                                       & 75.73                                                                       & 87.02                                                                    & \multicolumn{1}{l|}{80.98}                                                                & 75.08                                                                       & 58.55                                                                   & \multicolumn{1}{l|}{65.79}                                                                 & 75.41                                                                & 72.79                                                             & 74.08                                                         \\
\multicolumn{1}{l|}{\textbf{6000}}                                                       & 79.48                                                                       & 85.99                                                                    & \multicolumn{1}{l|}{82.61}                                                                & 75.94                                                                       & 65.82                                                                   & \multicolumn{1}{l|}{70.52}                                                                 & 77.71                                                                & 75.91                                                             & 76.80                                                         \\
\multicolumn{1}{l|}{\textbf{7000}}                                                       & 81.18                                                                       & 85.87                                                                    & \multicolumn{1}{l|}{83.46}                                                                & 77.52                                                                       & 69.19                                                                   & \multicolumn{1}{l|}{73.12}                                                                 & 79.35                                                                & 77.53                                                             & 78.43                                                         \\
\multicolumn{1}{l|}{\textbf{8000}}                                                       & 83.05                                                                       & 85.40                                                                    & \multicolumn{1}{l|}{84.21}                                                                & 78.70                                                                       & 73.36                                                                   & \multicolumn{1}{l|}{75.94}                                                                 & 80.87                                                                & 79.38                                                             & 80.12                                                         \\
\multicolumn{1}{l|}{\textbf{9000}}                                                       & 82.48                                                                       & 85.17                                                                    & \multicolumn{1}{l|}{83.80}                                                                & 78.39                                                                       & 72.41                                                                   & \multicolumn{1}{l|}{75.28}                                                                 & 80.44                                                                & 78.79                                                             & 79.61                                                         \\
\multicolumn{1}{l|}{\textbf{10000}}                                                      & 83.14                                                                       & 85.43                                                                    & \multicolumn{1}{l|}{84.27}                                                                & 78.92                                                                       & 73.88                                                                   & \multicolumn{1}{l|}{76.32}                                                                 & 81.03                                                                & 79.65                                                             & 80.33                                                         \\ \hline
\multicolumn{10}{c}{\textbf{Few shot Open set Recognition}}                                                                                                                                                                                                                                                                                                                                                                                                                                                                                                                                                                                                                                                                                                                                                   \\ \hline
\end{tabular}
    }
    \caption{
   Novelty Accommodation Stage: Dataset 1: Further Fine-tune using Sampled $D^T$ and $D^F$.
    }
    \label{tab:ds1_s23}
\end{table*}

\begin{table*}[t!]
    \centering
    \small
    \resizebox{\linewidth}{!}
    {
        \begin{tabular}{llllllllll}
\hline
\multicolumn{1}{l|}{\textbf{\begin{tabular}[c]{@{}l@{}}\# of \\ Novelties\end{tabular}}} & \textbf{\begin{tabular}[c]{@{}l@{}}Known \\ Class\\ precision\end{tabular}} & \textbf{\begin{tabular}[c]{@{}l@{}}Known\\ class \\ recall\end{tabular}} & \multicolumn{1}{l|}{\textbf{\begin{tabular}[c]{@{}l@{}}Known \\ Class\\ F1\end{tabular}}} & \textbf{\begin{tabular}[c]{@{}l@{}}Novel\\ Class \\ Precision\end{tabular}} & \textbf{\begin{tabular}[c]{@{}l@{}}Novel\\ Class\\ Recall\end{tabular}} & \multicolumn{1}{l|}{\textbf{\begin{tabular}[c]{@{}l@{}}Novel \\ Class \\ F1\end{tabular}}} & \textbf{\begin{tabular}[c]{@{}l@{}}Overall\\ Precision\end{tabular}} & \textbf{\begin{tabular}[c]{@{}l@{}}Overall\\ Recall\end{tabular}} & \textbf{\begin{tabular}[c]{@{}l@{}}Overall\\ F1\end{tabular}} \\ \hline
\multicolumn{1}{l|}{\textbf{4000}}                                                       & 57.47                                                                       & 89.90                                                                    & \multicolumn{1}{l|}{70.12}                                                                & 57.28                                                                       & 20.96                                                                   & \multicolumn{1}{l|}{30.69}                                                                 & 57.38                                                                & 55.43                                                             & 56.39                                                         \\
\multicolumn{1}{l|}{\textbf{8000}}                                                       & 66.31                                                                       & 88.96                                                                    & \multicolumn{1}{l|}{75.98}                                                                & 83.59                                                                       & 45.46                                                                   & \multicolumn{1}{l|}{58.89}                                                                 & 74.95                                                                & 67.21                                                             & 70.87                                                         \\
\multicolumn{1}{l|}{\textbf{12000}}                                                      & 71.26                                                                       & 88.93                                                                    & \multicolumn{1}{l|}{79.12}                                                                & 85.16                                                                       & 57.43                                                                   & \multicolumn{1}{l|}{68.60}                                                                 & 78.21                                                                & 73.18                                                             & 75.61                                                         \\
\multicolumn{1}{l|}{\textbf{16000}}                                                      & 75.42                                                                       & 89.55                                                                    & \multicolumn{1}{l|}{81.88}                                                                & 86.76                                                                       & 67.05                                                                   & \multicolumn{1}{l|}{75.64}                                                                 & 81.09                                                                & 78.30                                                             & 79.67                                                         \\
\multicolumn{1}{l|}{\textbf{20000}}                                                      & 77.47                                                                       & 89.41                                                                    & \multicolumn{1}{l|}{83.01}                                                                & 88.40                                                                       & 71.43                                                                   & \multicolumn{1}{l|}{79.01}                                                                 & 82.94                                                                & 80.42                                                             & 81.66                                                         \\ \hline
\multicolumn{10}{c}{\textbf{Compute Mean}}                                                                                                                                                                                                                                                                                                                                                                                                                                                                                                                                                                                                                                                                                                                                                                    \\ \hline
\multicolumn{1}{l|}{\textbf{4000}}                                                       & 56.84                                                                       & 89.63                                                                    & \multicolumn{1}{l|}{69.56}                                                                & 9.09                                                                        & 12.39                                                                   & \multicolumn{1}{l|}{10.49}                                                                 & 32.96                                                                & 51.01                                                             & 40.05                                                         \\
\multicolumn{1}{l|}{\textbf{8000}}                                                       & 60.52                                                                       & 89.47                                                                    & \multicolumn{1}{l|}{72.20}                                                                & 15.40                                                                       & 21.08                                                                   & \multicolumn{1}{l|}{17.80}                                                                 & 37.96                                                                & 55.27                                                             & 45.01                                                         \\
\multicolumn{1}{l|}{\textbf{12000}}                                                      & 62.75                                                                       & 89.67                                                                    & \multicolumn{1}{l|}{73.83}                                                                & 23.36                                                                       & 28.78                                                                   & \multicolumn{1}{l|}{25.79}                                                                 & 43.06                                                                & 59.22                                                             & 49.86                                                         \\
\multicolumn{1}{l|}{\textbf{16000}}                                                      & 66.27                                                                       & 89.51                                                                    & \multicolumn{1}{l|}{76.16}                                                                & 31.26                                                                       & 36.25                                                                   & \multicolumn{1}{l|}{33.57}                                                                 & 48.77                                                                & 62.88                                                             & 54.93                                                         \\
\multicolumn{1}{l|}{\textbf{20000}}                                                      & 68.62                                                                       & 89.19                                                                    & \multicolumn{1}{l|}{77.56}                                                                & 40.37                                                                       & 43.76                                                                   & \multicolumn{1}{l|}{42.00}                                                                 & 54.49                                                                & 66.47                                                             & 59.89                                                         \\ \hline
\multicolumn{10}{c}{\textbf{Compute Euclid Distance}}                                                                                                                                                                                                                                                                                                                                                                                                                                                                                                                                                                                                                                                                                                                                                         \\ \hline
\multicolumn{1}{l|}{\textbf{4000}}                                                       & 65.56                                                                       & 89.81                                                                    & \multicolumn{1}{l|}{75.79}                                                                & 73.37                                                                       & 39.72                                                                   & \multicolumn{1}{l|}{51.54}                                                                 & 69.47                                                                & 64.77                                                             & 67.04                                                         \\
\multicolumn{1}{l|}{\textbf{8000}}                                                       & 73.69                                                                       & 89.30                                                                    & \multicolumn{1}{l|}{80.75}                                                                & 85.84                                                                       & 63.55                                                                   & \multicolumn{1}{l|}{73.03}                                                                 & 79.77                                                                & 76.43                                                             & 78.06                                                         \\
\multicolumn{1}{l|}{\textbf{12000}}                                                      & 77.47                                                                       & 89.19                                                                    & \multicolumn{1}{l|}{82.92}                                                                & 88.01                                                                       & 71.65                                                                   & \multicolumn{1}{l|}{78.99}                                                                 & 82.74                                                                & 80.42                                                             & 81.56                                                         \\
\multicolumn{1}{l|}{\textbf{16000}}                                                      & 79.19                                                                       & 88.81                                                                    & \multicolumn{1}{l|}{83.72}                                                                & 88.18                                                                       & 74.73                                                                   & \multicolumn{1}{l|}{80.90}                                                                 & 83.68                                                                & 81.77                                                             & 82.71                                                         \\
\multicolumn{1}{l|}{\textbf{20000}}                                                      & 80.33                                                                       & 88.63                                                                    & \multicolumn{1}{l|}{84.28}                                                                & 89.57                                                                       & 77.62                                                                   & \multicolumn{1}{l|}{83.17}                                                                 & 84.95                                                                & 83.12                                                             & 84.03                                                         \\ \hline
\multicolumn{10}{c}{\textbf{Compute Mahalanobis Distance}}                                                                                                                                                                                                                                                                                                                                                                                                                                                                                                                                                                                                                                                                                                                                                    \\ \hline
\multicolumn{1}{l|}{\textbf{4000}}                                                       & 65.24                                                                       & 89.96                                                                    & \multicolumn{1}{l|}{75.63}                                                                & 75.98                                                                       & 38.95                                                                   & \multicolumn{1}{l|}{51.50}                                                                 & 70.61                                                                & 64.46                                                             & 67.39                                                         \\
\multicolumn{1}{l|}{\textbf{8000}}                                                       & 73.35                                                                       & 89.48                                                                    & \multicolumn{1}{l|}{80.62}                                                                & 85.56                                                                       & 61.38                                                                   & \multicolumn{1}{l|}{71.48}                                                                 & 79.45                                                                & 75.43                                                             & 77.39                                                         \\
\multicolumn{1}{l|}{\textbf{12000}}                                                      & 76.74                                                                       & 89.29                                                                    & \multicolumn{1}{l|}{82.54}                                                                & 87.99                                                                       & 70.02                                                                   & \multicolumn{1}{l|}{77.98}                                                                 & 82.37                                                                & 79.66                                                             & 80.99                                                         \\
\multicolumn{1}{l|}{\textbf{16000}}                                                      & 80.08                                                                       & 89.05                                                                    & \multicolumn{1}{l|}{84.33}                                                                & 89.10                                                                       & 76.62                                                                   & \multicolumn{1}{l|}{82.39}                                                                 & 84.59                                                                & 82.83                                                             & 83.70                                                         \\
\multicolumn{1}{l|}{\textbf{20000}}                                                      & 80.83                                                                       & 89.01                                                                    & \multicolumn{1}{l|}{84.72}                                                                & 90.47                                                                       & 78.57                                                                   & \multicolumn{1}{l|}{84.10}                                                                 & 85.65                                                                & 83.79                                                             & 84.71                                                         \\ \hline
\multicolumn{10}{c}{\textbf{Compute Max Probability}}                                                                                                                                                                                                                                                                                                                                                                                                                                                                                                                                                                                                                                                                                                                                                         \\ \hline
\multicolumn{1}{l|}{\textbf{4000}}                                                       & 61.55                                                                       & 90.02                                                                    & \multicolumn{1}{l|}{73.11}                                                                & 58.92                                                                       & 28.77                                                                   & \multicolumn{1}{l|}{38.66}                                                                 & 60.24                                                                & 59.40                                                             & 59.82                                                         \\
\multicolumn{1}{l|}{\textbf{8000}}                                                       & 70.28                                                                       & 89.56                                                                    & \multicolumn{1}{l|}{78.76}                                                                & 82.11                                                                       & 53.64                                                                   & \multicolumn{1}{l|}{64.89}                                                                 & 76.19                                                                & 71.60                                                             & 73.82                                                         \\
\multicolumn{1}{l|}{\textbf{12000}}                                                      & 76.39                                                                       & 88.87                                                                    & \multicolumn{1}{l|}{82.16}                                                                & 87.62                                                                       & 68.74                                                                   & \multicolumn{1}{l|}{77.04}                                                                 & 82.00                                                                & 78.80                                                             & 80.37                                                         \\
\multicolumn{1}{l|}{\textbf{16000}}                                                      & 80.09                                                                       & 89.08                                                                    & \multicolumn{1}{l|}{84.35}                                                                & 88.95                                                                       & 76.14                                                                   & \multicolumn{1}{l|}{82.05}                                                                 & 84.52                                                                & 82.61                                                             & 83.55                                                         \\
\multicolumn{1}{l|}{\textbf{20000}}                                                      & 81.25                                                                       & 88.80                                                                    & \multicolumn{1}{l|}{84.86}                                                                & 89.25                                                                       & 78.42                                                                   & \multicolumn{1}{l|}{83.49}                                                                 & 85.25                                                                & 83.61                                                             & 84.42                                                         \\ \hline
\multicolumn{10}{c}{\textbf{Placeholders Algorithm}}                                                                                                                                                                                                                                                                                                                                                                                                                                                                                                                                                                                                                                                                                                                                                          \\ \hline
\multicolumn{1}{l|}{\textbf{4000}}                                                       & 61.32                                                                       & 89.86                                                                    & \multicolumn{1}{l|}{72.90}                                                                & 57.04                                                                       & 30.12                                                                   & \multicolumn{1}{l|}{39.42}                                                                 & 59.18                                                                & 59.99                                                             & 59.58                                                         \\
\multicolumn{1}{l|}{\textbf{8000}}                                                       & 69.68                                                                       & 89.87                                                                    & \multicolumn{1}{l|}{78.50}                                                                & 85.11                                                                       & 52.89                                                                   & \multicolumn{1}{l|}{65.24}                                                                 & 77.39                                                                & 71.38                                                             & 74.26                                                         \\
\multicolumn{1}{l|}{\textbf{12000}}                                                      & 74.92                                                                       & 89.04                                                                    & \multicolumn{1}{l|}{81.37}                                                                & 87.46                                                                       & 65.95                                                                   & \multicolumn{1}{l|}{75.20}                                                                 & 81.19                                                                & 77.50                                                             & 79.30                                                         \\
\multicolumn{1}{l|}{\textbf{16000}}                                                      & 78.69                                                                       & 88.63                                                                    & \multicolumn{1}{l|}{83.36}                                                                & 88.87                                                                       & 74.19                                                                   & \multicolumn{1}{l|}{80.87}                                                                 & 83.78                                                                & 81.41                                                             & 82.58                                                         \\
\multicolumn{1}{l|}{\textbf{20000}}                                                      & 80.82                                                                       & 88.82                                                                    & \multicolumn{1}{l|}{84.63}                                                                & 90.20                                                                       & 78.17                                                                   & \multicolumn{1}{l|}{83.76}                                                                 & 85.51                                                                & 83.49                                                             & 84.49                                                         \\ \hline
\multicolumn{10}{c}{\textbf{Few shot Open set Recognition}}                                                                                                                                                                                                                                                                                                                                                                                                                                                                                                                                                                                                                                                                                                                                                   \\ \hline
\end{tabular}
    }
    \caption{
   Novelty Accommodation Stage: Dataset 2: Retrain using $D^T$ and $D^F$
    }
    \label{tab:ds2_s21}
\end{table*}
\begin{table*}[t!]
    \centering
    \small
    \resizebox{\linewidth}{!}
    {
       \begin{tabular}{llllllllll}
\hline
\multicolumn{1}{l|}{\textbf{\begin{tabular}[c]{@{}l@{}}\# of \\ Novelties\end{tabular}}} & \textbf{\begin{tabular}[c]{@{}l@{}}Known \\ Class\\ precision\end{tabular}} & \textbf{\begin{tabular}[c]{@{}l@{}}Known\\ class \\ recall\end{tabular}} & \multicolumn{1}{l|}{\textbf{\begin{tabular}[c]{@{}l@{}}Known \\ Class\\ F1\end{tabular}}} & \textbf{\begin{tabular}[c]{@{}l@{}}Novel\\ Class \\ Precision\end{tabular}} & \textbf{\begin{tabular}[c]{@{}l@{}}Novel\\ Class\\ Recall\end{tabular}} & \multicolumn{1}{l|}{\textbf{\begin{tabular}[c]{@{}l@{}}Novel \\ Class \\ F1\end{tabular}}} & \textbf{\begin{tabular}[c]{@{}l@{}}Overall\\ Precision\end{tabular}} & \textbf{\begin{tabular}[c]{@{}l@{}}Overall\\ Recall\end{tabular}} & \textbf{\begin{tabular}[c]{@{}l@{}}Overall\\ F1\end{tabular}} \\ \hline
\multicolumn{1}{l|}{\textbf{4000}}                                                       & 83.55                                                                       & 28.67                                                                    & \multicolumn{1}{l|}{42.69}                                                                & 33.25                                                                       & 32.69                                                                   & \multicolumn{1}{l|}{32.97}                                                                 & 58.40                                                                & 30.68                                                             & 40.23                                                         \\
\multicolumn{1}{l|}{\textbf{8000}}                                                       & 48.80                                                                       & 7.04                                                                     & \multicolumn{1}{l|}{12.30}                                                                & 45.59                                                                       & 61.05                                                                   & \multicolumn{1}{l|}{52.20}                                                                 & 47.20                                                                & 34.05                                                             & 39.56                                                         \\
\multicolumn{1}{l|}{\textbf{12000}}                                                      & 36.71                                                                       & 4.56                                                                     & \multicolumn{1}{l|}{8.11}                                                                 & 49.62                                                                       & 75.64                                                                   & \multicolumn{1}{l|}{59.93}                                                                 & 43.17                                                                & 40.10                                                             & 41.58                                                         \\
\multicolumn{1}{l|}{\textbf{16000}}                                                      & 18.00                                                                       & 2.26                                                                     & \multicolumn{1}{l|}{4.02}                                                                 & 53.16                                                                       & 80.24                                                                   & \multicolumn{1}{l|}{63.95}                                                                 & 35.58                                                                & 41.25                                                             & 38.21                                                         \\
\multicolumn{1}{l|}{\textbf{20000}}                                                      & 10.00                                                                       & 3.27                                                                     & \multicolumn{1}{l|}{4.93}                                                                 & 55.63                                                                       & 84.25                                                                   & \multicolumn{1}{l|}{67.01}                                                                 & 32.81                                                                & 43.76                                                             & 37.50                                                         \\ \hline
\multicolumn{10}{c}{\textbf{Compute Mean}}                                                                                                                                                                                                                                                                                                                                                                                                                                                                                                                                                                                                                                                                                                                                                                    \\ \hline
\multicolumn{1}{l|}{\textbf{4000}}                                                       & 12.89                                                                       & 0.76                                                                     & \multicolumn{1}{l|}{1.44}                                                                 & 1.90                                                                        & 14.41                                                                   & \multicolumn{1}{l|}{3.36}                                                                  & 7.39                                                                 & 7.58                                                              & 7.48                                                          \\
\multicolumn{1}{l|}{\textbf{8000}}                                                       & 6.00                                                                        & 1.75                                                                     & \multicolumn{1}{l|}{2.71}                                                                 & 4.59                                                                        & 22.88                                                                   & \multicolumn{1}{l|}{7.65}                                                                  & 5.29                                                                 & 12.32                                                             & 7.40                                                          \\
\multicolumn{1}{l|}{\textbf{12000}}                                                      & 13.98                                                                       & 2.31                                                                     & \multicolumn{1}{l|}{3.96}                                                                 & 8.50                                                                        & 31.69                                                                   & \multicolumn{1}{l|}{13.40}                                                                 & 11.24                                                                & 17.00                                                             & 13.53                                                         \\
\multicolumn{1}{l|}{\textbf{16000}}                                                      & 7.76                                                                        & 0.22                                                                     & \multicolumn{1}{l|}{0.43}                                                                 & 13.20                                                                       & 39.82                                                                   & \multicolumn{1}{l|}{19.83}                                                                 & 10.48                                                                & 20.02                                                             & 13.76                                                         \\
\multicolumn{1}{l|}{\textbf{20000}}                                                      & 10.00                                                                       & 0.13                                                                     & \multicolumn{1}{l|}{0.26}                                                                 & 19.13                                                                       & 48.70                                                                   & \multicolumn{1}{l|}{27.47}                                                                 & 14.57                                                                & 24.42                                                             & 18.25                                                         \\ \hline
\multicolumn{10}{c}{\textbf{Compute Euclid Distance}}                                                                                                                                                                                                                                                                                                                                                                                                                                                                                                                                                                                                                                                                                                                                                         \\ \hline
\multicolumn{1}{l|}{\textbf{4000}}                                                       & 79.08                                                                       & 15.55                                                                    & \multicolumn{1}{l|}{25.99}                                                                & 45.61                                                                       & 54.41                                                                   & \multicolumn{1}{l|}{49.62}                                                                 & 62.34                                                                & 34.98                                                             & 44.81                                                         \\
\multicolumn{1}{l|}{\textbf{8000}}                                                       & 30.00                                                                       & 3.81                                                                     & \multicolumn{1}{l|}{6.76}                                                                 & 51.96                                                                       & 79.00                                                                   & \multicolumn{1}{l|}{62.69}                                                                 & 40.98                                                                & 41.41                                                             & 41.19                                                         \\
\multicolumn{1}{l|}{\textbf{12000}}                                                      & 15.00                                                                       & 1.50                                                                     & \multicolumn{1}{l|}{2.73}                                                                 & 54.29                                                                       & 84.03                                                                   & \multicolumn{1}{l|}{65.96}                                                                 & 34.65                                                                & 42.76                                                             & 38.28                                                         \\
\multicolumn{1}{l|}{\textbf{16000}}                                                      & 10.00                                                                       & 0.75                                                                     & \multicolumn{1}{l|}{1.40}                                                                 & 55.82                                                                       & 86.85                                                                   & \multicolumn{1}{l|}{67.96}                                                                 & 32.91                                                                & 43.80                                                             & 37.58                                                         \\
\multicolumn{1}{l|}{\textbf{20000}}                                                      & 4.00                                                                        & 0.42                                                                     & \multicolumn{1}{l|}{0.76}                                                                 & 55.57                                                                       & 87.67                                                                   & \multicolumn{1}{l|}{68.02}                                                                 & 29.78                                                                & 44.04                                                             & 35.53                                                         \\ \hline
\multicolumn{10}{c}{\textbf{Compute Mahalanobis Distance}}                                                                                                                                                                                                                                                                                                                                                                                                                                                                                                                                                                                                                                                                                                                                                    \\ \hline
\multicolumn{1}{l|}{\textbf{4000}}                                                       & 64.78                                                                       & 14.62                                                                    & \multicolumn{1}{l|}{23.86}                                                                & 43.47                                                                       & 52.66                                                                   & \multicolumn{1}{l|}{47.63}                                                                 & 54.12                                                                & 33.64                                                             & 41.49                                                         \\
\multicolumn{1}{l|}{\textbf{8000}}                                                       & 27.95                                                                       & 2.69                                                                     & \multicolumn{1}{l|}{4.91}                                                                 & 53.26                                                                       & 77.48                                                                   & \multicolumn{1}{l|}{63.13}                                                                 & 40.61                                                                & 40.09                                                             & 40.35                                                         \\
\multicolumn{1}{l|}{\textbf{12000}}                                                      & 9.33                                                                        & 0.47                                                                     & \multicolumn{1}{l|}{0.89}                                                                 & 53.85                                                                       & 83.92                                                                   & \multicolumn{1}{l|}{65.60}                                                                 & 31.59                                                                & 42.20                                                             & 36.13                                                         \\
\multicolumn{1}{l|}{\textbf{16000}}                                                      & 11.00                                                                       & 1.31                                                                     & \multicolumn{1}{l|}{2.34}                                                                 & 55.60                                                                       & 86.41                                                                   & \multicolumn{1}{l|}{67.66}                                                                 & 33.30                                                                & 43.86                                                             & 37.86                                                         \\
\multicolumn{1}{l|}{\textbf{20000}}                                                      & 6.00                                                                        & 0.04                                                                     & \multicolumn{1}{l|}{0.08}                                                                 & 56.65                                                                       & 87.98                                                                   & \multicolumn{1}{l|}{68.92}                                                                 & 31.33                                                                & 44.01                                                             & 36.60                                                         \\ \hline
\multicolumn{10}{c}{\textbf{Compute Max Probability}}                                                                                                                                                                                                                                                                                                                                                                                                                                                                                                                                                                                                                                                                                                                                                         \\ \hline
\multicolumn{1}{l|}{\textbf{4000}}                                                       & 70.51                                                                       & 20.17                                                                    & \multicolumn{1}{l|}{31.37}                                                                & 36.49                                                                       & 39.94                                                                   & \multicolumn{1}{l|}{38.14}                                                                 & 53.50                                                                & 30.05                                                             & 38.48                                                         \\
\multicolumn{1}{l|}{\textbf{8000}}                                                       & 25.66                                                                       & 1.86                                                                     & \multicolumn{1}{l|}{3.47}                                                                 & 46.68                                                                       & 69.62                                                                   & \multicolumn{1}{l|}{55.89}                                                                 & 36.17                                                                & 35.74                                                             & 35.95                                                         \\
\multicolumn{1}{l|}{\textbf{12000}}                                                      & 8.00                                                                        & 1.15                                                                     & \multicolumn{1}{l|}{2.01}                                                                 & 53.00                                                                       & 82.74                                                                   & \multicolumn{1}{l|}{64.61}                                                                 & 30.50                                                                & 41.94                                                             & 35.32                                                         \\
\multicolumn{1}{l|}{\textbf{16000}}                                                      & 8.00                                                                        & 0.55                                                                     & \multicolumn{1}{l|}{1.03}                                                                 & 55.56                                                                       & 86.32                                                                   & \multicolumn{1}{l|}{67.61}                                                                 & 31.78                                                                & 43.43                                                             & 36.70                                                         \\
\multicolumn{1}{l|}{\textbf{20000}}                                                      & 7.00                                                                        & 0.24                                                                     & \multicolumn{1}{l|}{0.46}                                                                 & 57.11                                                                       & 87.49                                                                   & \multicolumn{1}{l|}{69.11}                                                                 & 32.05                                                                & 43.87                                                             & 37.04                                                         \\ \hline
\multicolumn{10}{c}{\textbf{Placeholders Algorithm}}                                                                                                                                                                                                                                                                                                                                                                                                                                                                                                                                                                                                                                                                                                                                                          \\ \hline
\multicolumn{1}{l|}{\textbf{4000}}                                                       & 67.19                                                                       & 13.68                                                                    & \multicolumn{1}{l|}{22.73}                                                                & 30.37                                                                       & 40.68                                                                   & \multicolumn{1}{l|}{34.78}                                                                 & 48.78                                                                & 27.18                                                             & 34.91                                                         \\
\multicolumn{1}{l|}{\textbf{8000}}                                                       & 31.93                                                                       & 2.62                                                                     & \multicolumn{1}{l|}{4.84}                                                                 & 47.52                                                                       & 69.70                                                                   & \multicolumn{1}{l|}{56.51}                                                                 & 39.72                                                                & 36.16                                                             & 37.86                                                         \\
\multicolumn{1}{l|}{\textbf{12000}}                                                      & 9.00                                                                        & 1.53                                                                     & \multicolumn{1}{l|}{2.62}                                                                 & 53.00                                                                       & 82.51                                                                   & \multicolumn{1}{l|}{64.54}                                                                 & 31.00                                                                & 42.02                                                             & 35.68                                                         \\
\multicolumn{1}{l|}{\textbf{16000}}                                                      & 7.00                                                                        & 0.63                                                                     & \multicolumn{1}{l|}{1.16}                                                                 & 55.64                                                                       & 86.24                                                                   & \multicolumn{1}{l|}{67.64}                                                                 & 31.32                                                                & 43.43                                                             & 36.39                                                         \\
\multicolumn{1}{l|}{\textbf{20000}}                                                      & 7.00                                                                        & 0.43                                                                     & \multicolumn{1}{l|}{0.81}                                                                 & 56.06                                                                       & 87.89                                                                   & \multicolumn{1}{l|}{68.46}                                                                 & 31.53                                                                & 44.16                                                             & 36.79                                                         \\ \hline
\multicolumn{10}{c}{\textbf{Few shot Open set Recognition}}                                                                                                                                                                                                                                                                                                                                                                                                                                                                                                                                                                                                                                                                                                                                                   \\ \hline
\end{tabular}
    }
    \caption{
   Novelty Accommodation Stage: Dataset 2: Further Fine-tune using $D^F$.
    }
    \label{tab:d2_s22}
\end{table*}
\begin{table*}[t!]
    \centering
    \small
    \resizebox{\linewidth}{!}
    {
       \begin{tabular}{llllllllll}
\hline
\multicolumn{1}{l|}{\textbf{\begin{tabular}[c]{@{}l@{}}\# of \\ Novelties\end{tabular}}} & \textbf{\begin{tabular}[c]{@{}l@{}}Known \\ Class\\ precision\end{tabular}} & \textbf{\begin{tabular}[c]{@{}l@{}}Known\\ class \\ recall\end{tabular}} & \multicolumn{1}{l|}{\textbf{\begin{tabular}[c]{@{}l@{}}Known \\ Class\\ F1\end{tabular}}} & \textbf{\begin{tabular}[c]{@{}l@{}}Novel\\ Class \\ Precision\end{tabular}} & \textbf{\begin{tabular}[c]{@{}l@{}}Novel\\ Class\\ Recall\end{tabular}} & \multicolumn{1}{l|}{\textbf{\begin{tabular}[c]{@{}l@{}}Novel \\ Class \\ F1\end{tabular}}} & \textbf{\begin{tabular}[c]{@{}l@{}}Overall\\ Precision\end{tabular}} & \textbf{\begin{tabular}[c]{@{}l@{}}Overall\\ Recall\end{tabular}} & \textbf{\begin{tabular}[c]{@{}l@{}}Overall\\ F1\end{tabular}} \\ \hline
\multicolumn{1}{l|}{\textbf{4000}}                                                       & 67.78                                                                       & 87.40                                                                    & \multicolumn{1}{l|}{76.35}                                                                & 53.65                                                                       & 28.48                                                                   & \multicolumn{1}{l|}{37.21}                                                                 & 60.71                                                                & 57.94                                                             & 59.29                                                         \\
\multicolumn{1}{l|}{\textbf{8000}}                                                       & 78.41                                                                       & 85.86                                                                    & \multicolumn{1}{l|}{81.97}                                                                & 73.39                                                                       & 59.38                                                                   & \multicolumn{1}{l|}{65.65}                                                                 & 75.90                                                                & 72.62                                                             & 74.22                                                         \\
\multicolumn{1}{l|}{\textbf{12000}}                                                      & 83.72                                                                       & 84.33                                                                    & \multicolumn{1}{l|}{84.02}                                                                & 77.60                                                                       & 72.56                                                                   & \multicolumn{1}{l|}{75.00}                                                                 & 80.66                                                                & 78.45                                                             & 79.54                                                         \\
\multicolumn{1}{l|}{\textbf{16000}}                                                      & 86.11                                                                       & 84.61                                                                    & \multicolumn{1}{l|}{85.35}                                                                & 80.39                                                                       & 78.25                                                                   & \multicolumn{1}{l|}{79.31}                                                                 & 83.25                                                                & 81.43                                                             & 82.33                                                         \\
\multicolumn{1}{l|}{\textbf{20000}}                                                      & 86.59                                                                       & 84.91                                                                    & \multicolumn{1}{l|}{85.74}                                                                & 82.77                                                                       & 81.47                                                                   & \multicolumn{1}{l|}{82.11}                                                                 & 84.68                                                                & 83.19                                                             & 83.93                                                         \\ \hline
\multicolumn{10}{c}{\textbf{Compute Mean}}                                                                                                                                                                                                                                                                                                                                                                                                                                                                                                                                                                                                                                                                                                                                                                    \\ \hline
\multicolumn{1}{l|}{\textbf{4000}}                                                       & 59.33                                                                       & 89.27                                                                    & \multicolumn{1}{l|}{71.28}                                                                & 7.03                                                                        & 13.27                                                                   & \multicolumn{1}{l|}{9.19}                                                                  & 33.18                                                                & 51.27                                                             & 40.29                                                         \\
\multicolumn{1}{l|}{\textbf{8000}}                                                       & 64.66                                                                       & 88.79                                                                    & \multicolumn{1}{l|}{74.83}                                                                & 12.25                                                                       & 21.84                                                                   & \multicolumn{1}{l|}{15.70}                                                                 & 38.46                                                                & 55.31                                                             & 45.37                                                         \\
\multicolumn{1}{l|}{\textbf{12000}}                                                      & 67.24                                                                       & 88.77                                                                    & \multicolumn{1}{l|}{76.52}                                                                & 19.30                                                                       & 29.96                                                                   & \multicolumn{1}{l|}{23.48}                                                                 & 43.27                                                                & 59.37                                                             & 50.06                                                         \\
\multicolumn{1}{l|}{\textbf{16000}}                                                      & 71.70                                                                       & 88.24                                                                    & \multicolumn{1}{l|}{79.11}                                                                & 26.76                                                                       & 37.66                                                                   & \multicolumn{1}{l|}{31.29}                                                                 & 49.23                                                                & 62.95                                                             & 55.25                                                         \\
\multicolumn{1}{l|}{\textbf{20000}}                                                      & 75.77                                                                       & 88.14                                                                    & \multicolumn{1}{l|}{81.49}                                                                & 35.22                                                                       & 46.40                                                                   & \multicolumn{1}{l|}{40.04}                                                                 & 55.49                                                                & 67.27                                                             & 60.81                                                         \\ \hline
\multicolumn{10}{c}{\textbf{Compute Euclid Distance}}                                                                                                                                                                                                                                                                                                                                                                                                                                                                                                                                                                                                                                                                                                                                                         \\ \hline
\multicolumn{1}{l|}{\textbf{4000}}                                                       & 74.92                                                                       & 87.17                                                                    & \multicolumn{1}{l|}{80.58}                                                                & 71.98                                                                       & 53.51                                                                   & \multicolumn{1}{l|}{61.39}                                                                 & 73.45                                                                & 70.34                                                             & 71.86                                                         \\
\multicolumn{1}{l|}{\textbf{8000}}                                                       & 83.65                                                                       & 86.13                                                                    & \multicolumn{1}{l|}{84.87}                                                                & 81.38                                                                       & 75.63                                                                   & \multicolumn{1}{l|}{78.40}                                                                 & 82.52                                                                & 80.88                                                             & 81.69                                                         \\
\multicolumn{1}{l|}{\textbf{12000}}                                                      & 85.43                                                                       & 85.89                                                                    & \multicolumn{1}{l|}{85.66}                                                                & 84.03                                                                       & 80.79                                                                   & \multicolumn{1}{l|}{82.38}                                                                 & 84.73                                                                & 83.34                                                             & 84.03                                                         \\
\multicolumn{1}{l|}{\textbf{16000}}                                                      & 86.43                                                                       & 86.35                                                                    & \multicolumn{1}{l|}{86.39}                                                                & 85.35                                                                       & 82.88                                                                   & \multicolumn{1}{l|}{84.10}                                                                 & 85.89                                                                & 84.61                                                             & 85.25                                                         \\
\multicolumn{1}{l|}{\textbf{20000}}                                                      & 87.02                                                                       & 86.11                                                                    & \multicolumn{1}{l|}{86.56}                                                                & 85.92                                                                       & 84.20                                                                   & \multicolumn{1}{l|}{85.05}                                                                 & 86.47                                                                & 85.15                                                             & 85.80                                                         \\ \hline
\multicolumn{10}{c}{\textbf{Compute Mahalanobis Distance}}                                                                                                                                                                                                                                                                                                                                                                                                                                                                                                                                                                                                                                                                                                                                                    \\ \hline
\multicolumn{1}{l|}{\textbf{4000}}                                                       & 75.00                                                                       & 87.16                                                                    & \multicolumn{1}{l|}{80.62}                                                                & 69.84                                                                       & 52.95                                                                   & \multicolumn{1}{l|}{60.23}                                                                 & 72.42                                                                & 70.05                                                             & 71.22                                                         \\
\multicolumn{1}{l|}{\textbf{8000}}                                                       & 82.97                                                                       & 86.17                                                                    & \multicolumn{1}{l|}{84.54}                                                                & 80.70                                                                       & 74.21                                                                   & \multicolumn{1}{l|}{77.32}                                                                 & 81.83                                                                & 80.19                                                             & 81.00                                                         \\
\multicolumn{1}{l|}{\textbf{12000}}                                                      & 85.56                                                                       & 85.86                                                                    & \multicolumn{1}{l|}{85.71}                                                                & 83.39                                                                       & 80.00                                                                   & \multicolumn{1}{l|}{81.66}                                                                 & 84.48                                                                & 82.93                                                             & 83.70                                                         \\
\multicolumn{1}{l|}{\textbf{16000}}                                                      & 86.82                                                                       & 86.13                                                                    & \multicolumn{1}{l|}{86.47}                                                                & 85.13                                                                       & 82.81                                                                   & \multicolumn{1}{l|}{83.95}                                                                 & 85.97                                                                & 84.47                                                             & 85.21                                                         \\
\multicolumn{1}{l|}{\textbf{20000}}                                                      & 87.49                                                                       & 86.27                                                                    & \multicolumn{1}{l|}{86.88}                                                                & 85.72                                                                       & 84.46                                                                   & \multicolumn{1}{l|}{85.09}                                                                 & 86.61                                                                & 85.36                                                             & 85.98                                                         \\ \hline
\multicolumn{10}{c}{\textbf{Compute Max Probability}}                                                                                                                                                                                                                                                                                                                                                                                                                                                                                                                                                                                                                                                                                                                                                         \\ \hline
\multicolumn{1}{l|}{\textbf{4000}}                                                       & 69.16                                                                       & 88.39                                                                    & \multicolumn{1}{l|}{77.60}                                                                & 61.06                                                                       & 39.69                                                                   & \multicolumn{1}{l|}{48.11}                                                                 & 65.11                                                                & 64.04                                                             & 64.57                                                         \\
\multicolumn{1}{l|}{\textbf{8000}}                                                       & 80.98                                                                       & 86.87                                                                    & \multicolumn{1}{l|}{83.82}                                                                & 78.15                                                                       & 68.53                                                                   & \multicolumn{1}{l|}{73.02}                                                                 & 79.57                                                                & 77.70                                                             & 78.62                                                         \\
\multicolumn{1}{l|}{\textbf{12000}}                                                      & 84.78                                                                       & 85.84                                                                    & \multicolumn{1}{l|}{85.31}                                                                & 82.84                                                                       & 78.16                                                                   & \multicolumn{1}{l|}{80.43}                                                                 & 83.81                                                                & 82.00                                                             & 82.90                                                         \\
\multicolumn{1}{l|}{\textbf{16000}}                                                      & 86.71                                                                       & 85.50                                                                    & \multicolumn{1}{l|}{86.10}                                                                & 84.36                                                                       & 82.70                                                                   & \multicolumn{1}{l|}{83.52}                                                                 & 85.54                                                                & 84.10                                                             & 84.81                                                         \\
\multicolumn{1}{l|}{\textbf{20000}}                                                      & 87.21                                                                       & 85.93                                                                    & \multicolumn{1}{l|}{86.57}                                                                & 85.63                                                                       & 84.05                                                                   & \multicolumn{1}{l|}{84.83}                                                                 & 86.42                                                                & 84.99                                                             & 85.70                                                         \\ \hline
\multicolumn{10}{c}{\textbf{Placeholders Algorithm}}                                                                                                                                                                                                                                                                                                                                                                                                                                                                                                                                                                                                                                                                                                                                                          \\ \hline
\multicolumn{1}{l|}{\textbf{4000}}                                                       & 67.89                                                                       & 89.14                                                                    & \multicolumn{1}{l|}{77.08}                                                                & 60.03                                                                       & 40.77                                                                   & \multicolumn{1}{l|}{48.56}                                                                 & 63.96                                                                & 64.95                                                             & 64.45                                                         \\
\multicolumn{1}{l|}{\textbf{8000}}                                                       & 77.86                                                                       & 88.05                                                                    & \multicolumn{1}{l|}{82.64}                                                                & 80.32                                                                       & 65.58                                                                   & \multicolumn{1}{l|}{72.21}                                                                 & 79.09                                                                & 76.82                                                             & 77.94                                                         \\
\multicolumn{1}{l|}{\textbf{12000}}                                                      & 83.66                                                                       & 87.03                                                                    & \multicolumn{1}{l|}{85.31}                                                                & 84.28                                                                       & 78.54                                                                   & \multicolumn{1}{l|}{81.31}                                                                 & 83.97                                                                & 82.78                                                             & 83.37                                                         \\
\multicolumn{1}{l|}{\textbf{16000}}                                                      & 85.80                                                                       & 86.50                                                                    & \multicolumn{1}{l|}{86.15}                                                                & 85.24                                                                       & 81.85                                                                   & \multicolumn{1}{l|}{83.51}                                                                 & 85.52                                                                & 84.18                                                             & 84.84                                                         \\
\multicolumn{1}{l|}{\textbf{20000}}                                                      & 86.72                                                                       & 86.51                                                                    & \multicolumn{1}{l|}{86.61}                                                                & 86.32                                                                       & 83.99                                                                   & \multicolumn{1}{l|}{85.14}                                                                 & 86.52                                                                & 85.25                                                             & 85.88                                                         \\ \hline
\multicolumn{10}{c}{\textbf{Few shot Open set Recognition}}                                                                                                                                                                                                                                                                                                                                                                                                                                                                                                                                                                                                                                                                                                                                                   \\ \hline
\end{tabular}
    }
    \caption{
   Novelty Accommodation Stage: Dataset 2: Further Fine-tune using Sampled $D^T$ and $D^F$.
    }
    \label{tab:ds2_s23}
\end{table*}

\begin{table*}[t!]
    \centering
    \small
    \resizebox{\linewidth}{!}
    {
        \begin{tabular}{llllllllll}
\hline
\multicolumn{1}{l|}{\textbf{\begin{tabular}[c]{@{}l@{}}\# of \\ Novelties\end{tabular}}} & \textbf{\begin{tabular}[c]{@{}l@{}}Known \\ Class\\ precision\end{tabular}} & \textbf{\begin{tabular}[c]{@{}l@{}}Known\\ class \\ recall\end{tabular}} & \multicolumn{1}{l|}{\textbf{\begin{tabular}[c]{@{}l@{}}Known \\ Class\\ F1\end{tabular}}} & \textbf{\begin{tabular}[c]{@{}l@{}}Novel\\ Class \\ Precision\end{tabular}} & \textbf{\begin{tabular}[c]{@{}l@{}}Novel\\ Class\\ Recall\end{tabular}} & \multicolumn{1}{l|}{\textbf{\begin{tabular}[c]{@{}l@{}}Novel \\ Class \\ F1\end{tabular}}} & \textbf{\begin{tabular}[c]{@{}l@{}}Overall\\ Precision\end{tabular}} & \textbf{\begin{tabular}[c]{@{}l@{}}Overall\\ Recall\end{tabular}} & \textbf{\begin{tabular}[c]{@{}l@{}}Overall\\ F1\end{tabular}} \\ \hline
\multicolumn{1}{l|}{\textbf{10000}}                                                      & 70.00                                                                       & 90.48                                                                    & \multicolumn{1}{l|}{78.93}                                                                & 86.80                                                                       & 54.61                                                                   & \multicolumn{1}{l|}{67.04}                                                                 & 78.40                                                                & 72.55                                                             & 75.36                                                         \\
\multicolumn{1}{l|}{\textbf{20000}}                                                      & 79.06                                                                       & 88.63                                                                    & \multicolumn{1}{l|}{83.57}                                                                & 88.26                                                                       & 74.07                                                                   & \multicolumn{1}{l|}{80.54}                                                                 & 83.66                                                                & 81.35                                                             & 82.49                                                         \\
\multicolumn{1}{l|}{\textbf{30000}}                                                      & 83.41                                                                       & 89.02                                                                    & \multicolumn{1}{l|}{86.12}                                                                & 91.10                                                                       & 82.49                                                                   & \multicolumn{1}{l|}{86.58}                                                                 & 87.25                                                                & 85.75                                                             & 86.49                                                         \\
\multicolumn{1}{l|}{\textbf{40000}}                                                      & 85.19                                                                       & 88.49                                                                    & \multicolumn{1}{l|}{86.81}                                                                & 91.91                                                                       & 86.37                                                                   & \multicolumn{1}{l|}{89.05}                                                                 & 88.55                                                                & 87.43                                                             & 87.99                                                         \\
\multicolumn{1}{l|}{\textbf{50000}}                                                      & 86.87                                                                       & 88.36                                                                    & \multicolumn{1}{l|}{87.61}                                                                & 92.39                                                                       & 88.83                                                                   & \multicolumn{1}{l|}{90.58}                                                                 & 89.63                                                                & 88.59                                                             & 89.11                                                         \\ \hline
\multicolumn{10}{c}{\textbf{Compute Mean}}                                                                                                                                                                                                                                                                                                                                                                                                                                                                                                                                                                                                                                                                                                                                                                    \\ \hline
\multicolumn{1}{l|}{\textbf{10000}}                                                      & 58.04                                                                       & 89.86                                                                    & \multicolumn{1}{l|}{70.53}                                                                & 8.73                                                                        & 14.06                                                                   & \multicolumn{1}{l|}{10.77}                                                                 & 33.39                                                                & 51.96                                                             & 40.65                                                         \\
\multicolumn{1}{l|}{\textbf{20000}}                                                      & 61.41                                                                       & 89.43                                                                    & \multicolumn{1}{l|}{72.82}                                                                & 15.30                                                                       & 22.70                                                                   & \multicolumn{1}{l|}{18.28}                                                                 & 38.35                                                                & 56.06                                                             & 45.54                                                         \\
\multicolumn{1}{l|}{\textbf{30000}}                                                      & 64.68                                                                       & 89.13                                                                    & \multicolumn{1}{l|}{74.96}                                                                & 23.33                                                                       & 31.21                                                                   & \multicolumn{1}{l|}{26.70}                                                                 & 44.00                                                                & 60.17                                                             & 50.83                                                         \\
\multicolumn{1}{l|}{\textbf{40000}}                                                      & 68.93                                                                       & 89.55                                                                    & \multicolumn{1}{l|}{77.90}                                                                & 30.62                                                                       & 39.67                                                                   & \multicolumn{1}{l|}{34.56}                                                                 & 49.78                                                                & 64.61                                                             & 56.23                                                         \\
\multicolumn{1}{l|}{\textbf{50000}}                                                      & 72.91                                                                       & 89.15                                                                    & \multicolumn{1}{l|}{80.22}                                                                & 40.10                                                                       & 48.95                                                                   & \multicolumn{1}{l|}{44.09}                                                                 & 56.50                                                                & 69.05                                                             & 62.15                                                         \\ \hline
\multicolumn{10}{c}{\textbf{Compute Euclid Distance}}                                                                                                                                                                                                                                                                                                                                                                                                                                                                                                                                                                                                                                                                                                                                                         \\ \hline
\multicolumn{1}{l|}{\textbf{10000}}                                                      & 78.29                                                                       & 89.46                                                                    & \multicolumn{1}{l|}{83.50}                                                                & 89.16                                                                       & 72.90                                                                   & \multicolumn{1}{l|}{80.21}                                                                 & 83.72                                                                & 81.18                                                             & 82.43                                                         \\
\multicolumn{1}{l|}{\textbf{20000}}                                                      & 84.24                                                                       & 88.78                                                                    & \multicolumn{1}{l|}{86.45}                                                                & 92.01                                                                       & 84.95                                                                   & \multicolumn{1}{l|}{88.34}                                                                 & 88.13                                                                & 86.87                                                             & 87.50                                                         \\
\multicolumn{1}{l|}{\textbf{30000}}                                                      & 86.28                                                                       & 88.35                                                                    & \multicolumn{1}{l|}{87.30}                                                                & 92.63                                                                       & 88.62                                                                   & \multicolumn{1}{l|}{90.58}                                                                 & 89.46                                                                & 88.48                                                             & 88.97                                                         \\
\multicolumn{1}{l|}{\textbf{40000}}                                                      & 87.84                                                                       & 88.96                                                                    & \multicolumn{1}{l|}{88.40}                                                                & 93.47                                                                       & 90.92                                                                   & \multicolumn{1}{l|}{92.18}                                                                 & 90.66                                                                & 89.94                                                             & 90.30                                                         \\
\multicolumn{1}{l|}{\textbf{50000}}                                                      & 88.36                                                                       & 88.41                                                                    & \multicolumn{1}{l|}{88.38}                                                                & 93.43                                                                       & 91.92                                                                   & \multicolumn{1}{l|}{92.67}                                                                 & 90.89                                                                & 90.16                                                             & 90.52                                                         \\ \hline
\multicolumn{10}{c}{\textbf{Compute Mahalanobis Distance}}                                                                                                                                                                                                                                                                                                                                                                                                                                                                                                                                                                                                                                                                                                                                                    \\ \hline
\multicolumn{1}{l|}{\textbf{10000}}                                                      & 77.53                                                                       & 89.67                                                                    & \multicolumn{1}{l|}{83.16}                                                                & 88.91                                                                       & 70.52                                                                   & \multicolumn{1}{l|}{78.65}                                                                 & 83.22                                                                & 80.10                                                             & 81.63                                                         \\
\multicolumn{1}{l|}{\textbf{20000}}                                                      & 83.83                                                                       & 88.98                                                                    & \multicolumn{1}{l|}{86.33}                                                                & 91.82                                                                       & 83.42                                                                   & \multicolumn{1}{l|}{87.42}                                                                 & 87.83                                                                & 86.20                                                             & 87.01                                                         \\
\multicolumn{1}{l|}{\textbf{30000}}                                                      & 86.29                                                                       & 88.76                                                                    & \multicolumn{1}{l|}{87.51}                                                                & 92.95                                                                       & 88.61                                                                   & \multicolumn{1}{l|}{90.73}                                                                 & 89.62                                                                & 88.68                                                             & 89.15                                                         \\
\multicolumn{1}{l|}{\textbf{40000}}                                                      & 87.90                                                                       & 88.79                                                                    & \multicolumn{1}{l|}{88.34}                                                                & 93.61                                                                       & 91.32                                                                   & \multicolumn{1}{l|}{92.45}                                                                 & 90.75                                                                & 90.05                                                             & 90.40                                                         \\
\multicolumn{1}{l|}{\textbf{50000}}                                                      & 88.45                                                                       & 88.49                                                                    & \multicolumn{1}{l|}{88.47}                                                                & 93.58                                                                       & 92.15                                                                   & \multicolumn{1}{l|}{92.86}                                                                 & 91.02                                                                & 90.32                                                             & 90.67                                                         \\ \hline
\multicolumn{10}{c}{\textbf{Compute Max Probability}}                                                                                                                                                                                                                                                                                                                                                                                                                                                                                                                                                                                                                                                                                                                                                         \\ \hline
\multicolumn{1}{l|}{\textbf{10000}}                                                      & 70.30                                                                       & 89.58                                                                    & \multicolumn{1}{l|}{78.78}                                                                & 77.38                                                                       & 53.36                                                                   & \multicolumn{1}{l|}{63.16}                                                                 & 73.84                                                                & 71.47                                                             & 72.64                                                         \\
\multicolumn{1}{l|}{\textbf{20000}}                                                      & 81.93                                                                       & 88.82                                                                    & \multicolumn{1}{l|}{85.24}                                                                & 90.36                                                                       & 79.85                                                                   & \multicolumn{1}{l|}{84.78}                                                                 & 86.14                                                                & 84.34                                                             & 85.23                                                         \\
\multicolumn{1}{l|}{\textbf{30000}}                                                      & 86.22                                                                       & 88.71                                                                    & \multicolumn{1}{l|}{87.45}                                                                & 92.28                                                                       & 87.57                                                                   & \multicolumn{1}{l|}{89.86}                                                                 & 89.25                                                                & 88.14                                                             & 88.69                                                         \\
\multicolumn{1}{l|}{\textbf{40000}}                                                      & 88.01                                                                       & 89.02                                                                    & \multicolumn{1}{l|}{88.51}                                                                & 93.34                                                                       & 90.93                                                                   & \multicolumn{1}{l|}{92.12}                                                                 & 90.68                                                                & 89.97                                                             & 90.32                                                         \\
\multicolumn{1}{l|}{\textbf{50000}}                                                      & 88.47                                                                       & 88.88                                                                    & \multicolumn{1}{l|}{88.67}                                                                & 93.77                                                                       & 92.06                                                                   & \multicolumn{1}{l|}{92.91}                                                                 & 91.12                                                                & 90.47                                                             & 90.79                                                         \\ \hline
\multicolumn{10}{c}{\textbf{Placeholders Algorithm}}                                                                                                                                                                                                                                                                                                                                                                                                                                                                                                                                                                                                                                                                                                                                                          \\ \hline
\multicolumn{1}{l|}{\textbf{10000}}                                                      & 69.76                                                                       & 89.82                                                                    & \multicolumn{1}{l|}{78.53}                                                                & 79.02                                                                       & 52.47                                                                   & \multicolumn{1}{l|}{63.06}                                                                 & 74.39                                                                & 71.14                                                             & 72.73                                                         \\
\multicolumn{1}{l|}{\textbf{20000}}                                                      & 80.36                                                                       & 89.10                                                                    & \multicolumn{1}{l|}{84.50}                                                                & 90.91                                                                       & 77.81                                                                   & \multicolumn{1}{l|}{83.85}                                                                 & 85.63                                                                & 83.45                                                             & 84.53                                                         \\
\multicolumn{1}{l|}{\textbf{30000}}                                                      & 85.01                                                                       & 88.88                                                                    & \multicolumn{1}{l|}{86.90}                                                                & 92.68                                                                       & 86.46                                                                   & \multicolumn{1}{l|}{89.46}                                                                 & 88.84                                                                & 87.67                                                             & 88.25                                                         \\
\multicolumn{1}{l|}{\textbf{40000}}                                                      & 87.65                                                                       & 89.18                                                                    & \multicolumn{1}{l|}{88.41}                                                                & 93.48                                                                       & 90.50                                                                   & \multicolumn{1}{l|}{91.97}                                                                 & 90.56                                                                & 89.84                                                             & 90.20                                                         \\
\multicolumn{1}{l|}{\textbf{50000}}                                                      & 88.53                                                                       & 88.51                                                                    & \multicolumn{1}{l|}{88.52}                                                                & 93.38                                                                       & 91.89                                                                   & \multicolumn{1}{l|}{92.63}                                                                 & 90.96                                                                & 90.20                                                             & 90.58                                                         \\ \hline
\multicolumn{10}{c}{\textbf{Few shot Open set Recognition}}                                                                                                                                                                                                                                                                                                                                                                                                                                                                                                                                                                                                                                                                                                                                                   \\ \hline
\end{tabular}
    }
    \caption{
   Novelty Accommodation Stage: Dataset 3: Retrain using $D^T$ and $D^F$
    }
    \label{tab:ds3_s21}
\end{table*}
\begin{table*}[t!]
    \centering
    \small
    \resizebox{\linewidth}{!}
    {
        \begin{tabular}{llllllllll}
\hline
\multicolumn{1}{l|}{\textbf{\begin{tabular}[c]{@{}l@{}}\# of \\ Novelties\end{tabular}}} & \textbf{\begin{tabular}[c]{@{}l@{}}Known \\ Class\\ precision\end{tabular}} & \textbf{\begin{tabular}[c]{@{}l@{}}Known\\ class \\ recall\end{tabular}} & \multicolumn{1}{l|}{\textbf{\begin{tabular}[c]{@{}l@{}}Known \\ Class\\ F1\end{tabular}}} & \textbf{\begin{tabular}[c]{@{}l@{}}Novel\\ Class \\ Precision\end{tabular}} & \textbf{\begin{tabular}[c]{@{}l@{}}Novel\\ Class\\ Recall\end{tabular}} & \multicolumn{1}{l|}{\textbf{\begin{tabular}[c]{@{}l@{}}Novel \\ Class \\ F1\end{tabular}}} & \textbf{\begin{tabular}[c]{@{}l@{}}Overall\\ Precision\end{tabular}} & \textbf{\begin{tabular}[c]{@{}l@{}}Overall\\ Recall\end{tabular}} & \textbf{\begin{tabular}[c]{@{}l@{}}Overall\\ F1\end{tabular}} \\ \hline
\multicolumn{1}{l|}{\textbf{10000}}                                                      & 61.09                                                                       & 8.26                                                                     & \multicolumn{1}{l|}{14.55}                                                                & 53.03                                                                       & 69.85                                                                   & \multicolumn{1}{l|}{60.29}                                                                 & 57.06                                                                & 39.06                                                             & 46.37                                                         \\
\multicolumn{1}{l|}{\textbf{20000}}                                                      & 11.00                                                                       & 1.10                                                                     & \multicolumn{1}{l|}{2.00}                                                                 & 56.49                                                                       & 86.58                                                                   & \multicolumn{1}{l|}{68.37}                                                                 & 33.74                                                                & 43.84                                                             & 38.13                                                         \\
\multicolumn{1}{l|}{\textbf{30000}}                                                      & 7.00                                                                        & 2.34                                                                     & \multicolumn{1}{l|}{3.51}                                                                 & 59.43                                                                       & 89.90                                                                   & \multicolumn{1}{l|}{71.56}                                                                 & 33.22                                                                & 46.12                                                             & 38.62                                                         \\
\multicolumn{1}{l|}{\textbf{40000}}                                                      & 5.00                                                                        & 0.02                                                                     & \multicolumn{1}{l|}{0.04}                                                                 & 58.84                                                                       & 92.42                                                                   & \multicolumn{1}{l|}{71.90}                                                                 & 31.92                                                                & 46.22                                                             & 37.76                                                         \\
\multicolumn{1}{l|}{\textbf{50000}}                                                      & 2.00                                                                        & 0.16                                                                     & \multicolumn{1}{l|}{0.30}                                                                 & 59.19                                                                       & 93.52                                                                   & \multicolumn{1}{l|}{72.50}                                                                 & 30.59                                                                & 46.84                                                             & 37.01                                                         \\ \hline
\multicolumn{10}{c}{\textbf{Compute Mean}}                                                                                                                                                                                                                                                                                                                                                                                                                                                                                                                                                                                                                                                                                                                                                                    \\ \hline
\multicolumn{1}{l|}{\textbf{10000}}                                                      & 20.73                                                                       & 1.59                                                                     & \multicolumn{1}{l|}{2.95}                                                                 & 2.04                                                                        & 14.72                                                                   & \multicolumn{1}{l|}{3.58}                                                                  & 11.38                                                                & 8.16                                                              & 9.50                                                          \\
\multicolumn{1}{l|}{\textbf{20000}}                                                      & 15.00                                                                       & 0.93                                                                     & \multicolumn{1}{l|}{1.75}                                                                 & 5.22                                                                        & 23.50                                                                   & \multicolumn{1}{l|}{8.54}                                                                  & 10.11                                                                & 12.22                                                             & 11.07                                                         \\
\multicolumn{1}{l|}{\textbf{30000}}                                                      & 12.00                                                                       & 0.47                                                                     & \multicolumn{1}{l|}{0.90}                                                                 & 8.75                                                                        & 32.31                                                                   & \multicolumn{1}{l|}{13.77}                                                                 & 10.38                                                                & 16.39                                                             & 12.71                                                         \\
\multicolumn{1}{l|}{\textbf{40000}}                                                      & 6.00                                                                        & 0.01                                                                     & \multicolumn{1}{l|}{0.02}                                                                 & 14.42                                                                       & 40.90                                                                   & \multicolumn{1}{l|}{21.32}                                                                 & 10.21                                                                & 20.46                                                             & 13.62                                                         \\
\multicolumn{1}{l|}{\textbf{50000}}                                                      & 4.00                                                                        & 0.04                                                                     & \multicolumn{1}{l|}{0.08}                                                                 & 20.16                                                                       & 50.39                                                                   & \multicolumn{1}{l|}{28.80}                                                                 & 12.08                                                                & 25.21                                                             & 16.33                                                         \\ \hline
\multicolumn{10}{c}{\textbf{Compute Euclid Distance}}                                                                                                                                                                                                                                                                                                                                                                                                                                                                                                                                                                                                                                                                                                                                                         \\ \hline
\multicolumn{1}{l|}{\textbf{10000}}                                                      & 24.98                                                                       & 1.43                                                                     & \multicolumn{1}{l|}{2.71}                                                                 & 55.64                                                                       & 86.18                                                                   & \multicolumn{1}{l|}{67.62}                                                                 & 40.31                                                                & 43.81                                                             & 41.99                                                         \\
\multicolumn{1}{l|}{\textbf{20000}}                                                      & 8.00                                                                        & 0.24                                                                     & \multicolumn{1}{l|}{0.47}                                                                 & 57.99                                                                       & 91.57                                                                   & \multicolumn{1}{l|}{71.01}                                                                 & 32.99                                                                & 45.91                                                             & 38.39                                                         \\
\multicolumn{1}{l|}{\textbf{30000}}                                                      & 5.00                                                                        & 0.36                                                                     & \multicolumn{1}{l|}{0.67}                                                                 & 58.09                                                                       & 93.42                                                                   & \multicolumn{1}{l|}{71.64}                                                                 & 31.55                                                                & 46.89                                                             & 37.72                                                         \\
\multicolumn{1}{l|}{\textbf{40000}}                                                      & 3.00                                                                        & 0.02                                                                     & \multicolumn{1}{l|}{0.04}                                                                 & 57.68                                                                       & 94.68                                                                   & \multicolumn{1}{l|}{71.69}                                                                 & 30.34                                                                & 47.35                                                             & 36.98                                                         \\
\multicolumn{1}{l|}{\textbf{50000}}                                                      & 3.00                                                                        & 0.01                                                                     & \multicolumn{1}{l|}{0.02}                                                                 & 58.78                                                                       & 94.69                                                                   & \multicolumn{1}{l|}{72.53}                                                                 & 30.89                                                                & 47.35                                                             & 37.39                                                         \\ \hline
\multicolumn{10}{c}{\textbf{Compute Mahalanobis Distance}}                                                                                                                                                                                                                                                                                                                                                                                                                                                                                                                                                                                                                                                                                                                                                    \\ \hline
\multicolumn{1}{l|}{\textbf{10000}}                                                      & 18.00                                                                       & 2.07                                                                     & \multicolumn{1}{l|}{3.71}                                                                 & 55.95                                                                       & 84.56                                                                   & \multicolumn{1}{l|}{67.34}                                                                 & 36.98                                                                & 43.31                                                             & 39.90                                                         \\
\multicolumn{1}{l|}{\textbf{20000}}                                                      & 7.00                                                                        & 0.55                                                                     & \multicolumn{1}{l|}{1.02}                                                                 & 57.81                                                                       & 91.36                                                                   & \multicolumn{1}{l|}{70.81}                                                                 & 32.40                                                                & 45.95                                                             & 38.00                                                         \\
\multicolumn{1}{l|}{\textbf{30000}}                                                      & 4.00                                                                        & 0.57                                                                     & \multicolumn{1}{l|}{1.00}                                                                 & 58.52                                                                       & 93.62                                                                   & \multicolumn{1}{l|}{72.02}                                                                 & 31.26                                                                & 47.10                                                             & 37.58                                                         \\
\multicolumn{1}{l|}{\textbf{40000}}                                                      & 2.00                                                                        & 0.14                                                                     & \multicolumn{1}{l|}{0.26}                                                                 & 59.38                                                                       & 94.64                                                                   & \multicolumn{1}{l|}{72.97}                                                                 & 30.69                                                                & 47.39                                                             & 37.25                                                         \\
\multicolumn{1}{l|}{\textbf{50000}}                                                      & 0.00                                                                        & 0.00                                                                     & \multicolumn{1}{l|}{0.00}                                                                 & 59.54                                                                       & 95.01                                                                   & \multicolumn{1}{l|}{73.20}                                                                 & 29.77                                                                & 47.51                                                             & 36.60                                                         \\ \hline
\multicolumn{10}{c}{\textbf{Compute Max Probability}}                                                                                                                                                                                                                                                                                                                                                                                                                                                                                                                                                                                                                                                                                                                                                         \\ \hline
\multicolumn{1}{l|}{\textbf{10000}}                                                      & 23.87                                                                       & 4.33                                                                     & \multicolumn{1}{l|}{7.33}                                                                 & 48.60                                                                       & 68.95                                                                   & \multicolumn{1}{l|}{57.01}                                                                 & 36.23                                                                & 36.64                                                             & 36.43                                                         \\
\multicolumn{1}{l|}{\textbf{20000}}                                                      & 8.00                                                                        & 0.06                                                                     & \multicolumn{1}{l|}{0.12}                                                                 & 56.57                                                                       & 89.78                                                                   & \multicolumn{1}{l|}{69.41}                                                                 & 32.28                                                                & 44.92                                                             & 37.57                                                         \\
\multicolumn{1}{l|}{\textbf{30000}}                                                      & 3.00                                                                        & 0.13                                                                     & \multicolumn{1}{l|}{0.25}                                                                 & 58.05                                                                       & 92.88                                                                   & \multicolumn{1}{l|}{71.45}                                                                 & 30.52                                                                & 46.51                                                             & 36.86                                                         \\
\multicolumn{1}{l|}{\textbf{40000}}                                                      & 2.00                                                                        & 0.02                                                                     & \multicolumn{1}{l|}{0.04}                                                                 & 59.85                                                                       & 94.23                                                                   & \multicolumn{1}{l|}{73.20}                                                                 & 30.93                                                                & 47.12                                                             & 37.35                                                         \\
\multicolumn{1}{l|}{\textbf{50000}}                                                      & 1.00                                                                        & 0.00                                                                     & \multicolumn{1}{l|}{0.00}                                                                 & 59.12                                                                       & 94.99                                                                   & \multicolumn{1}{l|}{72.88}                                                                 & 30.06                                                                & 47.50                                                             & 36.82                                                         \\ \hline
\multicolumn{10}{c}{\textbf{Placeholders Algorithm}}                                                                                                                                                                                                                                                                                                                                                                                                                                                                                                                                                                                                                                                                                                                                                          \\ \hline
\multicolumn{1}{l|}{\textbf{10000}}                                                      & 38.14                                                                       & 3.87                                                                     & \multicolumn{1}{l|}{7.03}                                                                 & 42.99                                                                       & 69.57                                                                   & \multicolumn{1}{l|}{53.14}                                                                 & 40.57                                                                & 36.72                                                             & 38.55                                                         \\
\multicolumn{1}{l|}{\textbf{20000}}                                                      & 9.97                                                                        & 0.21                                                                     & \multicolumn{1}{l|}{0.41}                                                                 & 56.54                                                                       & 89.17                                                                   & \multicolumn{1}{l|}{69.20}                                                                 & 33.26                                                                & 44.69                                                             & 38.14                                                         \\
\multicolumn{1}{l|}{\textbf{30000}}                                                      & 4.00                                                                        & 0.06                                                                     & \multicolumn{1}{l|}{0.12}                                                                 & 58.77                                                                       & 92.85                                                                   & \multicolumn{1}{l|}{71.98}                                                                 & 31.38                                                                & 46.46                                                             & 37.46                                                         \\
\multicolumn{1}{l|}{\textbf{40000}}                                                      & 2.00                                                                        & 0.01                                                                     & \multicolumn{1}{l|}{0.02}                                                                 & 59.66                                                                       & 94.14                                                                   & \multicolumn{1}{l|}{73.04}                                                                 & 30.83                                                                & 47.07                                                             & 37.26                                                         \\
\multicolumn{1}{l|}{\textbf{50000}}                                                      & 1.00                                                                        & 0.05                                                                     & \multicolumn{1}{l|}{0.10}                                                                 & 59.44                                                                       & 94.73                                                                   & \multicolumn{1}{l|}{73.05}                                                                 & 30.22                                                                & 47.39                                                             & 36.90                                                         \\ \hline
\multicolumn{10}{c}{\textbf{Few shot Open set Recognition}}                                                                                                                                                                                                                                                                                                                                                                                                                                                                                                                                                                                                                                                                                                                                                   \\ \hline
\end{tabular}
    }
    \caption{
   Novelty Accommodation Stage: Dataset 3: Further Fine-tune using $D^F$.
    }
    \label{tab:ds3_s22}
\end{table*}
\begin{table*}[t!]
    \centering
    \small
    \resizebox{\linewidth}{!}
    {
        \begin{tabular}{llllllllll}
\hline
\multicolumn{1}{l|}{\textbf{\begin{tabular}[c]{@{}l@{}}\# of \\ Novelties\end{tabular}}} & \textbf{\begin{tabular}[c]{@{}l@{}}Known \\ Class\\ precision\end{tabular}} & \textbf{\begin{tabular}[c]{@{}l@{}}Known\\ class \\ recall\end{tabular}} & \multicolumn{1}{l|}{\textbf{\begin{tabular}[c]{@{}l@{}}Known \\ Class\\ F1\end{tabular}}} & \textbf{\begin{tabular}[c]{@{}l@{}}Novel\\ Class \\ Precision\end{tabular}} & \textbf{\begin{tabular}[c]{@{}l@{}}Novel\\ Class\\ Recall\end{tabular}} & \multicolumn{1}{l|}{\textbf{\begin{tabular}[c]{@{}l@{}}Novel \\ Class \\ F1\end{tabular}}} & \textbf{\begin{tabular}[c]{@{}l@{}}Overall\\ Precision\end{tabular}} & \textbf{\begin{tabular}[c]{@{}l@{}}Overall\\ Recall\end{tabular}} & \textbf{\begin{tabular}[c]{@{}l@{}}Overall\\ F1\end{tabular}} \\ \hline
\multicolumn{1}{l|}{\textbf{10000}}                                                      & 81.85                                                                       & 85.73                                                                    & \multicolumn{1}{l|}{83.75}                                                                & 77.75                                                                       & 67.71                                                                   & \multicolumn{1}{l|}{72.38}                                                                 & 79.80                                                                & 76.72                                                             & 78.23                                                         \\
\multicolumn{1}{l|}{\textbf{20000}}                                                      & 87.54                                                                       & 85.57                                                                    & \multicolumn{1}{l|}{86.54}                                                                & 84.59                                                                       & 83.74                                                                   & \multicolumn{1}{l|}{84.16}                                                                 & 86.06                                                                & 84.65                                                             & 85.35                                                         \\
\multicolumn{1}{l|}{\textbf{30000}}                                                      & 89.39                                                                       & 85.82                                                                    & \multicolumn{1}{l|}{87.57}                                                                & 87.03                                                                       & 88.01                                                                   & \multicolumn{1}{l|}{87.52}                                                                 & 88.21                                                                & 86.91                                                             & 87.56                                                         \\
\multicolumn{1}{l|}{\textbf{40000}}                                                      & 90.01                                                                       & 86.48                                                                    & \multicolumn{1}{l|}{88.21}                                                                & 88.53                                                                       & 89.79                                                                   & \multicolumn{1}{l|}{89.16}                                                                 & 89.27                                                                & 88.14                                                             & 88.70                                                         \\
\multicolumn{1}{l|}{\textbf{50000}}                                                      & 90.35                                                                       & 87.23                                                                    & \multicolumn{1}{l|}{88.76}                                                                & 89.81                                                                       & 91.12                                                                   & \multicolumn{1}{l|}{90.46}                                                                 & 90.08                                                                & 89.17                                                             & 89.62                                                         \\ \hline
\multicolumn{10}{c}{\textbf{Compute Mean}}                                                                                                                                                                                                                                                                                                                                                                                                                                                                                                                                                                                                                                                                                                                                                                    \\ \hline
\multicolumn{1}{l|}{\textbf{10000}}                                                      & 60.69                                                                       & 90.58                                                                    & \multicolumn{1}{l|}{72.68}                                                                & 7.41                                                                        & 14.23                                                                   & \multicolumn{1}{l|}{9.75}                                                                  & 34.05                                                                & 52.40                                                             & 41.28                                                         \\
\multicolumn{1}{l|}{\textbf{20000}}                                                      & 65.10                                                                       & 90.15                                                                    & \multicolumn{1}{l|}{75.60}                                                                & 13.30                                                                       & 22.92                                                                   & \multicolumn{1}{l|}{16.83}                                                                 & 39.20                                                                & 56.53                                                             & 46.30                                                         \\
\multicolumn{1}{l|}{\textbf{30000}}                                                      & 67.89                                                                       & 90.59                                                                    & \multicolumn{1}{l|}{77.61}                                                                & 20.53                                                                       & 31.52                                                                   & \multicolumn{1}{l|}{24.86}                                                                 & 44.21                                                                & 61.05                                                             & 51.28                                                         \\
\multicolumn{1}{l|}{\textbf{40000}}                                                      & 73.00                                                                       & 90.40                                                                    & \multicolumn{1}{l|}{80.77}                                                                & 28.22                                                                       & 39.92                                                                   & \multicolumn{1}{l|}{33.07}                                                                 & 50.61                                                                & 65.16                                                             & 56.97                                                         \\
\multicolumn{1}{l|}{\textbf{50000}}                                                      & 76.60                                                                       & 90.26                                                                    & \multicolumn{1}{l|}{82.87}                                                                & 36.91                                                                       & 49.37                                                                   & \multicolumn{1}{l|}{42.24}                                                                 & 56.75                                                                & 69.81                                                             & 62.61                                                         \\ \hline
\multicolumn{10}{c}{\textbf{Compute Euclid Distance}}                                                                                                                                                                                                                                                                                                                                                                                                                                                                                                                                                                                                                                                                                                                                                         \\ \hline
\multicolumn{1}{l|}{\textbf{10000}}                                                      & 85.32                                                                       & 87.97                                                                    & \multicolumn{1}{l|}{86.62}                                                                & 86.95                                                                       & 81.83                                                                   & \multicolumn{1}{l|}{84.31}                                                                 & 86.14                                                                & 84.90                                                             & 85.52                                                         \\
\multicolumn{1}{l|}{\textbf{20000}}                                                      & 88.53                                                                       & 88.27                                                                    & \multicolumn{1}{l|}{88.40}                                                                & 89.91                                                                       & 88.44                                                                   & \multicolumn{1}{l|}{89.17}                                                                 & 89.22                                                                & 88.35                                                             & 88.78                                                         \\
\multicolumn{1}{l|}{\textbf{30000}}                                                      & 89.74                                                                       & 89.35                                                                    & \multicolumn{1}{l|}{89.54}                                                                & 91.56                                                                       & 90.55                                                                   & \multicolumn{1}{l|}{91.05}                                                                 & 90.65                                                                & 89.95                                                             & 90.30                                                         \\
\multicolumn{1}{l|}{\textbf{40000}}                                                      & 90.03                                                                       & 89.28                                                                    & \multicolumn{1}{l|}{89.65}                                                                & 92.15                                                                       & 91.62                                                                   & \multicolumn{1}{l|}{91.88}                                                                 & 91.09                                                                & 90.45                                                             & 90.77                                                         \\
\multicolumn{1}{l|}{\textbf{50000}}                                                      & 90.96                                                                       & 89.34                                                                    & \multicolumn{1}{l|}{90.14}                                                                & 92.21                                                                       & 92.58                                                                   & \multicolumn{1}{l|}{92.39}                                                                 & 91.58                                                                & 90.96                                                             & 91.27                                                         \\ \hline
\multicolumn{10}{c}{\textbf{Compute Mahalanobis Distance}}                                                                                                                                                                                                                                                                                                                                                                                                                                                                                                                                                                                                                                                                                                                                                    \\ \hline
\multicolumn{1}{l|}{\textbf{10000}}                                                      & 84.75                                                                       & 87.44                                                                    & \multicolumn{1}{l|}{86.07}                                                                & 85.76                                                                       & 79.61                                                                   & \multicolumn{1}{l|}{82.57}                                                                 & 85.26                                                                & 83.52                                                             & 84.38                                                         \\
\multicolumn{1}{l|}{\textbf{20000}}                                                      & 88.19                                                                       & 88.47                                                                    & \multicolumn{1}{l|}{88.33}                                                                & 90.19                                                                       & 88.10                                                                   & \multicolumn{1}{l|}{89.13}                                                                 & 89.19                                                                & 88.28                                                             & 88.73                                                         \\
\multicolumn{1}{l|}{\textbf{30000}}                                                      & 89.59                                                                       & 88.70                                                                    & \multicolumn{1}{l|}{89.14}                                                                & 91.41                                                                       & 90.69                                                                   & \multicolumn{1}{l|}{91.05}                                                                 & 90.50                                                                & 89.70                                                             & 90.10                                                         \\
\multicolumn{1}{l|}{\textbf{40000}}                                                      & 90.82                                                                       & 89.31                                                                    & \multicolumn{1}{l|}{90.06}                                                                & 91.88                                                                       & 92.13                                                                   & \multicolumn{1}{l|}{92.00}                                                                 & 91.35                                                                & 90.72                                                             & 91.03                                                         \\
\multicolumn{1}{l|}{\textbf{50000}}                                                      & 91.16                                                                       & 89.51                                                                    & \multicolumn{1}{l|}{90.33}                                                                & 92.55                                                                       & 93.09                                                                   & \multicolumn{1}{l|}{92.82}                                                                 & 91.85                                                                & 91.30                                                             & 91.57                                                         \\ \hline
\multicolumn{10}{c}{\textbf{Compute Max Probability}}                                                                                                                                                                                                                                                                                                                                                                                                                                                                                                                                                                                                                                                                                                                                                         \\ \hline
\multicolumn{1}{l|}{\textbf{10000}}                                                      & 78.52                                                                       & 89.09                                                                    & \multicolumn{1}{l|}{83.47}                                                                & 80.19                                                                       & 64.44                                                                   & \multicolumn{1}{l|}{71.46}                                                                 & 79.36                                                                & 76.76                                                             & 78.04                                                         \\
\multicolumn{1}{l|}{\textbf{20000}}                                                      & 87.03                                                                       & 88.68                                                                    & \multicolumn{1}{l|}{87.85}                                                                & 89.04                                                                       & 85.28                                                                   & \multicolumn{1}{l|}{87.12}                                                                 & 88.03                                                                & 86.98                                                             & 87.50                                                         \\
\multicolumn{1}{l|}{\textbf{30000}}                                                      & 89.45                                                                       & 88.28                                                                    & \multicolumn{1}{l|}{88.86}                                                                & 90.24                                                                       & 89.79                                                                   & \multicolumn{1}{l|}{90.01}                                                                 & 89.84                                                                & 89.03                                                             & 89.43                                                         \\
\multicolumn{1}{l|}{\textbf{40000}}                                                      & 90.66                                                                       & 88.28                                                                    & \multicolumn{1}{l|}{89.45}                                                                & 90.99                                                                       & 91.76                                                                   & \multicolumn{1}{l|}{91.37}                                                                 & 90.82                                                                & 90.02                                                             & 90.42                                                         \\
\multicolumn{1}{l|}{\textbf{50000}}                                                      & 91.05                                                                       & 89.12                                                                    & \multicolumn{1}{l|}{90.07}                                                                & 91.95                                                                       & 92.62                                                                   & \multicolumn{1}{l|}{92.28}                                                                 & 91.50                                                                & 90.87                                                             & 91.18                                                         \\ \hline
\multicolumn{10}{c}{\textbf{Placeholders Algorithm}}                                                                                                                                                                                                                                                                                                                                                                                                                                                                                                                                                                                                                                                                                                                                                          \\ \hline
\multicolumn{1}{l|}{\textbf{10000}}                                                      & 76.23                                                                       & 89.90                                                                    & \multicolumn{1}{l|}{82.50}                                                                & 80.92                                                                       & 64.15                                                                   & \multicolumn{1}{l|}{71.57}                                                                 & 78.57                                                                & 77.03                                                             & 77.79                                                         \\
\multicolumn{1}{l|}{\textbf{20000}}                                                      & 85.80                                                                       & 89.42                                                                    & \multicolumn{1}{l|}{87.57}                                                                & 89.87                                                                       & 83.76                                                                   & \multicolumn{1}{l|}{86.71}                                                                 & 87.83                                                                & 86.59                                                             & 87.21                                                         \\
\multicolumn{1}{l|}{\textbf{30000}}                                                      & 89.06                                                                       & 89.68                                                                    & \multicolumn{1}{l|}{89.37}                                                                & 91.44                                                                       & 89.43                                                                   & \multicolumn{1}{l|}{90.42}                                                                 & 90.25                                                                & 89.55                                                             & 89.90                                                         \\
\multicolumn{1}{l|}{\textbf{40000}}                                                      & 90.03                                                                       & 89.08                                                                    & \multicolumn{1}{l|}{89.55}                                                                & 91.97                                                                       & 91.42                                                                   & \multicolumn{1}{l|}{91.69}                                                                 & 91.00                                                                & 90.25                                                             & 90.62                                                         \\
\multicolumn{1}{l|}{\textbf{50000}}                                                      & 90.96                                                                       & 88.93                                                                    & \multicolumn{1}{l|}{89.93}                                                                & 91.83                                                                       & 92.40                                                                   & \multicolumn{1}{l|}{92.11}                                                                 & 91.39                                                                & 90.66                                                             & 91.02                                                         \\ \hline
\multicolumn{10}{c}{\textbf{Few shot Open set Recognition}}                                                                                                                                                                                                                                                                                                                                                                                                                                                                                                                                                                                                                                                                                                                                                   \\ \hline
\end{tabular}
    }
    \caption{
   Novelty Accommodation Stage: Dataset 3: Further Fine-tune using Sampled $D^T$ and $D^F$.
    }
    \label{tab:ds3_s23}
\end{table*}


\end{document}